\documentclass[10pt,twocolumn,letterpaper]{article}

\usepackage[pagenumbers]{cvpr} %

\usepackage[dvipsnames]{xcolor}
\usepackage{multirow}

\definecolor{cvprblue}{rgb}{0.21,0.49,0.74}
\usepackage[pagebackref,breaklinks,colorlinks,citecolor=cvprblue]{hyperref}
\usepackage{colortbl}
\usepackage{tikz}

\usepackage{appendix}

\def\etal{\emph{et al}\onedot}

\title{Cross-Image Attention for Zero-Shot Appearance Transfer \vspace*{-0.35cm}}

\begin{document}

\author{
Yuval Alaluf$^*$ \hspace{0.65cm} 
Daniel Garibi$^*$ \hspace{0.65cm} 
Or Patashnik \hspace{0.65cm} 
Hadar Averbuch-Elor \hspace{0.65cm}
Daniel Cohen-Or \\ \\[-0.15cm]
Tel Aviv University \\ \\[-0.25cm]
\small\url{https://garibida.github.io/cross-image-attention/}
}

\twocolumn[{%
\maketitle
\renewcommand\twocolumn[1][]{#1}%
\begin{center}
    \centering
    \vspace*{-0.6cm}
    \includegraphics[width=0.975\linewidth]{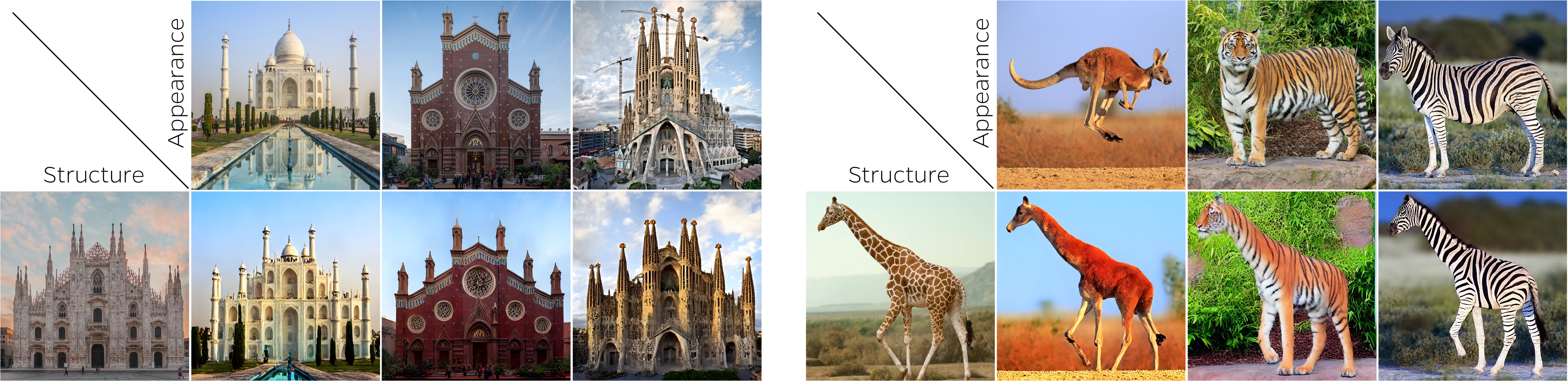}
    \vspace{-0.2cm}
    \captionsetup{type=figure}\caption{
    Given two images depicting a source structure and a target appearance, our method generates an image merging the structure of one image with the appearance of the other. We do so in a \textit{zero-shot} manner, with no optimization or model training required while supporting appearance transfer across images that may differ in size and shape. 
    }
    \vspace{0.4cm}
    \label{fig:teaser}
\end{center}
}]

\begin{abstract}
\vspace*{-0.2cm}
Recent advancements in text-to-image generative models have demonstrated a remarkable ability to capture a deep semantic understanding of images. In this work, we leverage this semantic knowledge to transfer the visual appearance between objects that share similar semantics but may differ significantly in shape. To achieve this, we build upon the self-attention layers of these generative models and introduce a cross-image attention mechanism that implicitly establishes semantic correspondences across images. Specifically, given a pair of images --- one depicting the target structure and the other specifying the desired appearance --- our cross-image attention combines the queries corresponding to the structure image with the keys and values of the appearance image. This operation, when applied during the denoising process, leverages the established semantic correspondences to generate an image combining the desired structure and appearance. In addition, to improve the output image quality, we harness three mechanisms that either manipulate the noisy latent codes or the model's internal representations throughout the denoising process. Importantly, our approach is zero-shot, requiring no optimization or training. Experiments show that our method is effective across a wide range of object categories and is robust to variations in shape, size, and viewpoint between the two input images.
\end{abstract}
\section{Introduction}

The rapid growth and adoption of powerful generative models have granted users an unprecedented level of freedom to create stunning, diverse visual content with relative ease~\cite{ramesh2022hierarchical, nichol2021glide, rombach2022high, saharia2022photorealistic, balaji2023ediffi, kandinsky2, ding2022cogview2}. In parallel with these advancements in generative capabilities, many have sought new avenues to gain greater control over the \textit{manipulation} of visual content using these generative models.

In this work, we explore image manipulation within the context of appearance transfer, where we aim to transfer the visual appearance of a concept from one image to a concept present in another image.
Consider, for example, transferring the appearance of a zebra to a giraffe (see~\Cref{fig:teaser}). Successfully accomplishing this task requires first associating semantically similar regions between the giraffe and zebra (e.g., their legs, head, and neck) and then transferring the zebra's appearance in a realistic manner through these associations without altering the structure of the giraffe.  
Furthermore, a particular challenge in this task is establishing these associations across images containing objects from different categories that vary in shape, as well as images with differing viewpoints and illuminations.
Previous attempts assume that appearance transfer is performed between objects of similar shape~\cite{tumanyan2022splicing,epstein2023selfguidance,mou2023dragondiffusion}, or require training a model for a specific class of objects~\cite{park2020swapping,zhu2017unpaired}.

\def\thefootnote{*}\footnotetext{Denotes equal contribution}

\begin{figure}
    \centering
    \setlength{\tabcolsep}{0.5pt}
    \addtolength{\belowcaptionskip}{-10pt}
    {\scriptsize
    \begin{tabular}{c c c c c}
        \begin{tabular}{c} Structure \\ Image \end{tabular} & 
        \begin{tabular}{c} Appearance \\ Image \end{tabular} & 
        \begin{tabular}{c} Output \\ Image \end{tabular} &
        \begin{tabular}{c} Appearance \\ Colormap \end{tabular} & 
        \begin{tabular}{c} Output \\ Correspondences \end{tabular} \\ 

        \includegraphics[width=0.0925\textwidth]{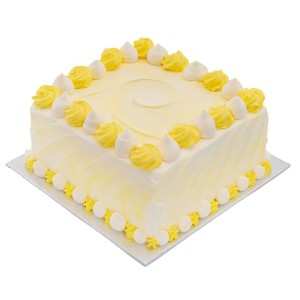} &
        \includegraphics[width=0.0925\textwidth]{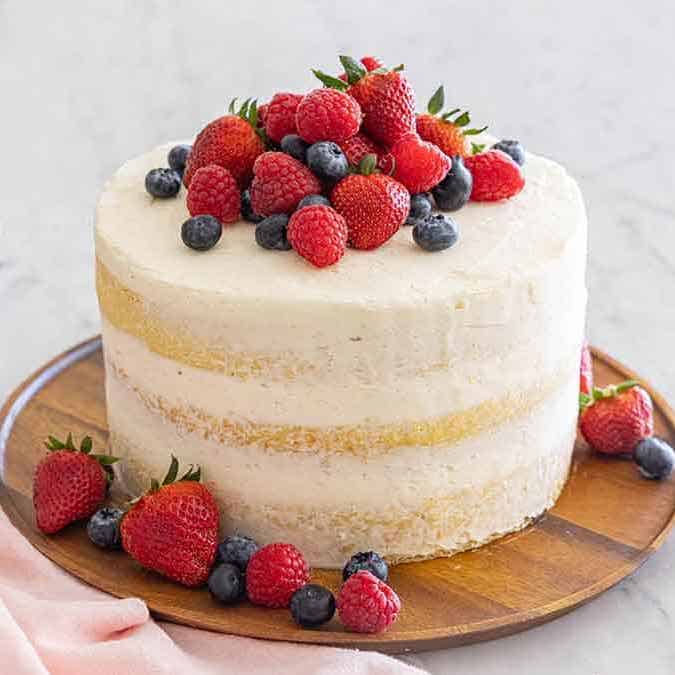} &
        \includegraphics[width=0.0925\textwidth]{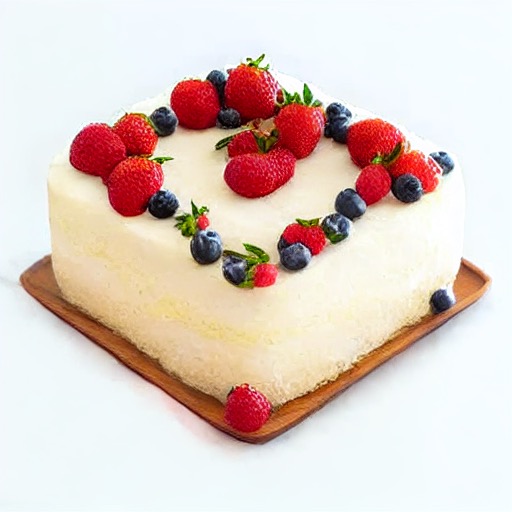} &
        \includegraphics[width=0.0925\textwidth]{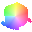} &
        \includegraphics[width=0.0925\textwidth]{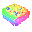} \\

        \includegraphics[width=0.0925\textwidth]{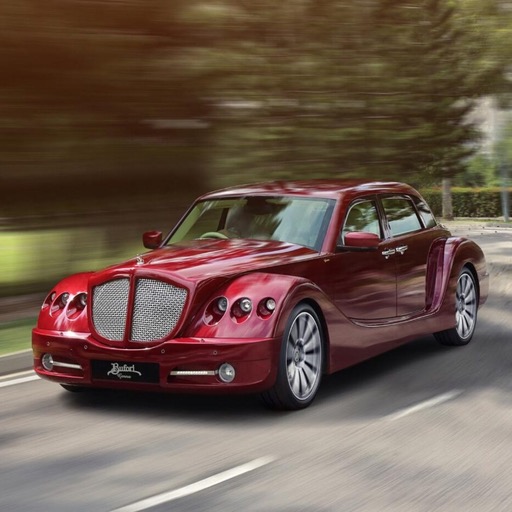} &
        \includegraphics[width=0.0925\textwidth]{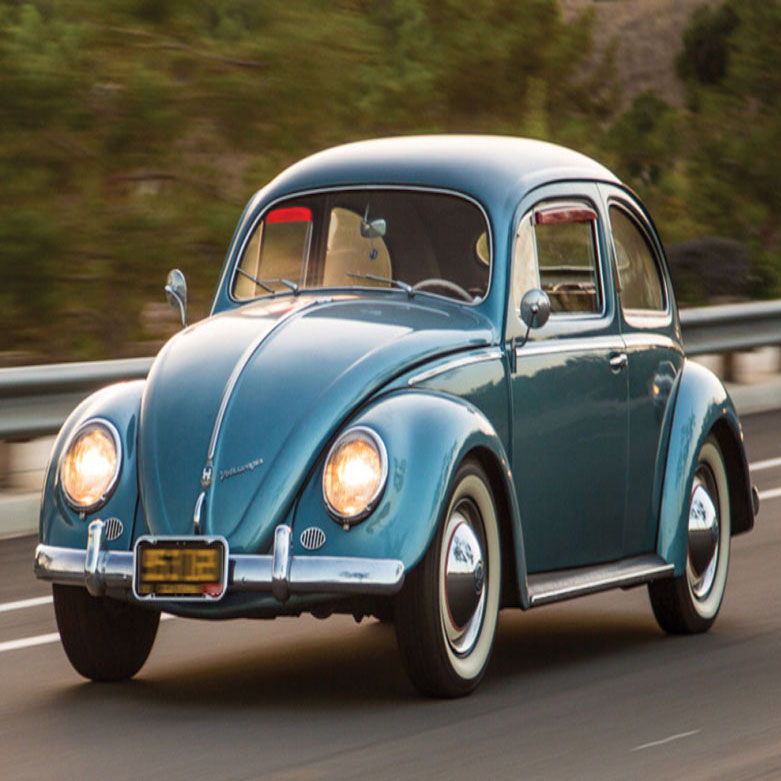} &
        \includegraphics[width=0.0925\textwidth]{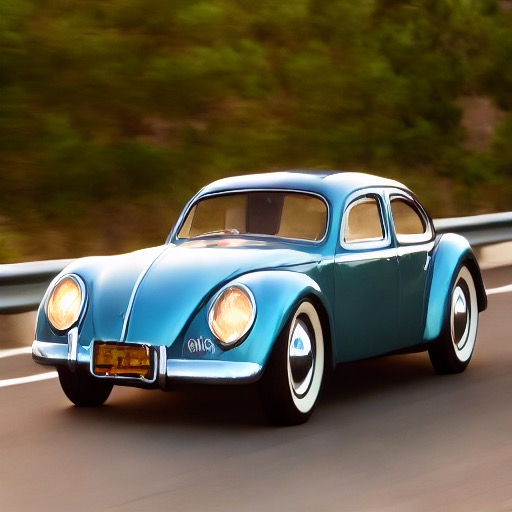} &    
        \includegraphics[width=0.0925\textwidth]{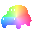} &
        \includegraphics[width=0.0925\textwidth]{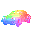} \\
        
    \end{tabular}
    }
    \vspace{-0.3cm}
    \caption{
    Implicitly finding correspondences via our cross-image attention applied between two images. For each pixel in the structure image, we identify the pixel in the appearance image that attains the highest activation in its cross-image attention map. The output correspondences represent the pixel mapping from the structure to the appearance image using the maximum activations. As shown, these correspondences are semantically aligned (e.g., matching the fruits on the cake and the bodies of the cars). 
    }
    \label{fig:colormap_figure}
\end{figure}

Analyzing the inner workings of recent large-scale diffusion models, many works have demonstrated that the cross- and self-attention mechanisms of the denoising network implicitly encode strong semantic information from the generated image~\cite{hertz2022prompt,pnpDiffusion2022,cao2023masactrl, geyer2023tokenflow,patashnik2023localizing,tang2023emergent}. Building on the functions of the queries, keys, and values within these self-attention layers, our key insight is to employ the self-attention mechanism across \textit{different} images, which we term \textit{Cross-Image Attention}. As illustrated in~\Cref{fig:colormap_figure}, when applied to images featuring distinct subjects with varying shapes and structures, this cross-image attention forms strong associations between similar semantic regions in the two images.

More specifically, given an appearance image and a structure image, we begin by inverting the two images into the latent space of a pretrained text-to-image diffusion model~\cite{rombach2022high}.
Then, at each timestep of the denoising process, we compute a modified self-attention map by multiplying the \textit{queries} corresponding to the structure image with the \textit{keys} of the appearance image. This cross-image operation establishes implicit semantic correspondences between the two images, without requiring additional supervision, as illustrated in~\Cref{fig:colormap_figure}. Then, by multiplying the resulting cross-image attention map by the \textit{values} of the appearance image, we can accurately transfer each pixel from the appearance image to the corresponding, semantically similar pixel(s) in the structure image.

While the cross-image mechanism is conceptually simple, we observe that it alone is not sufficient for attaining an accurate semantic transfer between the two images, often leading to noticeable artifacts in the resulting image. 
We attribute these artifacts to the domain gap between the queries of the structure image and the keys and values of the appearance image.
To address this challenge, we employ three mechanisms aimed at enhancing transfer quality.
First, we amplify the variance of the cross-image attention maps, making them more focused on capturing only the most semantically similar image regions.
Second, we adapt the classifier-free guidance technique~\cite{nichol2021glide,ho2022classifier} to the task of appearance transfer and strengthen the influence of our cross-image attention operation on the generated image during the denoising process.
Finally, we leverage the AdaIN~\cite{huang2017adain} mechanism to align the image statistics of the appearance and output images, better preserving the color of the appearance image.

We illustrate the versatility of our cross-image attention and show its effectiveness for zero-shot appearance transfer across a wide range of object domains. This includes challenging image pairs containing objects with substantial variations in shape, viewpoint, and number of instances. We also perform quantitative comparisons to existing methods, further demonstrating that our results better capture the target appearance while preserving the source structure.
\section{Related Works}

\paragraph{Appearance Transfer}
The task of appearance transfer can be seen as a specialized form of image-to-image translation. However, unlike Neural Style Transfer~\cite{gatys2017controlling,jing2019neural,kolkin2019style, yoo2019photorealistic,park2019arbitrary}, which focuses on transferring a \textit{global} artistic style across images, our focus is on \textit{semantic style transfer}, where we aim to transfer the appearance between semantically related regions in two images.

Early generative-based approaches trained a Generative Adversarial Network (GAN)\cite{goodfellow2020generative} on a large collection of either paired\cite{isola2017image} or unpaired images~\cite{zhu2017unpaired,liu2017unsupervised,park2020contrastive, Katzir2020CrossDomainCD, 2017dualgan}. Notably, Park~\etal~\cite{park2020swapping} introduced Swapping Autoencoders, where they train an autoencoder, separately encoding the structure and the appearance of an image. Then, to transfer the appearance from one image to another, they take the structure representation from one image and the appearance code from the other and pass them together to the decoder. However, this approach necessitates training a dedicated generator for each target domain (e.g., churches or cats) and requires collecting a large domain-specific dataset.

To reduce the level of supervision required, several methods learn a mapping using a single exemplar~\cite{liao2017visual,cohen2019bidirectional, benaim2021structural,lin2020tuigan,sohn2023styledrop}. 
Tumanyan~\etal~\cite{tumanyan2022splicing} train a dedicated generator for a single image pair and utilize a pretrained DINO-ViT~\cite{caron2021emerging,dosovitskiy2020image} to extract structure and style information from input images, injecting them into the training process to guide the transformation. 
This approach, however, requires training a dedicated generator for each pair of images, which takes dozens of minutes per input. Moreover, the technique mainly works well between images with relatively similar shapes. 

Most similar to our approach, recent works have sought to leverage powerful large-scale diffusion models for appearance transfer without additional inputs or model training~\cite{kwon2022diffusion, epstein2023selfguidance, mou2023dragondiffusion}.
These methods typically incorporate losses applied over the noisy latent codes to guide the denoising process toward generating images depicting the structure of a given image while adapting its appearance.
However, unlike our method, the appearance losses in these works rely on a global appearance descriptor and do not consider the semantic correspondences between the images. As a result, these methods are often limited to a coarse appearance transfer or constrained to transferring appearance between objects of the same category, size, and shape.

In contrast to the above approaches, our method operates with no training or per-image optimization and respects the semantic correspondences between the two images when transferring appearance. Furthermore, our method requires only a single forward pass through a pretrained diffusion model and is applicable to diverse image pairs that may contain cross-domain subjects.

\vspace{-0.15cm}
\paragraph{Semantic Correspondences with Generative Models}
The task of finding correspondences between two images has been a longstanding challenge in computer vision, ranging from classical approaches~\cite{lowe2004distinctive,bay2006surf,rublee2011orb} to learning-based techniques~\cite{simonyan2014learning,aberman2018neural,hong2022integrative,cho2022cats++,min2021convolutional,rocco2018neighbourhood,seo2018attentive,ofri2023neural}.
With the rapid improvements in generative models, many have explored adapting these models for the task of semantic correspondence.
Peebles~\etal~\cite{peebles2022gan} leverage the latent space of a pretrained GAN to train a Spatial Transformer~\cite{jaderberg2015spatial} tasked with densely aligning a set of images from a specific domain.
In the context of diffusion models, numerous works have demonstrated that the intermediate features of the denoising network of a pretrained diffusion model~\cite{rombach2022high} can effectively establish semantic correspondences across different object categories~\cite{tang2023emergent, luo2023diffusion, zhang2023tale, hedlin2023unsupervised}.

\vspace{-0.15cm}
\paragraph{Image Editing with Diffusion Models}
Building on the recent advancements in large-scale diffusion models~\cite{ramesh2022hierarchical,nichol2021glide,rombach2022high,saharia2022photorealistic,kandinsky2}, many works have explored new avenues to gain more precise control over the generative process~\cite{nichol2021glide,avrahami2023spatext,bar2023multidiffusion,li2023gligen,voynov2023sketch,zhang2023adding,lhhuang2023composer}, further utilizing these models for various downstream editing tasks~\cite{avrahami2022blended,brooks2022instructpix2pix,couairon2022diffedit,meng2022sdedit,2021ilvr,kawar2023imagic}.
To provide users with additional control over the generation and editing process, recent works have also utilized user-provided spatial conditions to specify the region that should be altered~\cite{avrahami2022blended,couairon2022diffedit,avrahami2023spatext,bar2023multidiffusion,li2023gligen,voynov2023sketch,zhang2023adding}.
Notably, numerous works have shown that manipulating the internal representations of the denoising network, particularly its attention layers, is effective for image editing~\cite{hertz2022prompt,pnpDiffusion2022,epstein2023selfguidance,cao2023masactrl,geyer2023tokenflow,patashnik2023localizing,parmar2023zero,ge2023expressive,liew2022magicmix}, as well as for finer control over the image generation process~\cite{chefer2023attend,li2023divide,rassin2023linguistic}.

Recently, MasaCtrl~\cite{cao2023masactrl} demonstrated that freezing the keys and values of the self-attention layers when performing non-rigid edits of an image more faithfully preserves the image's original appearance.
TokenFlow~\cite{geyer2023tokenflow} and Infusion~\cite{khandelwal2023infusion} extend this technique to preserve the appearance of different frames when editing a video.
In our method, we also inject keys and values into the self-attention layers. However, unlike the methods mentioned above, this injection is performed between \textit{different} images rather than between an image and its output edit.

\setlength{\abovedisplayskip}{5pt}
\setlength{\belowdisplayskip}{5pt}

\section{Method} \label{sec:method}
Given a pair of images $(I^{struct},I^{app})$, we wish to generate an output image $I^{out}$ depicting the structure of the subject present in $I^{struct}$ with the appearance of the subject in $I^{app}$.
To perform the transfer, we utilize a pretrained text-to-image diffusion model, namely Stable Diffusion~\cite{rombach2022high}. 
We first briefly review concepts related to the self-attention layers within image diffusion models. We then introduce our Cross-Image Attention mechanism, which is the core of our proposed method, and then describe how it can be utilized for zero-shot appearance transfer.

\subsection{Preliminaries}
We begin by describing the self-attention layers that compose the denoising U-Net within the image diffusion model. At each timestep $t$ of the denoising process, the noised latent code $z_t$ is fed as input to the denoising network. Consider a specific self-attention layer $\ell$. The intermediate features of the network at $\ell$, denoted $\phi_\ell(z_t)$, are first projected into queries $Q=f_Q(\phi(z_t))$, keys $K=f_K(\phi(z_t))$, and values $V=f_V(\phi(z_t))$ through learned linear projections $f_Q, f_K, f_V$.

For each query vector $q_{i,j}$ located at spatial location $(i,j)$ of $Q$, we calculate a similarity score, or attention score, with respect to all keys in $K$, reflecting how relevant each key is to the corresponding query. These attention scores are then normalized using a softmax operation, defining the weight each value will have when updating the features at position $(i,j)$. Finally, the weighted values are aggregated to produce the output for each query position. Formally, we compute:
\begin{equation}
\begin{split}
    A_{(i,j)} & = \text{softmax} \left ( \frac{q_{i,j}\cdot K^T}{\sqrt{d}} \right ) \\
    \Delta \phi_{(i,j)} & =  A_{(i,j)} \cdot V,
\end{split}
\end{equation}
where $A_{(i,j)}$ represents the attention map at $(i,j)$ and $\Delta \phi_(i,j)$ denotes the aggregated output feature at $(i,j)$ used to update the spatial features $\phi(z_t)$.
This process is applied independently for all queries, enabling the model to capture correspondences across the entire image.

\subsection{Cross-Image Attention}
In a recent work, Cao~\etal~\cite{cao2023masactrl} explored the self-attention layers within the denoising network of a text-to-image diffusion model. They show that keeping the keys and values of these self-attention layers fixed aids in preserving the visual characteristics of objects when applying non-rigid manipulations over a given image. 

Building on this insight, in this work we demonstrate the significant roles played by the queries, keys, and values in encoding semantic information of the generated image. More specifically, we observe that one can utilize the queries, keys, and values from these self-attention layers to transfer semantic information between \textit{different} images. 
As shall be shown, the \textit{queries} determine the semantic meaning of each spatial location. Next, the \textit{keys} offer context for each query, allowing the model to weigh the importance of different parts of the image for a specific query position. Lastly, the values represent the content we aim to generate and define the information that will be used for determining the features of each query position.

To define our cross-image attention layer, we replace the keys and values corresponding to one image with the keys and values of another image. We demonstrate that by doing so, it is possible to implicitly transfer the visual appearance between semantically similar objects in the images. More precisely, we replace the keys and values corresponding to the output image $I^{out}$ with the keys and values corresponding to the appearance image $I^{app}$. Formally, the output of our cross-image attention layer is given by:
\begin{equation}~\label{eq:cross_image}
    \Delta \phi^{\text{cross}} = \text{softmax} \left ( \frac{Q_{out}\cdot {K_{app}}^T}{\sqrt{d}} \right ) V_{app}.
\end{equation}

\definecolor{point1}{HTML}{FA5434}
\definecolor{point2}{HTML}{FFFB00}
\definecolor{point3}{HTML}{00FF05}

\begin{figure}
    \centering
    \setlength{\tabcolsep}{1pt}
    \addtolength{\belowcaptionskip}{-10pt}
    {\small
    \centering

    \begin{tabular}{c c c c c}

        & &
        \raisebox{0.0025\textwidth}
        {\begin{tikzpicture}
            \fill[fill=point1, draw=black] (0,0) circle (0.10);
        \end{tikzpicture}} &
        \raisebox{0.0025\textwidth}
        {\begin{tikzpicture}
            \fill[fill=point2, draw=black] (0,0) circle (0.10);
        \end{tikzpicture}} &
        \raisebox{0.0025\textwidth}
        {\begin{tikzpicture}
            \fill[fill=point3, draw=black] (0,0) circle (0.10);
        \end{tikzpicture}} \\

        \raisebox{15pt}{\rotatebox{90}{$I^{struct}$}} &
        \includegraphics[width=0.1015\textwidth]{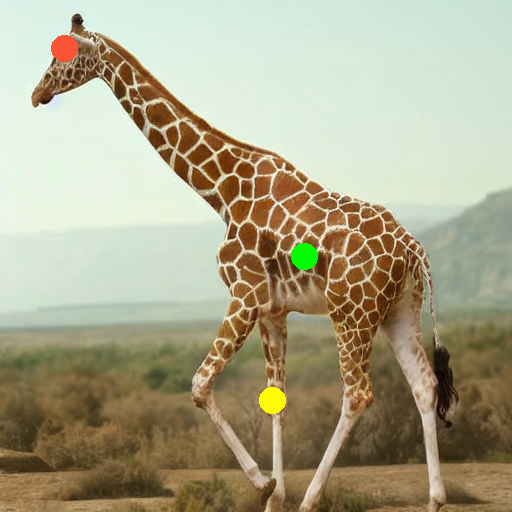} &
        \includegraphics[width=0.1015\textwidth]{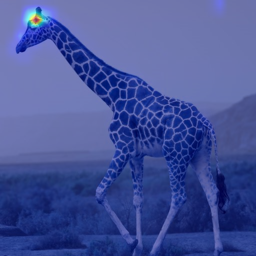} &
        \includegraphics[width=0.1015\textwidth]{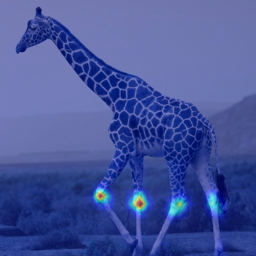} &
        \includegraphics[width=0.1015\textwidth]{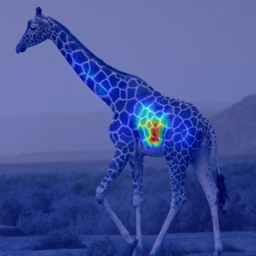} \\
        & & \multicolumn{3}{c}{ $Q_{struct} \cdot K_{struct}^T$  } \\ \\[-0.25cm]
        
        \raisebox{17.5pt}{\rotatebox{90}{$I^{app}$}} &
        \includegraphics[width=0.1015\textwidth]{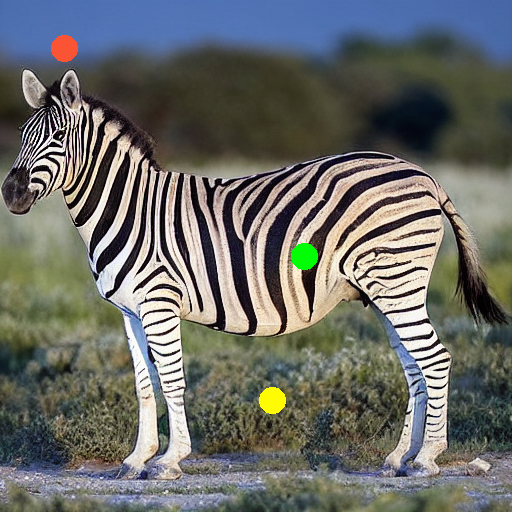} &
        \includegraphics[width=0.1015\textwidth]{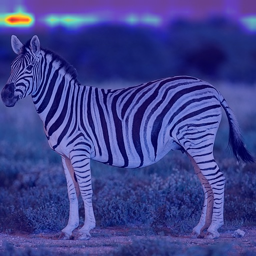} &
        \includegraphics[width=0.1015\textwidth]{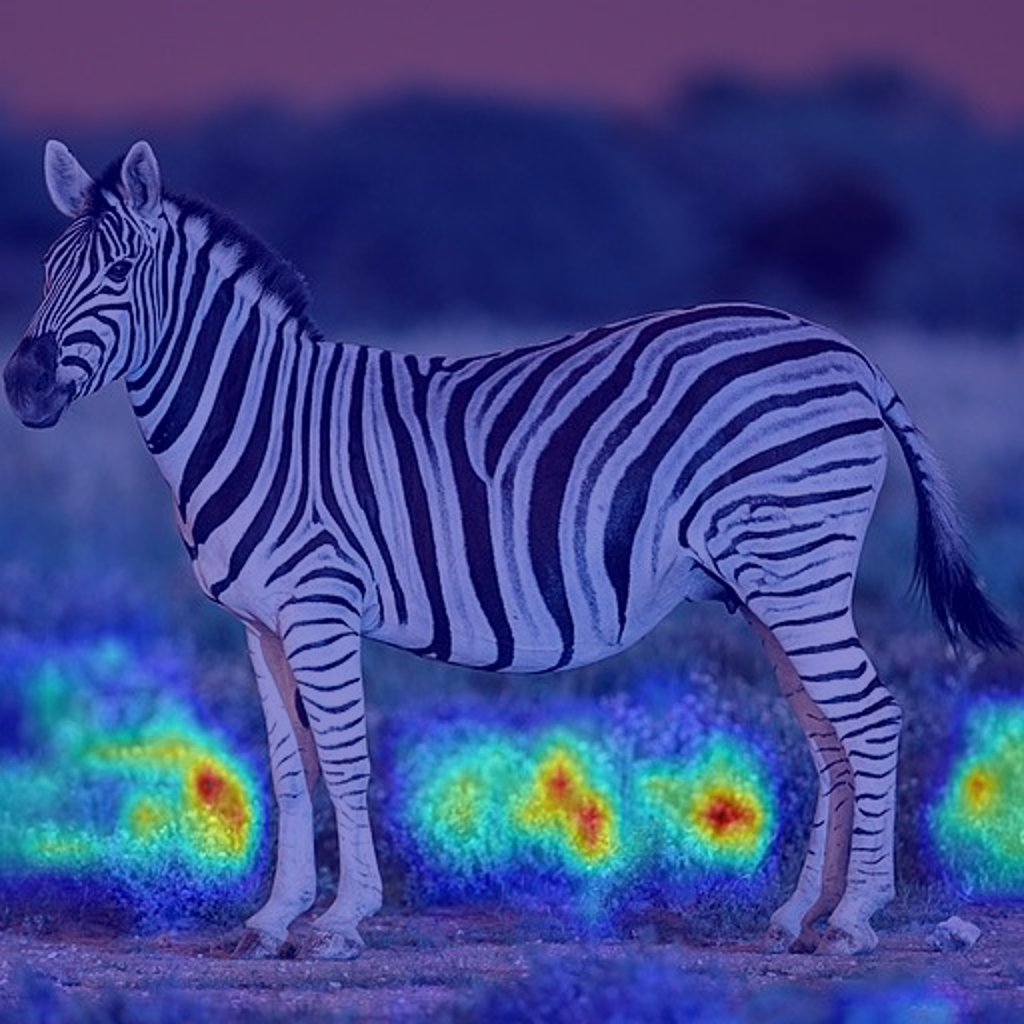}  &
        \includegraphics[width=0.1015\textwidth]{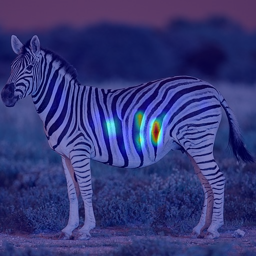} \\
        & & \multicolumn{3}{c}{ $Q_{app} \cdot K_{app}^T$ } \\ \\[-0.25cm]

        \raisebox{17.5pt}{\rotatebox{90}{$I^{out}$}} &
        \includegraphics[width=0.1015\textwidth]{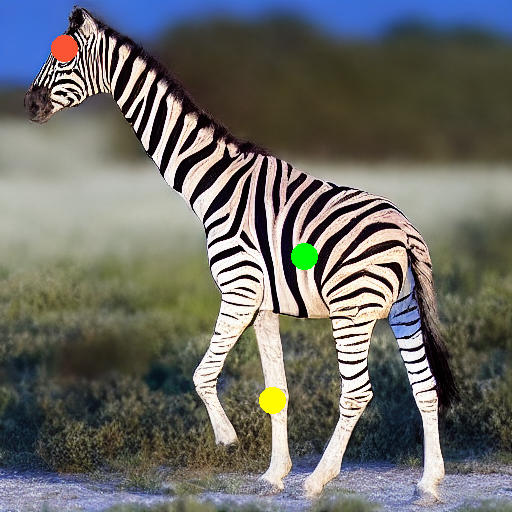} &
        \includegraphics[width=0.1015\textwidth]{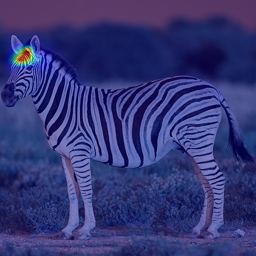} &
        \includegraphics[width=0.1015\textwidth]{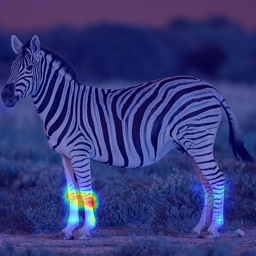 } &   
        \includegraphics[width=0.1015\textwidth]{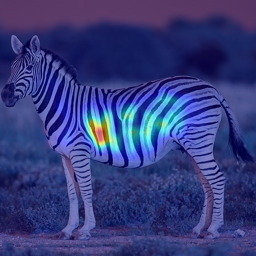} \\
        & & \multicolumn{3}{c}{ $Q_{struct} \cdot K_{app}^T$ }
        
    \end{tabular}
    
    }
    \vspace{-0.225cm}    
    \caption{
    Establishing correspondences through the keys and queries of self-attention and cross-image attention. Using color-coded markers, we denote three queries corresponding to different semantic regions of the structure image (the giraffe's head, leg, and body). These markers are placed in the same pixel locations in the three images.
    In each row, we illustrate the attention maps obtained by computing various dot products of the queries and keys of the two images computed at a single layer. In the first two rows, we show the self-attention maps obtained using queries and keys originating from the same image, resulting in each query focusing on semantically similar regions in the image. For instance, the yellow query attends to the legs of the giraffe in the structure image and to nearby grass pixels in the background of the appearance image. In the bottom row, we use our cross-image attention, aligning the queries $Q_{struct}$ with the keys $K_{app}$. In doing so, each query on the giraffe corresponds to semantically similar regions of the zebra. For example, the red query attends to the head of the zebra while the yellow query points to its legs.
    }
    \label{fig:points_correspondence}
    \vspace{-0.15cm}
\end{figure}

We illustrate the roles of the keys and queries in~\Cref{fig:points_correspondence}. In each column, we highlight one of three query locations marked by red, yellow, and green circles. For each row, we then display the attention maps obtained for each query location using different combinations of queries and keys corresponding to $I^{struct}$ and $I^{app}$.
As shown, when multiplying the keys and queries corresponding to the same image (i.e., computing $Q_{struct}\cdot K^T_{struct}$ or $Q_{app}\cdot K^T_{app}$), each query attends to semantically similar regions within the image. For example, consider the yellow query. In the attention of the structure image (row one), the query attends to the legs of the giraffe as it is located on the giraffe's leg. Conversely, in the appearance image (row two), the yellow query lies on the grass of the image background, and hence, the query attends to nearby grass-like pixels in the image. 

In the bottom row, we apply our cross-image attention mechanism and compute $Q_{struct}\cdot K_{app}^T$. As shown, the queries now attend to a semantically corresponding region in the zebra image. For example, the red query now attends to the head of the zebra while the yellow query attends to the leg of the zebra. Interestingly, these associations are established even though the two animals differ significantly in shape.  
Consequently, by multiplying the resulting attention maps with the values $V_{app}$ from the appearance image, we can accurately project semantically similar regions from the zebra image onto the giraffe image. This enables one to transfer the zebra's appearance onto the giraffe's structure.

\begin{figure*}
    \centering
    \includegraphics[width=0.975\textwidth]{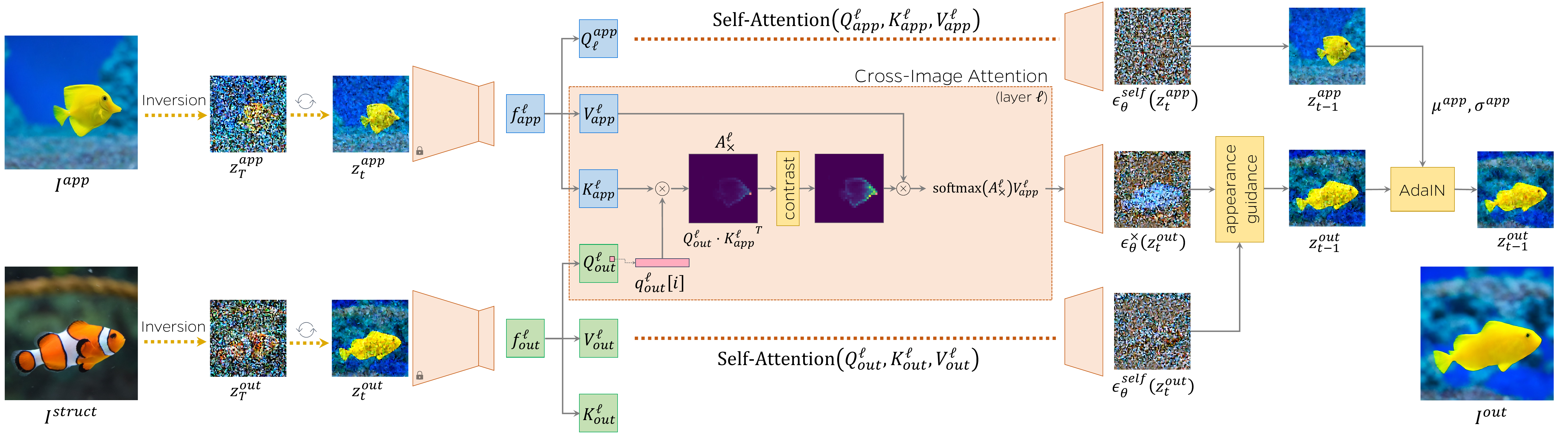} 
    \vspace{-0.1cm}
    \caption{
    Method overview. Given $I^{struct}$ and $I^{app}$, we begin by inverting the two images into the latent space of a pretrained image diffusion model, resulting in latents $z_T^{struct}$ and $z_T^{app}$. To initialize our output latent, we set $z_T^{out} = z_T^{struct}$. Consider some timestep $t$ and self-attention layer $\ell$. To compute the next latent $z_{T-1}^{out}$, we compute our cross-image attention map defined in~\Cref{eq:cross_image} by mixing the keys and values from $z_t^{app}$ with the query of $z_t^{out}$. To improve the output image quality, we introduce three extensions. First, we apply a contrast operation over the cross-image attention map, encouraging the $Q_{out}$ to attend to a smaller set of keys in $K_{app}$. Next, we introduce an appearance guidance mechanism akin to classifier-free guidance used for text-guided image synthesis. Finally, we apply an AdaIN operation over $z_{t-1}^{out}$ to better align with the feature statistics of $z_{t-1}^{app}$. This process is repeated across multiple timesteps of the denoising process and across multiple layers of the network decoder, resulting in the gradual appearance transfer from $I^{app}$ to $I^{struct}$.
    }
    \label{fig:method}
\end{figure*}

\subsection{Appearance Transfer}~\label{sec:app_transfer}
We now turn to describe how our cross-image attention mechanism can be utilized for semantic-based appearance transfer, as depicted in~\Cref{fig:method}. Given input images $I^{struct}$ and $I^{app}$, we begin by inverting them using the edit-friendly DDPM inversion introduced in Huberman~\etal~\cite{huberman2023edit}.  After obtaining the inverted latents, denoted by $z_T^{struct}$ and $z_T^{app}$, we perform a denoising process along two parallel paths, resulting in the reconstruction of $I^{app}$ and the generation of $I^{out}$. To initialize this denoising process, we set the latent $z_{T}^{out}$ representing our output image to be equal to $z_T^{struct}$. 

At each timestep $t$, we pass the two latent codes $z_{t}^{out}$ and $z_{t}^{app}$ to the denoising U-Net model. Within the decoder of the U-Net, we replace the standard self-attention with our cross-image attention and compute the modified output using~\Cref{eq:cross_image}. In practice, our cross-image attention replaces the standard self-attention layers in the U-Net decoder layers with output resolutions of $32\times 32$ and $64\times 64$. 

While this simple injection mechanism allows for the transfer of pixels from the appearance image to semantically similar regions in the structure image, it may still result in noticeable artifacts in the generated output image. We attribute this issue to the presence of a domain gap between the queries, keys, and values that are computed from latent codes of two distinct images, resulting in a lower-quality output image. To improve the output image quality, we introduce several additional mechanisms to guide the appearance transfer, detailed below.

\begin{figure}
    \centering
    \setlength{\tabcolsep}{0.5pt}
    \addtolength{\belowcaptionskip}{-10pt}
    {\scriptsize
    \begin{tabular}{c c c c c}

        \includegraphics[width=0.09\textwidth]{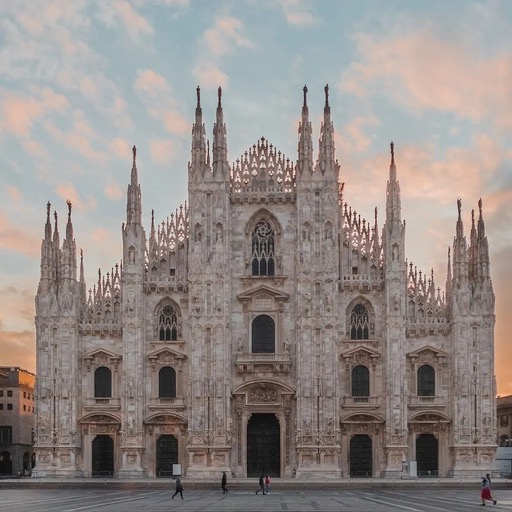} &
        \includegraphics[width=0.09\textwidth]{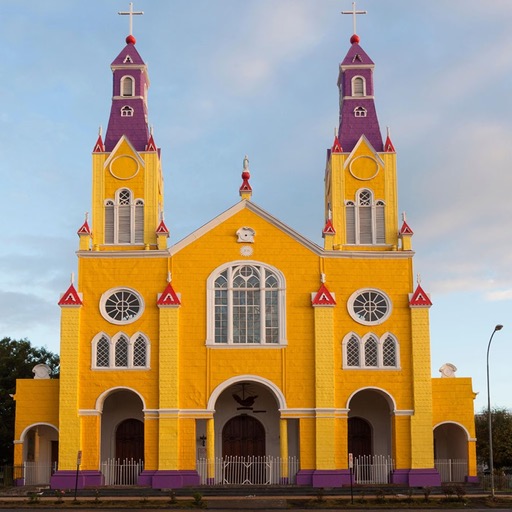} &
        \includegraphics[width=0.09\textwidth]{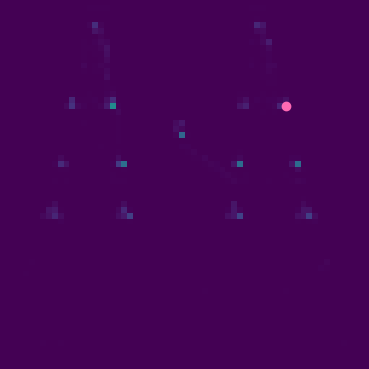} &
        \includegraphics[width=0.09\textwidth]{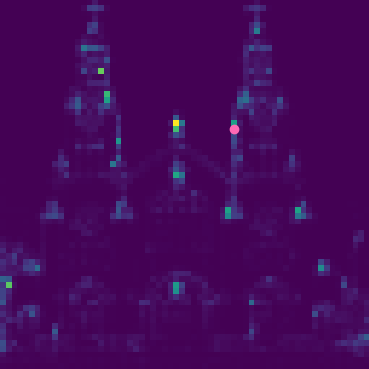} &
        \includegraphics[width=0.09\textwidth]{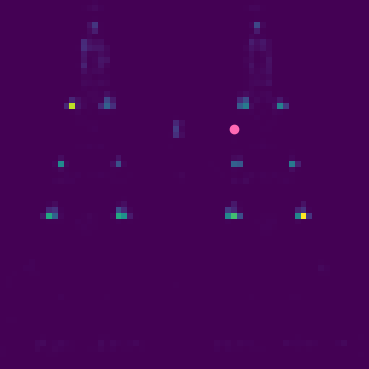} \\

        \includegraphics[width=0.09\textwidth]{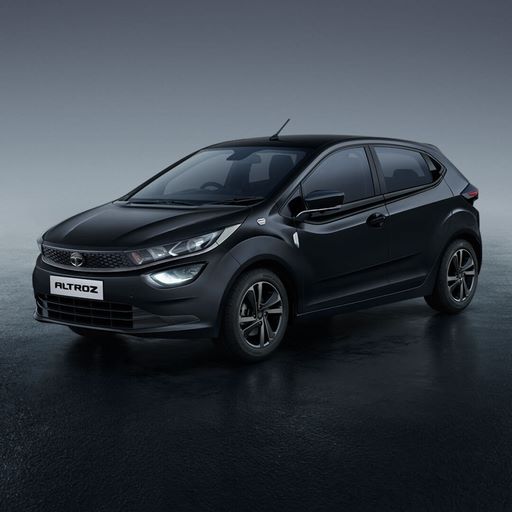} &
        \includegraphics[width=0.09\textwidth]{images/inputs/cars/vintage.jpg} &
        \includegraphics[width=0.09\textwidth]{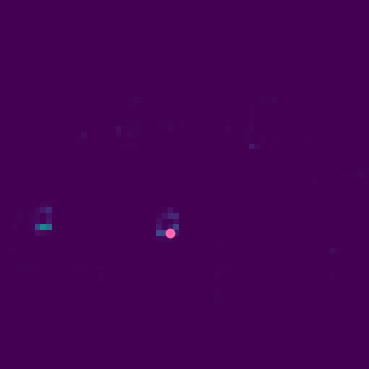} &
        \includegraphics[width=0.09\textwidth]{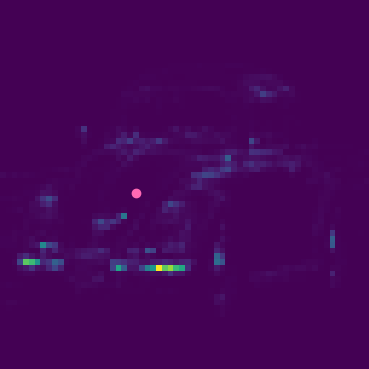} &
        \includegraphics[width=0.09\textwidth]{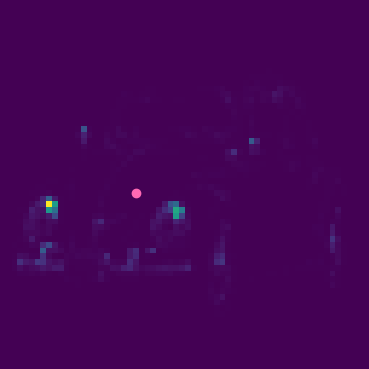} \\
        
        \begin{tabular}{c} Structure \\ Image  \end{tabular} & 
        \begin{tabular}{c} Appearance \\ Image \end{tabular} & 
        \begin{tabular}{c} Self- \\ Attention \end{tabular} &
        \begin{tabular}{c} Cross-Image \\ w/o Contrast \end{tabular} & 
        \begin{tabular}{c} Cross-Image \\ w/ Contrast \end{tabular}  \\
        
    \end{tabular}
    }
    \caption{
        Attention map contrasting. When applying the standard self-attention, queries often attend to a small set of semantically similar pixels (third column). In contrast, our cross-image attention map may cause a specific query (highlighted in pink) to attain high activations across many pixels across the entire image (fourth column). 
        By applying our contrast operation over the cross-image attention maps, the maps behave more similarly to the standard self-attention, focusing on the more semantically similar image regions (rightmost column).
    }
    \label{fig:contrast_visualization}
\end{figure}

\vspace{-0.2cm}
\paragraph{Attention Map Contrasting}
First, we observe that some queries of $z_t^{out}$ attain a high similarity to many keys of $z_t^{app}$. 
This can be observed in~\Cref{fig:contrast_visualization} where the cross-image attention map obtained by our method returns a sparse and unfocused attention map (fourth column). This is in contrast to the attention map obtained by the standard self-attention mechanism, which attains high similarities in a concentrated region of the image (third column). 
These sparse attention maps may lead to inaccurate transfers because the output value of each query is computed using an aggregation of pixels spanning many different image regions. This, in turn, can result in unwanted artifacts in the final image.

To encourage the attention maps to focus on more concentrated regions in the image, we apply a contrast operation to increase the variance of the attention maps. Given $A^\ell_\times$ obtained from our cross-image operation, we update
\begin{equation}
    A^\ell_\times \gets (A^\ell_\times - \mu(A^\ell_\times)) \beta + \mu(A^\ell_\times),
\end{equation}
where $\mu$ is the mean operation and $\beta$ is the contrast factor, empirically set to $\beta=1.67$. 
Note that this operation is applied before the attention map is multiplied with $V_{app}$. 

\begin{figure*}
    \centering
    \setlength{\tabcolsep}{0.5pt}
    \renewcommand{\arraystretch}{0.3}
    \addtolength{\belowcaptionskip}{-5pt}
    {\small

    \hspace{-0.2cm}
    \begin{minipage}{0.32\textwidth}
        \centering
        \begin{tabular}{c c c c}

        \includegraphics[width=0.25\textwidth]{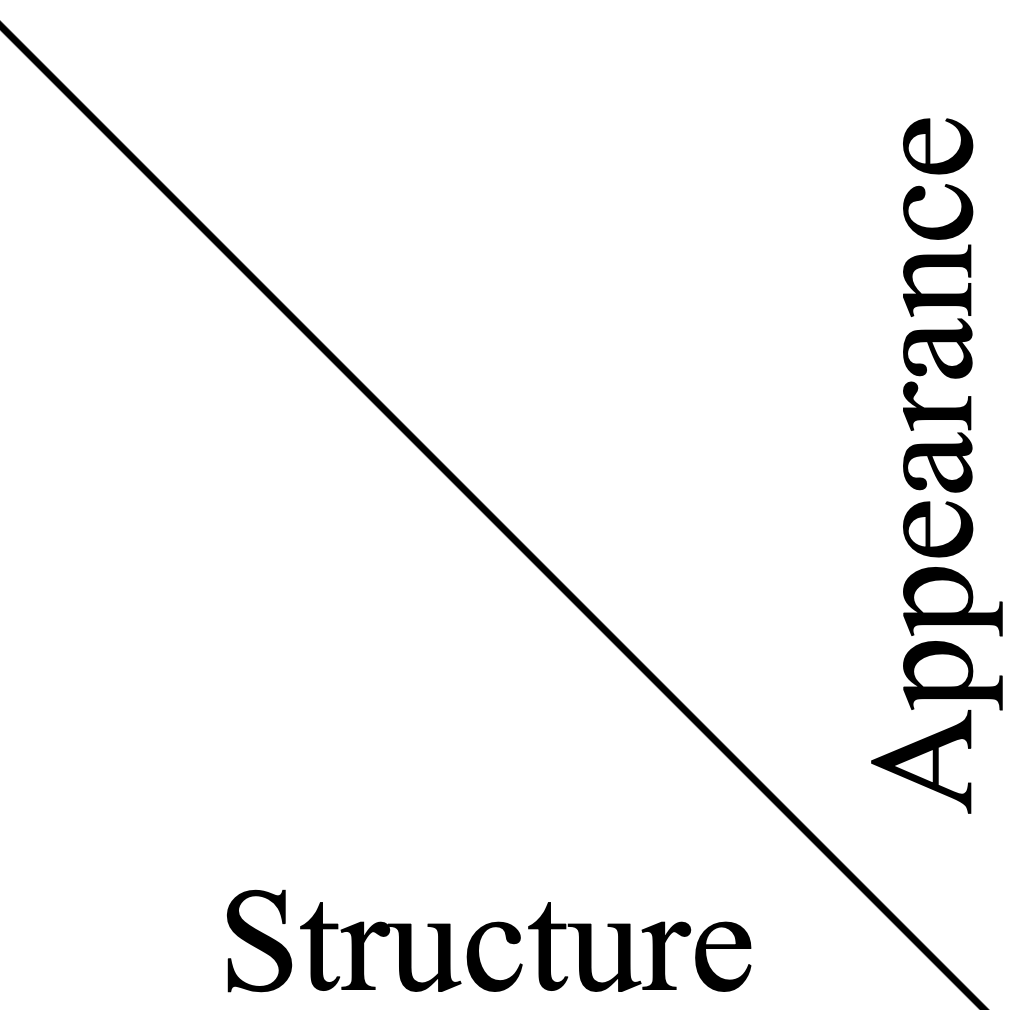} &
        \includegraphics[width=0.25\textwidth]{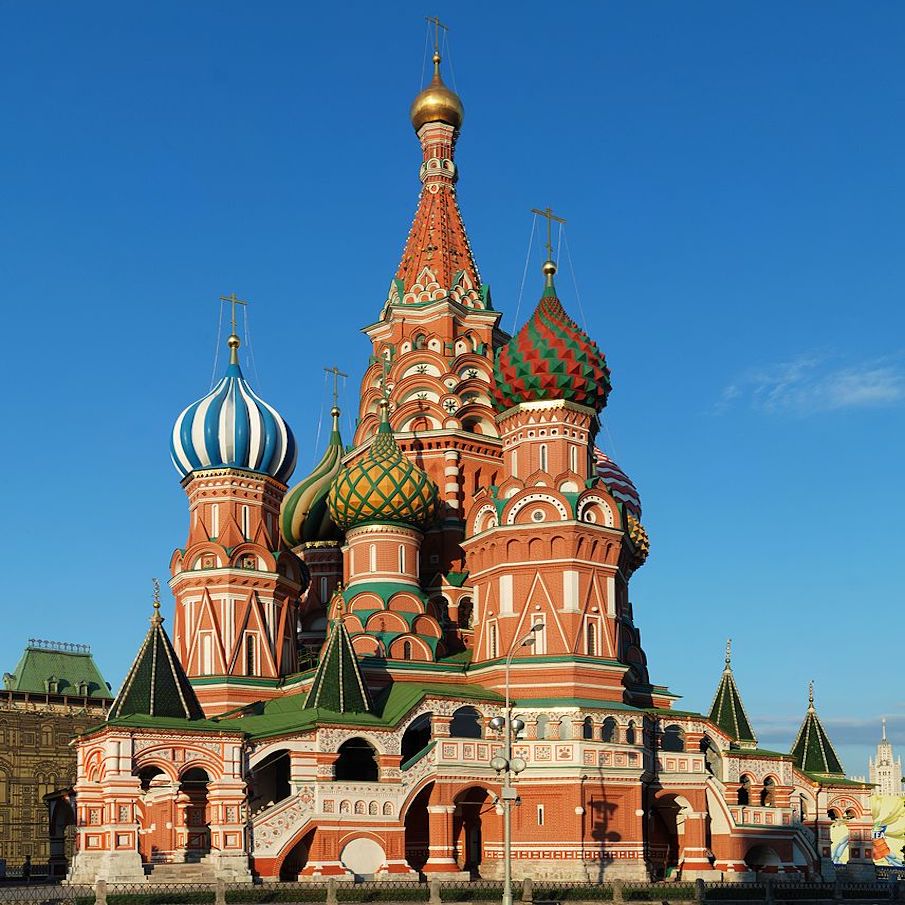} &
        \includegraphics[width=0.25\textwidth]{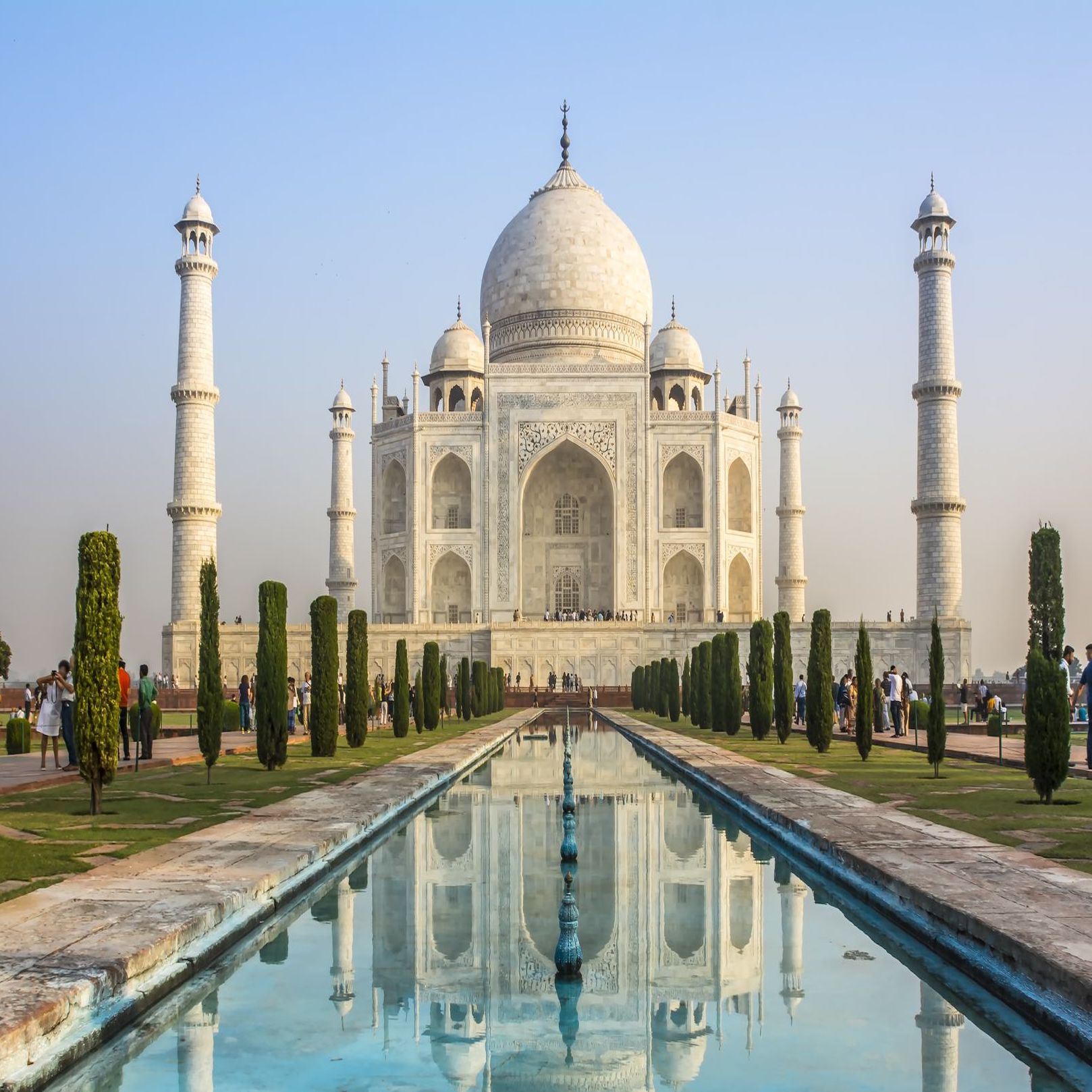} &
        \includegraphics[width=0.25\textwidth]{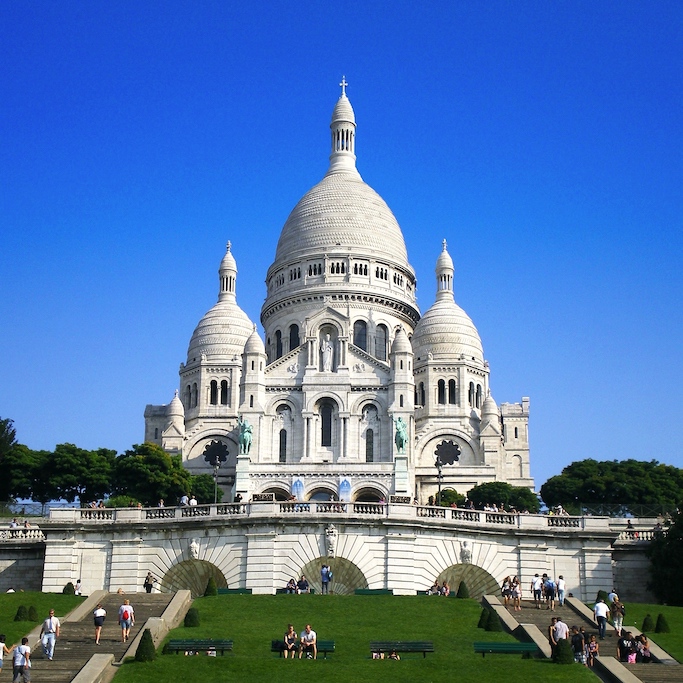} \\

        \includegraphics[width=0.25\textwidth]{images/inputs/buildings/saint_basil.jpg} &
        \includegraphics[width=0.25\textwidth]{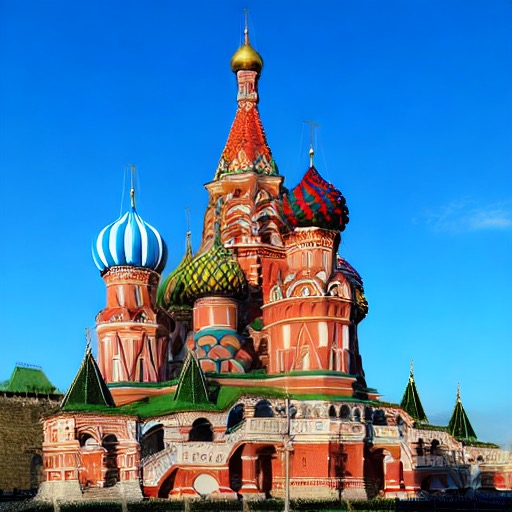} &
        \includegraphics[width=0.25\textwidth]{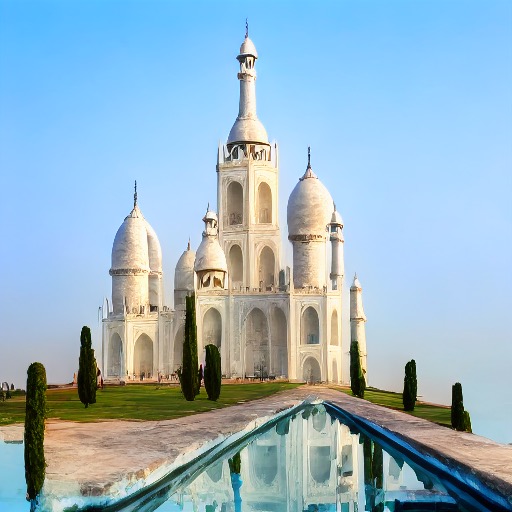} &
        \includegraphics[width=0.25\textwidth]{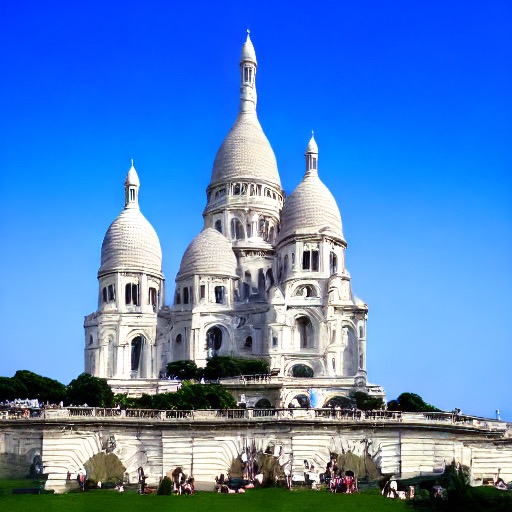} \\

        \includegraphics[width=0.25\textwidth]{images/inputs/buildings/taj_mahal.jpg} &
        \includegraphics[width=0.25\textwidth]{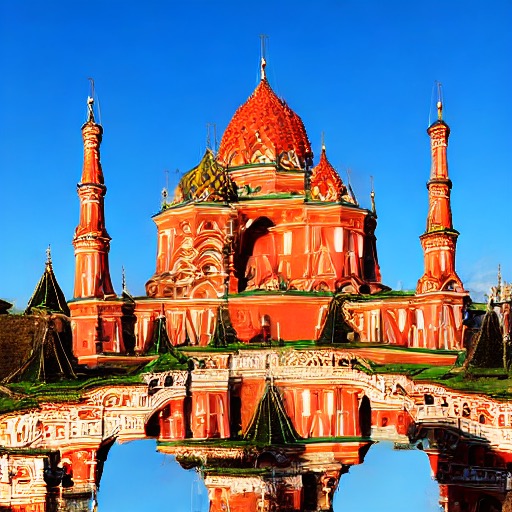} &
        \includegraphics[width=0.25\textwidth]{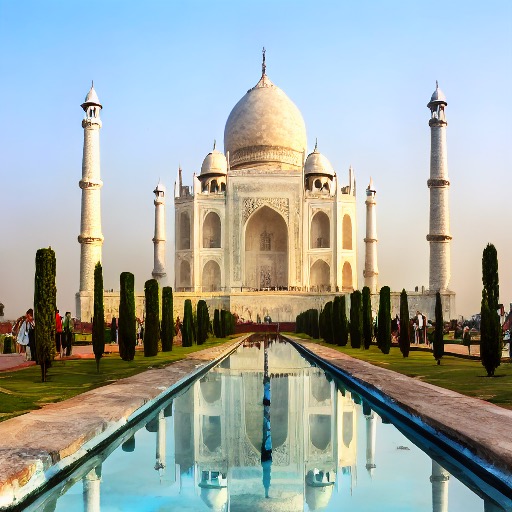} &
        \includegraphics[width=0.25\textwidth]{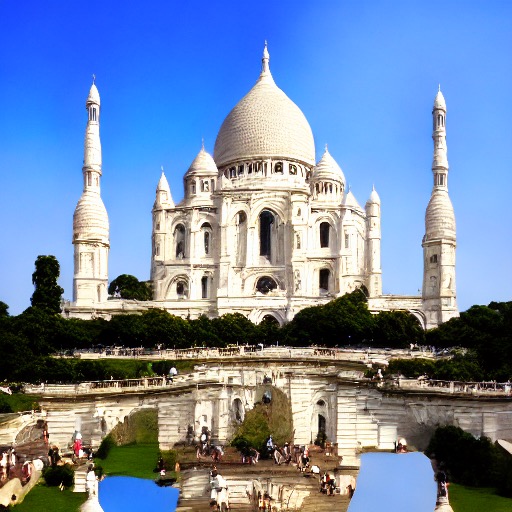} \\
        
        \includegraphics[width=0.25\textwidth]{images/inputs/buildings/Le_sacre_coeur.jpg} &
        \includegraphics[width=0.25\textwidth]{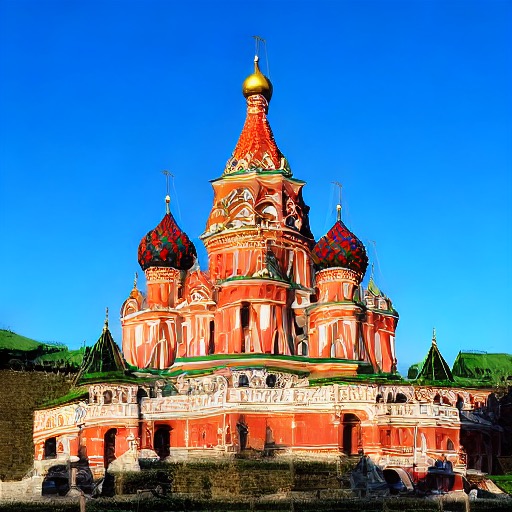} &
        \includegraphics[width=0.25\textwidth]{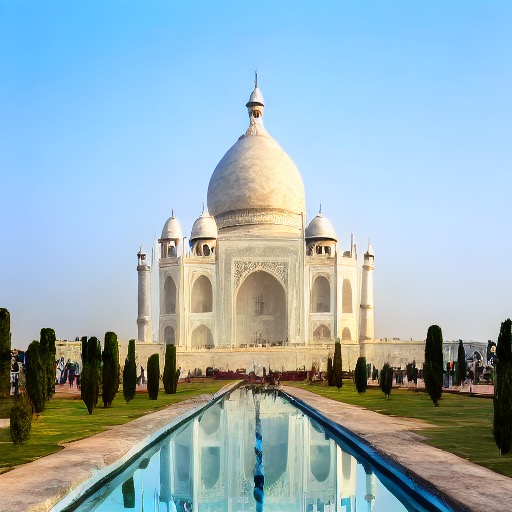} &
        \includegraphics[width=0.25\textwidth]{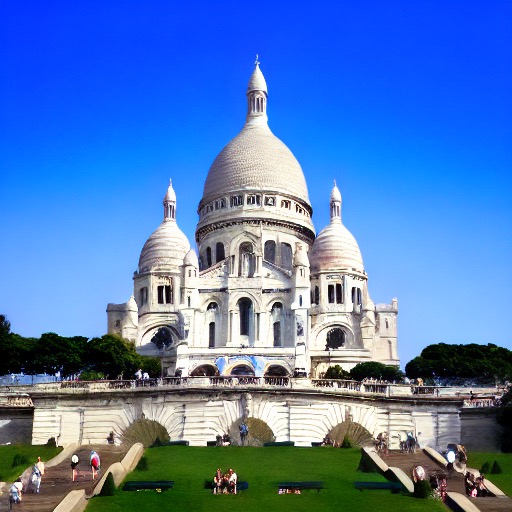} \\
        
        \end{tabular}
        
    \end{minipage}%
    \hspace{0.2cm}
    \begin{minipage}{0.32\textwidth}
        \centering
        \begin{tabular}{c c c c c}

        \includegraphics[width=0.25\textwidth]{images/struct_app.png} &
        \includegraphics[width=0.25\textwidth]{images/inputs/cars/vintage.jpg} &
        \includegraphics[width=0.25\textwidth]{images/inputs/cars/red_vintage.jpg} &
        \includegraphics[width=0.25\textwidth]{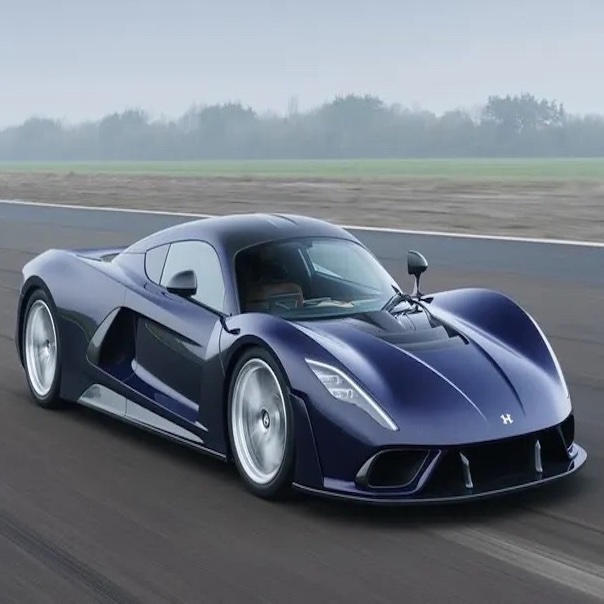} \\

        \includegraphics[width=0.25\textwidth]{images/inputs/cars/vintage.jpg} &
        \includegraphics[width=0.25\textwidth]{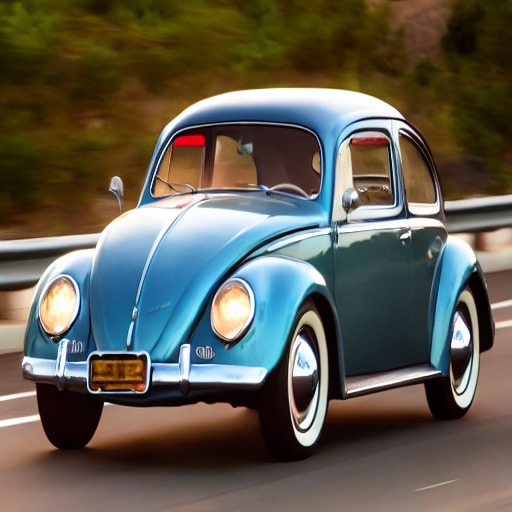} &
        \includegraphics[width=0.25\textwidth]{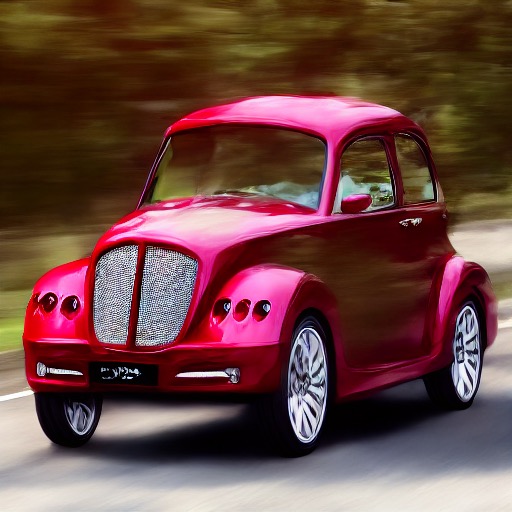} &
        \includegraphics[width=0.25\textwidth]{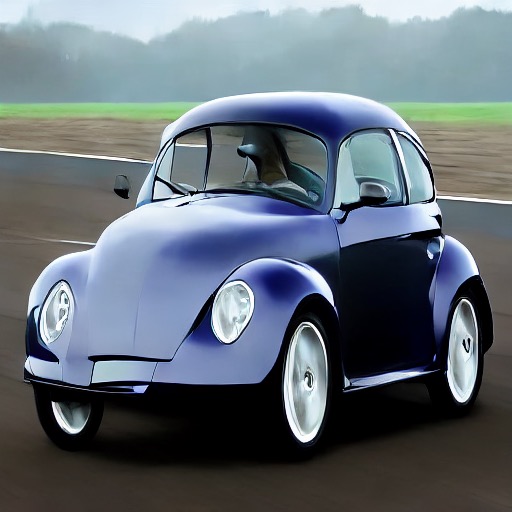} \\

        \includegraphics[width=0.25\textwidth]{images/inputs/cars/red_vintage.jpg} &
        \includegraphics[width=0.25\textwidth]{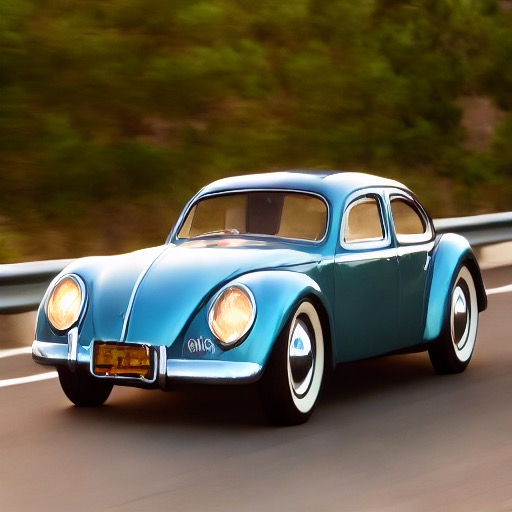} &
        \includegraphics[width=0.25\textwidth]{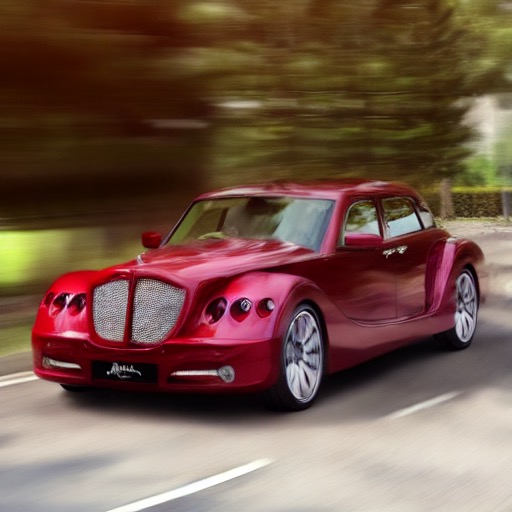} &
        \includegraphics[width=0.25\textwidth]{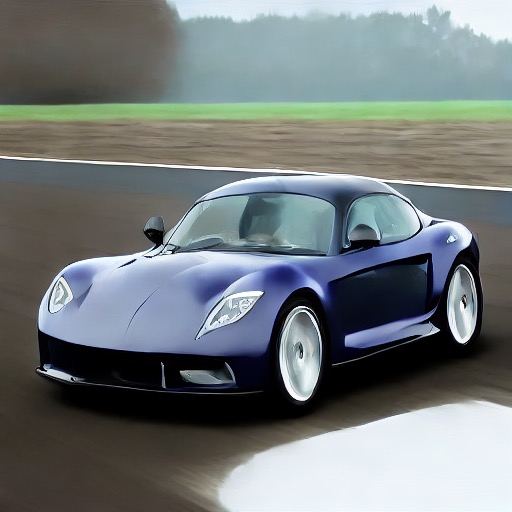} \\
        
        \includegraphics[width=0.25\textwidth]{images/inputs/cars/black_sports_flipped.jpeg} &
        \includegraphics[width=0.25\textwidth]{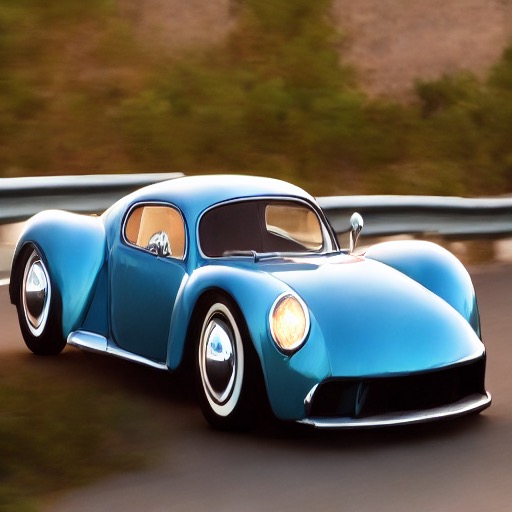} &
        \includegraphics[width=0.25\textwidth]{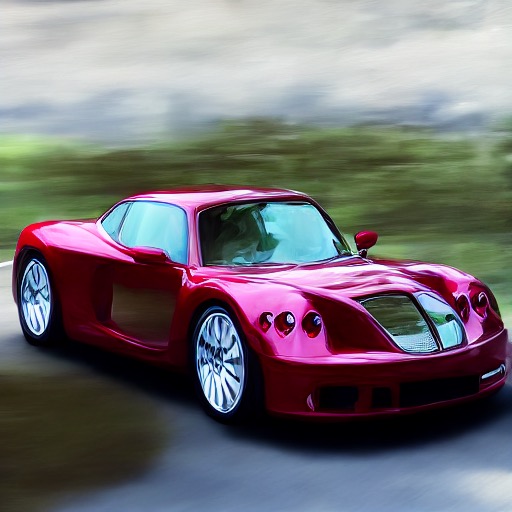} &
        \includegraphics[width=0.25\textwidth]{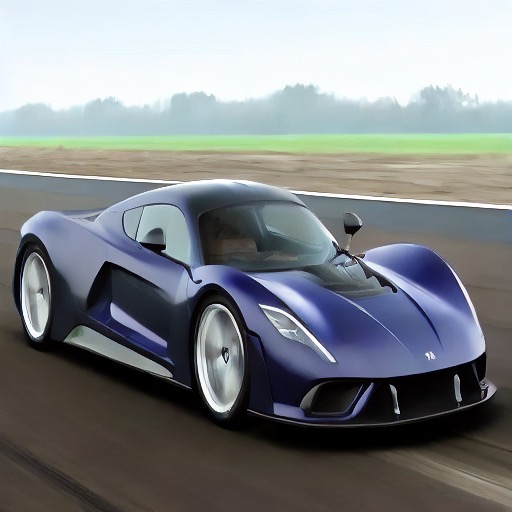} \\
        
        \end{tabular}
        
    \end{minipage}%
    \hspace{0.2cm}
    \begin{minipage}{0.32\textwidth}
        \centering
        \begin{tabular}{c c c c c}

        \includegraphics[width=0.25\textwidth]{images/struct_app.png} &
        \includegraphics[width=0.25\textwidth]{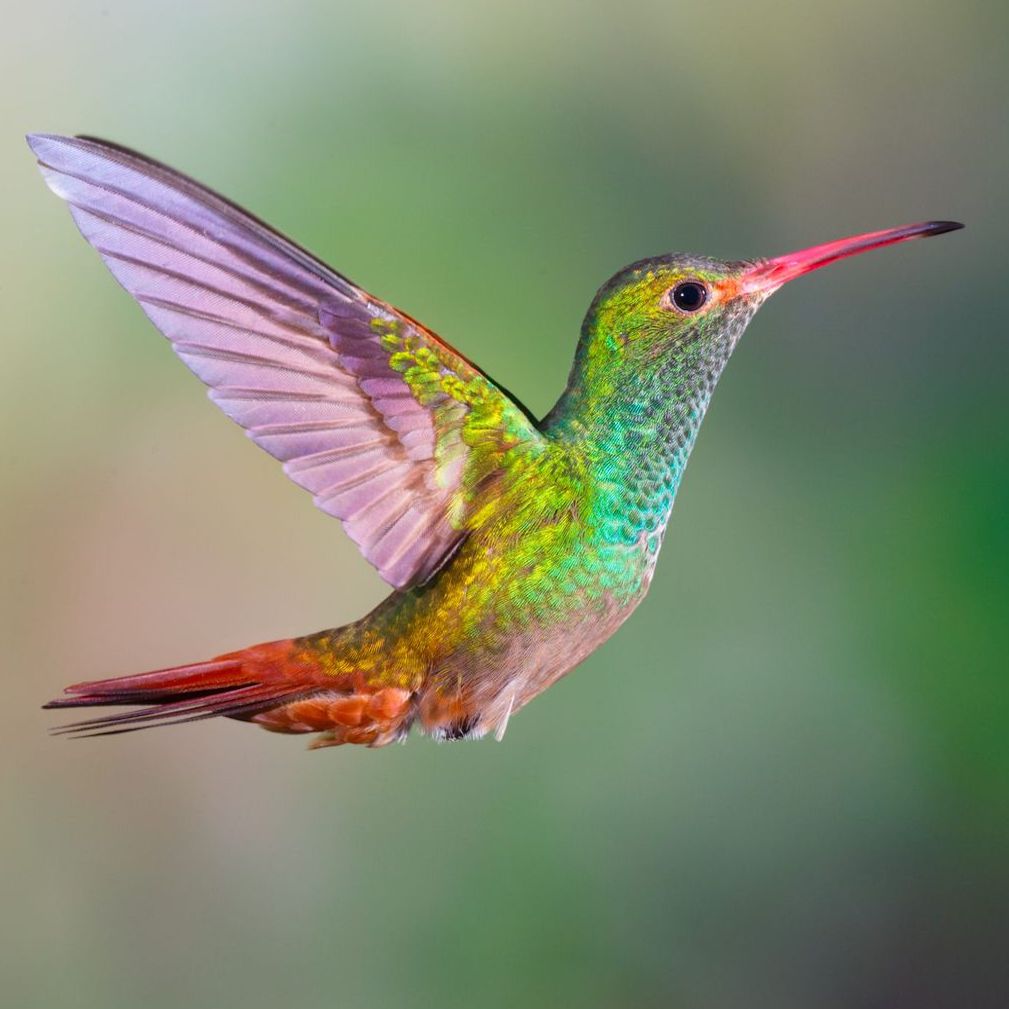} &
        \includegraphics[width=0.25\textwidth]{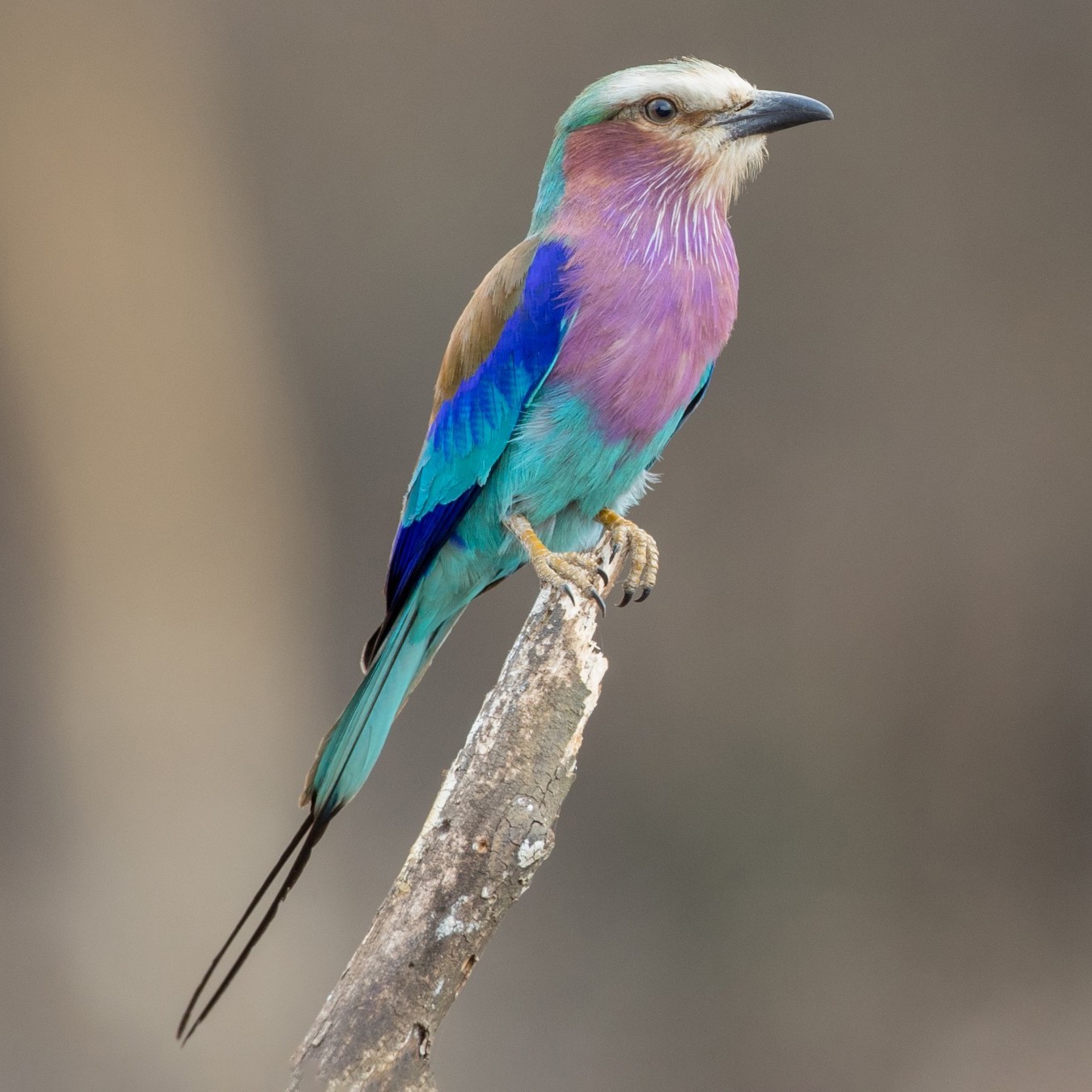} &
        \includegraphics[width=0.25\textwidth]{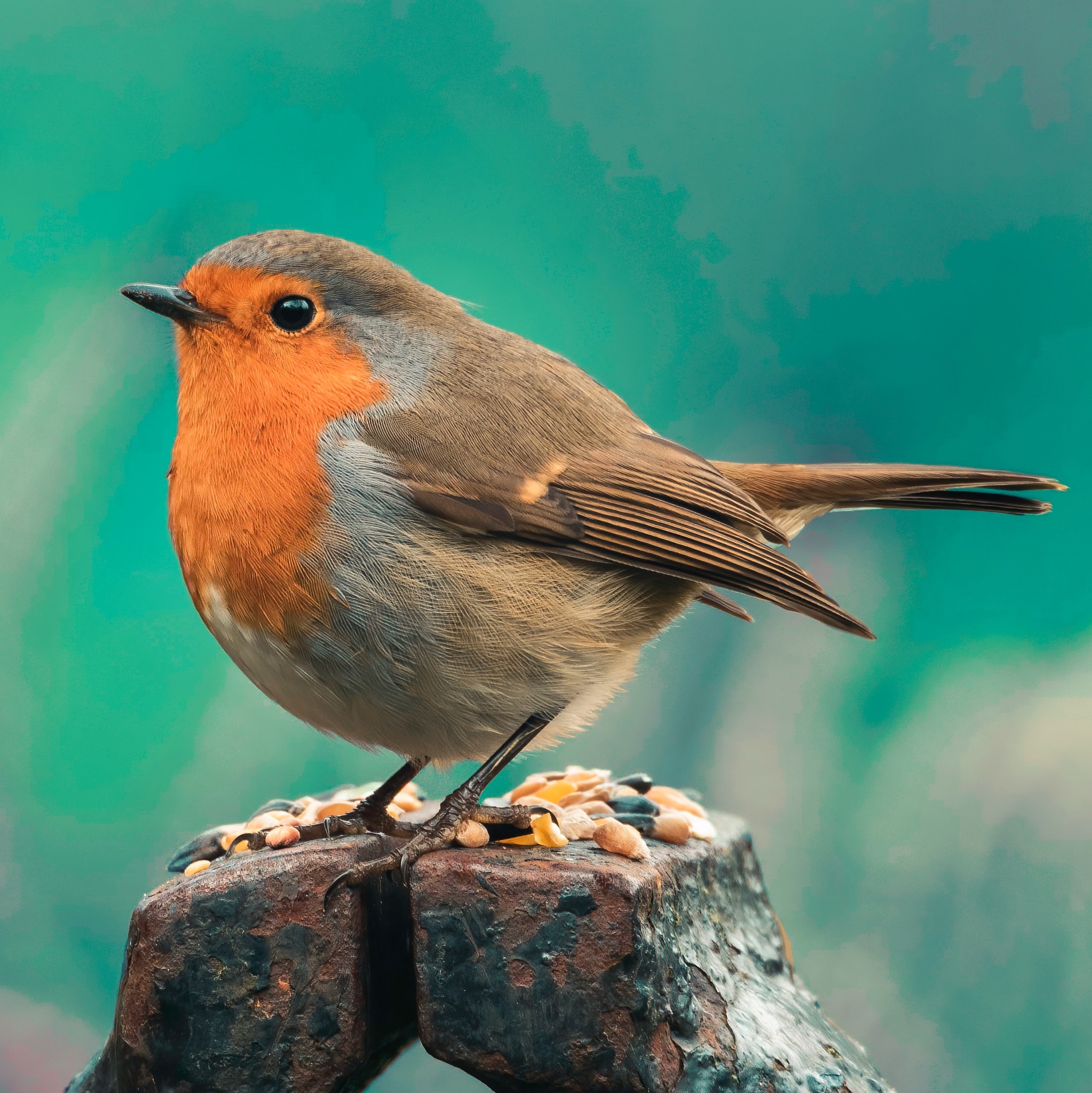} \\

        \includegraphics[width=0.25\textwidth]{images/inputs/birds/hummingbird.jpg} &
        \includegraphics[width=0.25\textwidth]{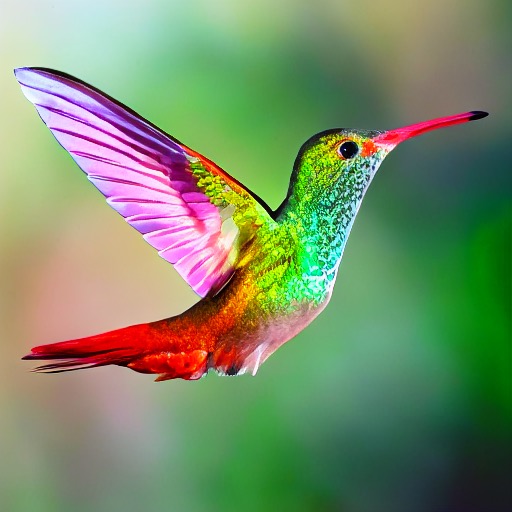} &
        \includegraphics[width=0.25\textwidth]{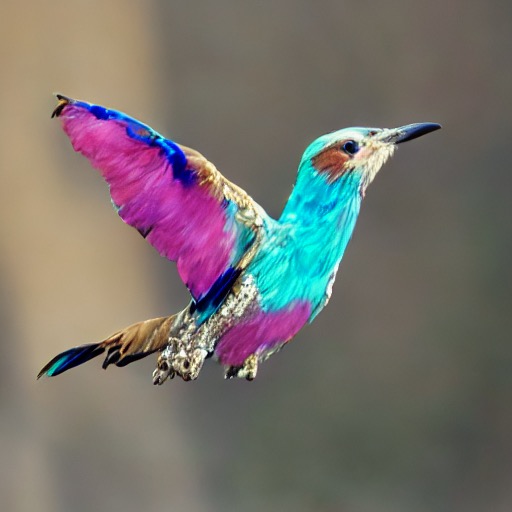} &
        \includegraphics[width=0.25\textwidth]{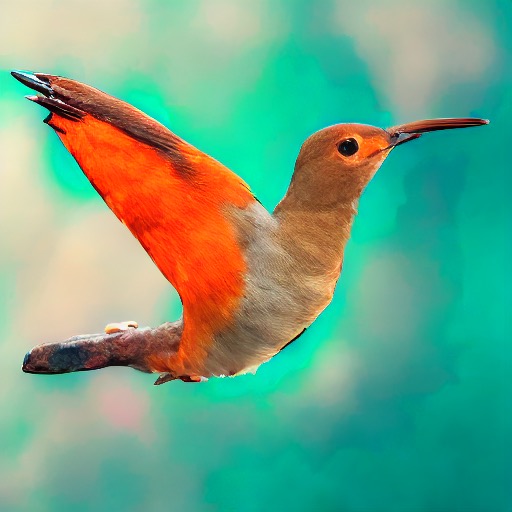} \\

        \includegraphics[width=0.25\textwidth]{images/inputs/birds/lilac_roller.jpg} &
        \includegraphics[width=0.25\textwidth]{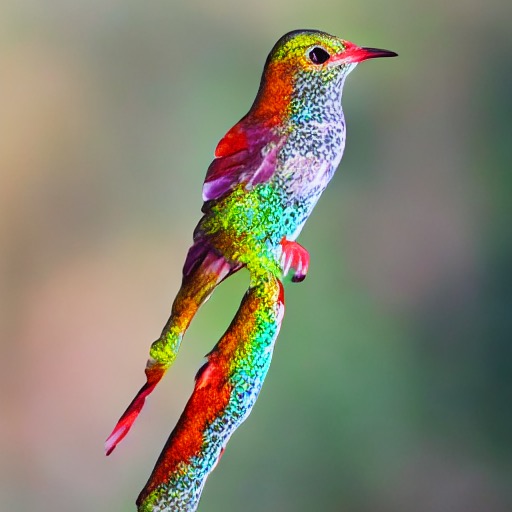} &
        \includegraphics[width=0.25\textwidth]{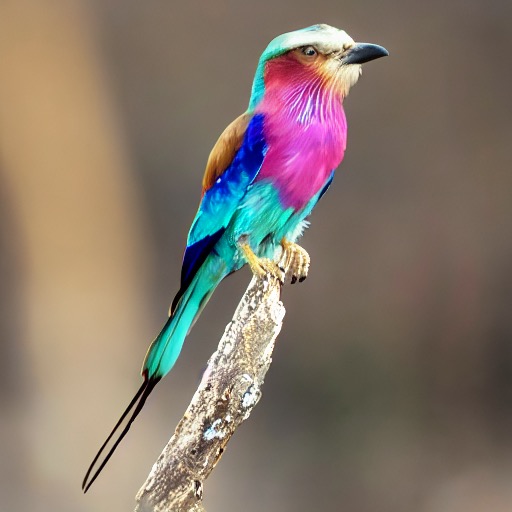} &
        \includegraphics[width=0.25\textwidth]{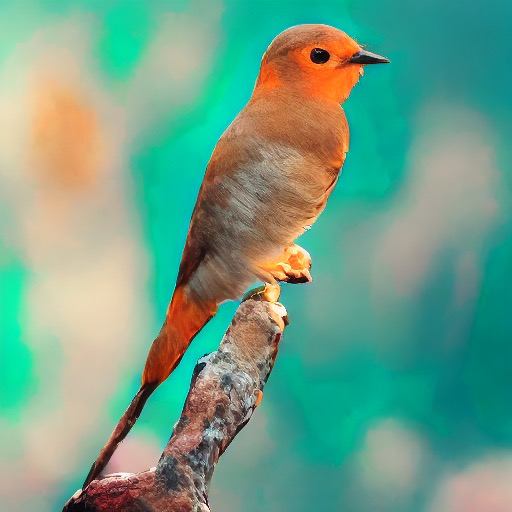} \\
        
        \includegraphics[width=0.25\textwidth]{images/inputs/birds/small_orange.jpg} &
        \includegraphics[width=0.25\textwidth]{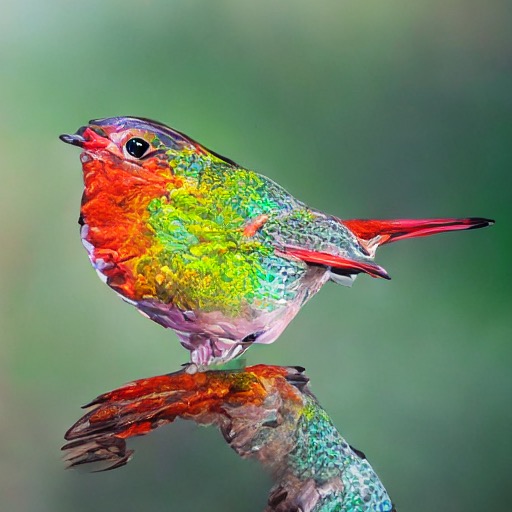} &
        \includegraphics[width=0.25\textwidth]{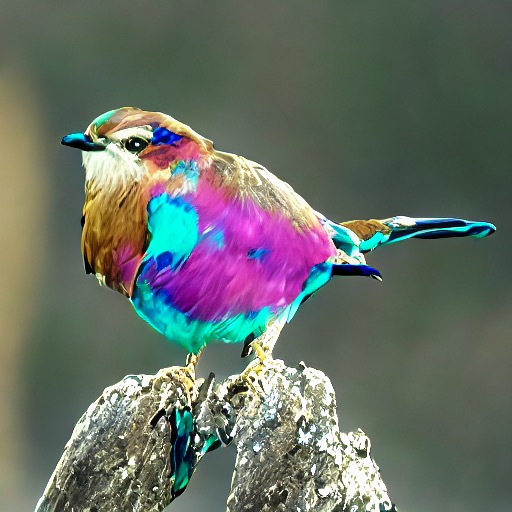} &
        \includegraphics[width=0.25\textwidth]{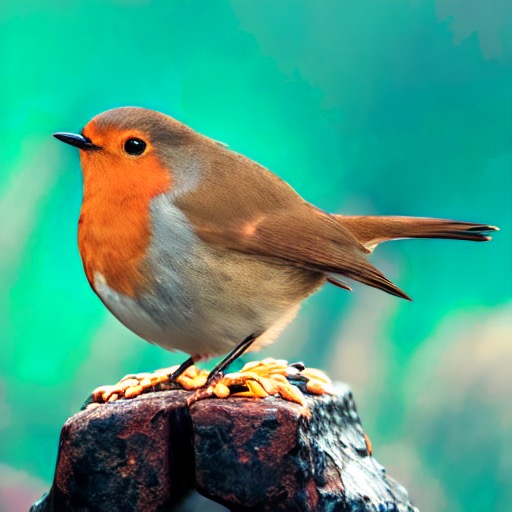} \\
        \end{tabular}
        
    \end{minipage}%

    }
    \vspace{-0.2cm}
    \caption{
    Semantic-based appearance transfer results obtained by our method. In each grid, the leftmost column displays the input structure images, while the topmost row presents the input appearance images. The remaining $3\times3$ grid showcases the results of the appearance transfer between each corresponding structure and appearance image.
    }
    \label{fig:our_results_grid}
    \vspace{-0.1cm}
\end{figure*}

\vspace{-0.15cm}
\paragraph{Appearance Guidance}
Next, we adapt the concept of classifier-free guidance~\cite{ho2022classifier}, which has been shown to improve the overall quality of generated images, to the realm of appearance transfer. At each denoising step $t$ we perform two forward passes through the denoising network: (i) $\epsilon^{\times} = \epsilon^\times_\theta(z_t^{out})$ using our cross-image attention layer, and (ii) $\epsilon^{\text{self}} = \epsilon^{\text{self}}_\theta(z_t^{out})$ using the original self-attention layer of the network. Given the two noise predictions, we then define the final predicted noise $\epsilon^t$ as:
\begin{equation}
\epsilon^t = \epsilon^{\text{self}} + \alpha \left ( \epsilon^{\times} - \epsilon^{\text{self}} \right ),
\end{equation}
where $\alpha$ is the guidance scale. The next latent code $z_{t-1}^{out}$ is then sampled using the modified noise $\epsilon^t$. 

Intuitively, this guidance mechanism shifts the noisy latent code towards denser regions of the distribution associated with the target appearance while moving away from the original appearance. By reaching denser regions of the distribution, we obtain more plausible images, resulting in fewer artifacts.

\vspace{-0.1cm}
\paragraph{AdaIN}
In addition to the artifacts handled by the previously described mechanisms, we observe a shift in the color distribution between the output image and the input appearance image. To address this, we utilize the AdaIN operation~\cite{huang2017adain}, originally introduced for style transfer and known for effectively matching feature statistics between latent representations. We find that applying AdaIN on $z_{t}^{out}$ with respect to $z_{t}^{app}$ assists in gradually aligning the color distribution of the output and appearance images. 
Specifically, we update
\begin{equation}
    z_t^{out}  \gets \text{AdaIN}(z_t^{out}, z_t^{app}),
\end{equation}
where the statistics of $z_t^{out}$ are adjusted to match those of $z_t^{app}$, assisting in aligning their color distributions.

However, we notice that the statistics computed by the AdaIN operation are sensitive to the size of the objects. As a result, AdaIN may not be effective when the objects depicted in the images significantly vary in size. To address this, we apply a mask over the latents $z_t^{out}$ and $z_t^{app}$ and restrict the AdaIN operation to compute the feature statistics only on a foreground mask containing the object. To create the object masks, we employ the unsupervised self-segmentation technique introduced in Patashnik~\etal~\cite{patashnik2023localizing}.

\section{Experiments} \label{sec:exp}

In the following section, we demonstrate the effectiveness of our cross-image attention technique for the task of appearance transfer.

\subsection{Evaluations and Comparisons}

\begin{figure}
    \centering
    \setlength{\tabcolsep}{0.3pt}
    \renewcommand{\arraystretch}{0.3}
    \addtolength{\belowcaptionskip}{-5pt}
    {\small
    \begin{tabular}{c c c c c c}

        \raisebox{0.05in}{\rotatebox{90}{ Structure }} &
        \includegraphics[width=0.09\textwidth]{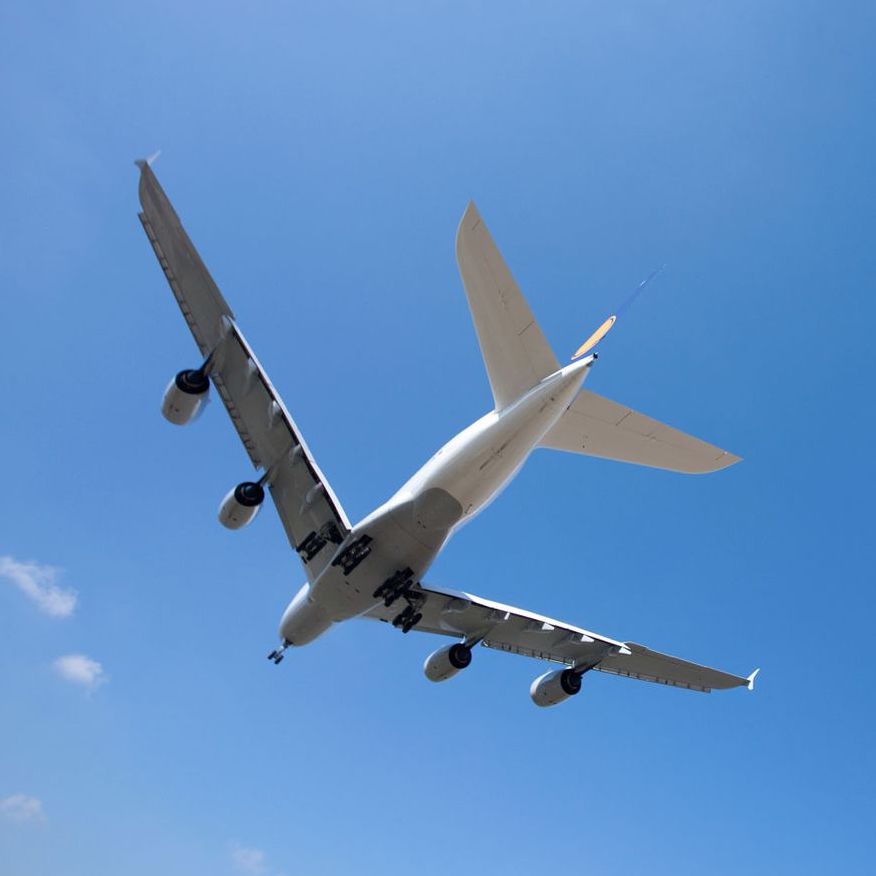} &
        \includegraphics[width=0.09\textwidth]{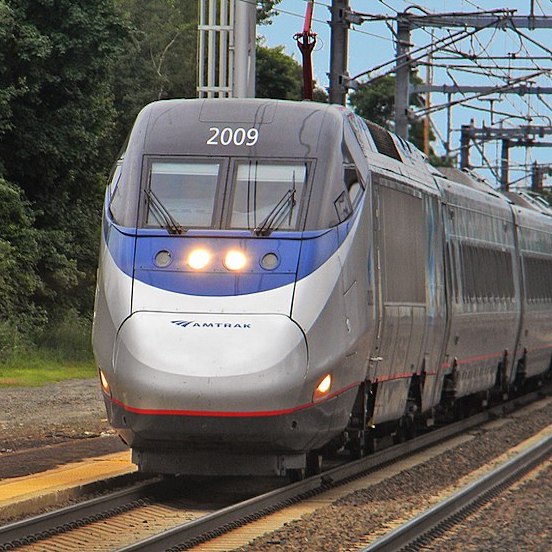} &
        \includegraphics[width=0.09\textwidth]{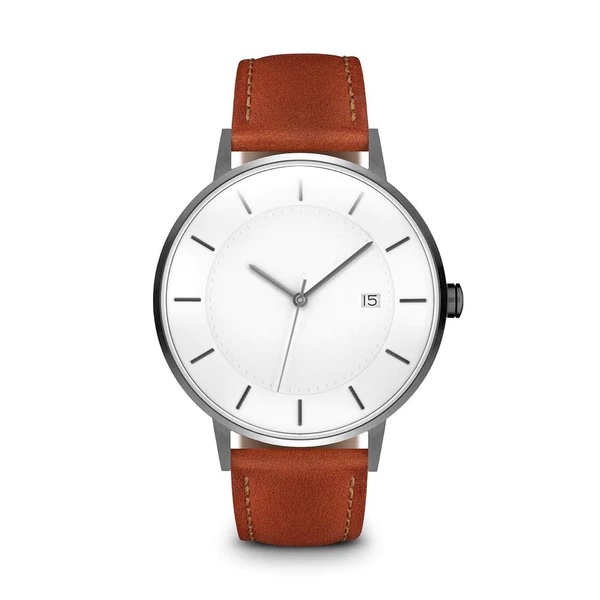} &
        \includegraphics[width=0.09\textwidth]{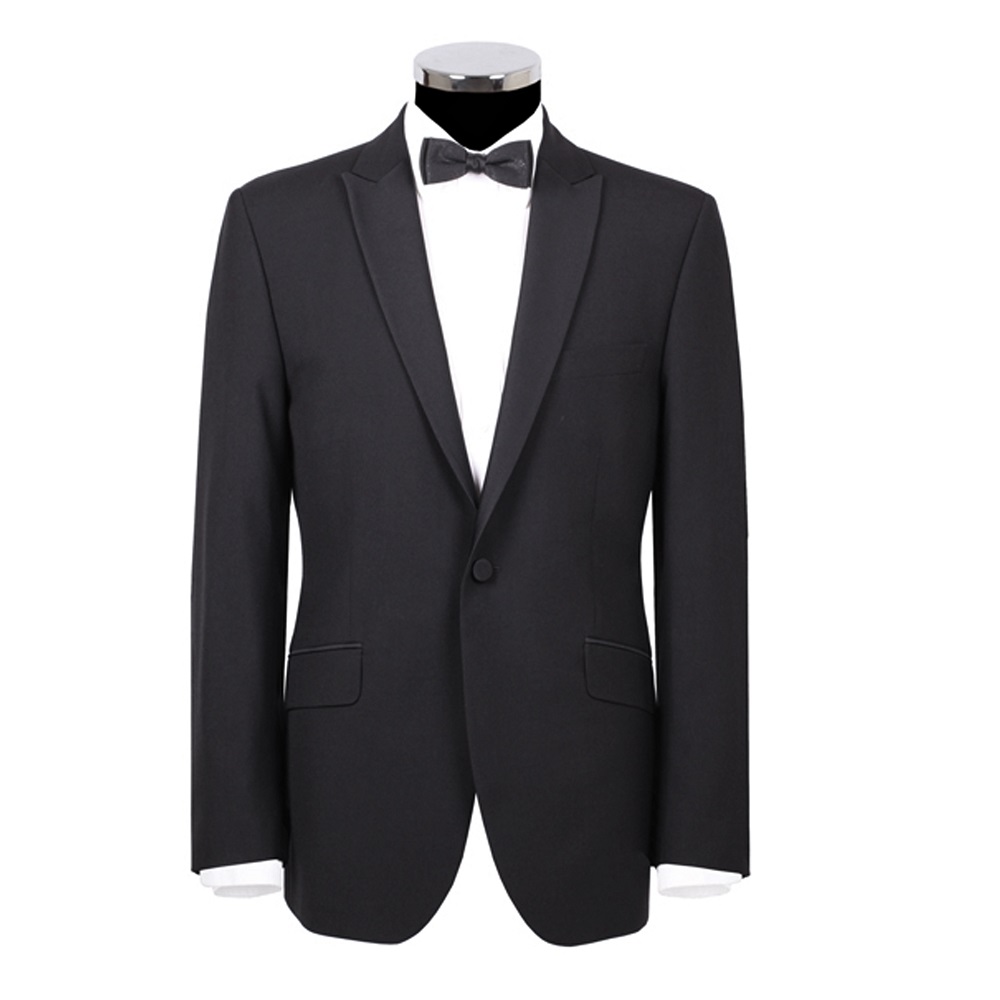} &
        \includegraphics[width=0.09\textwidth]{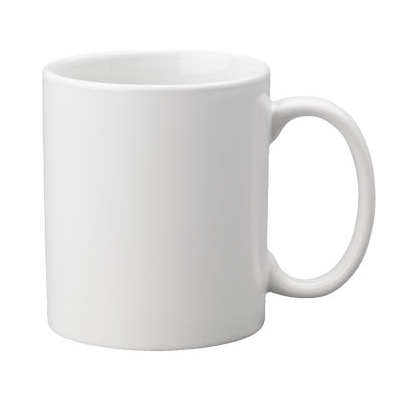} \\

        \raisebox{-0.025in}{\rotatebox{90}{ Appearance }} &
        \includegraphics[width=0.09\textwidth]{images/inputs/birds/hummingbird.jpg} &
        \includegraphics[width=0.09\textwidth]{images/inputs/cars/vintage.jpg} &
        \includegraphics[width=0.09\textwidth]{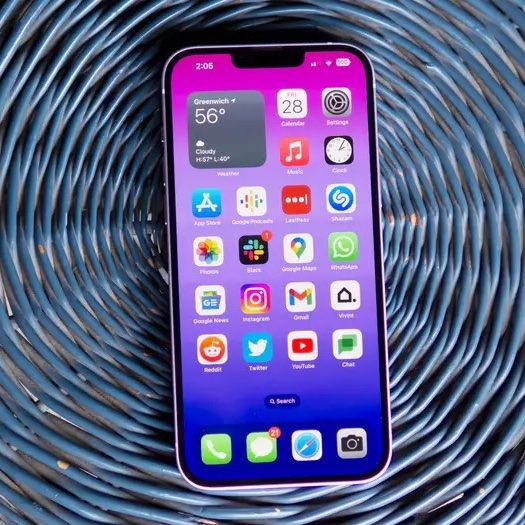} &
        \includegraphics[width=0.09\textwidth]{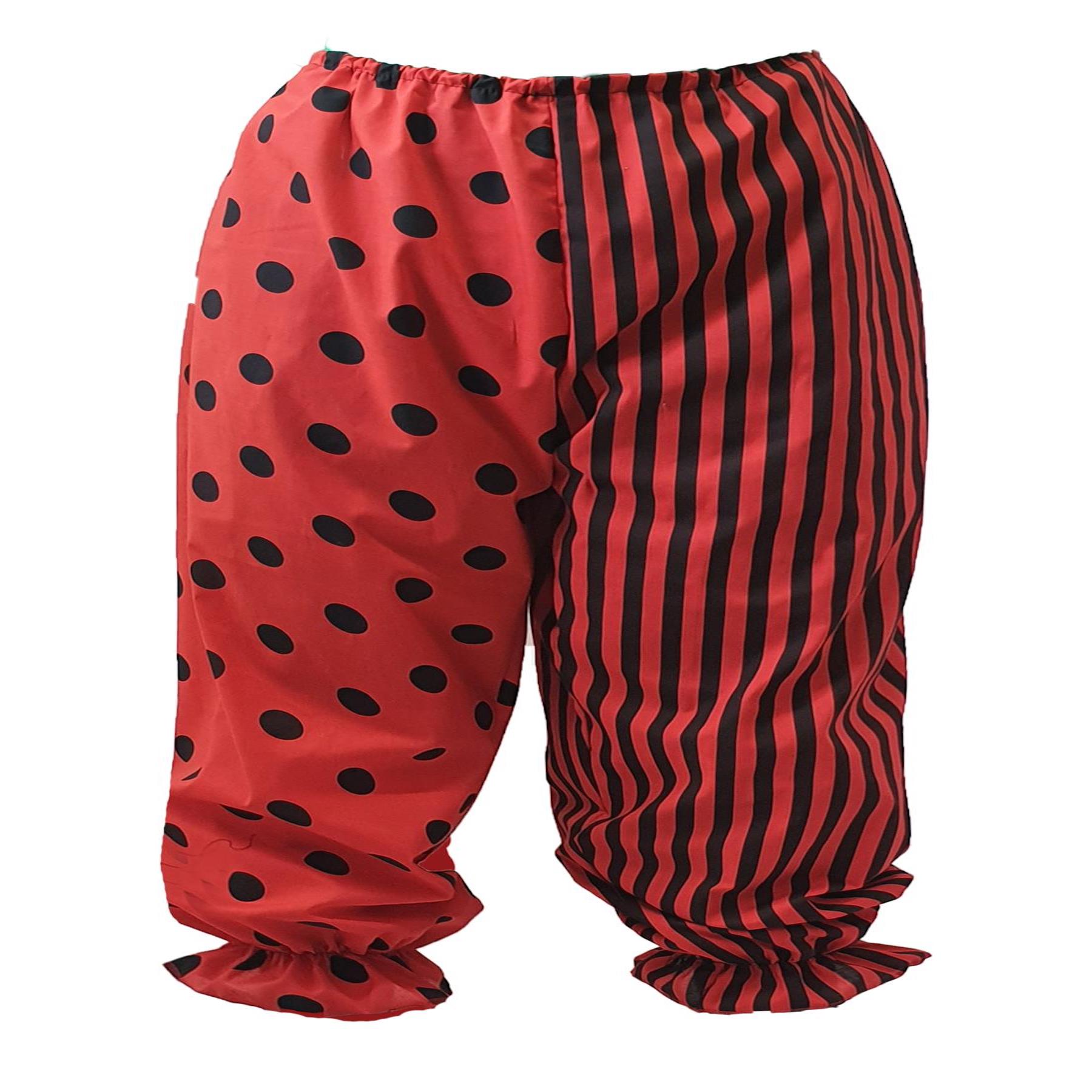} &
        \includegraphics[width=0.09\textwidth]{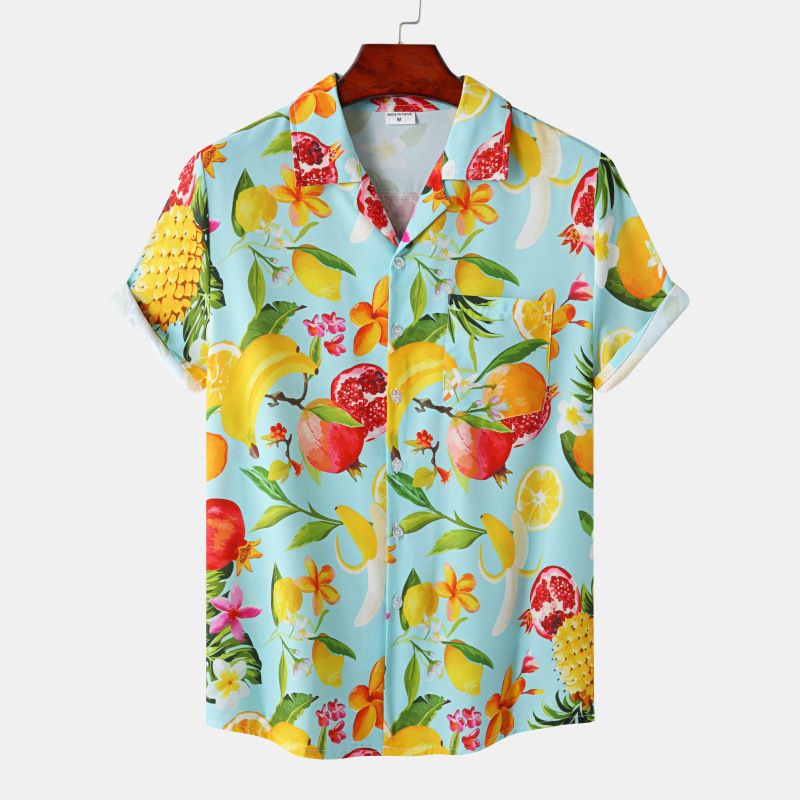} \\

        \raisebox{0.1in}{\rotatebox{90}{ Output }} &
        \includegraphics[width=0.09\textwidth]{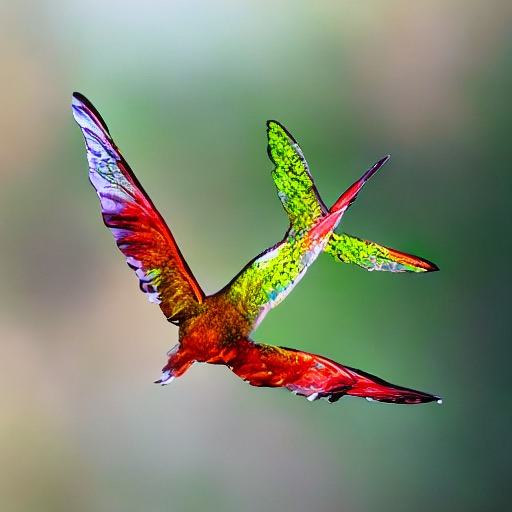} &
        \includegraphics[width=0.09\textwidth]{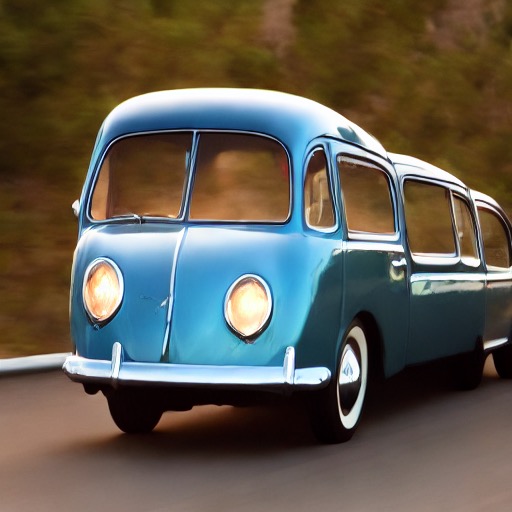} &
        \includegraphics[width=0.09\textwidth]{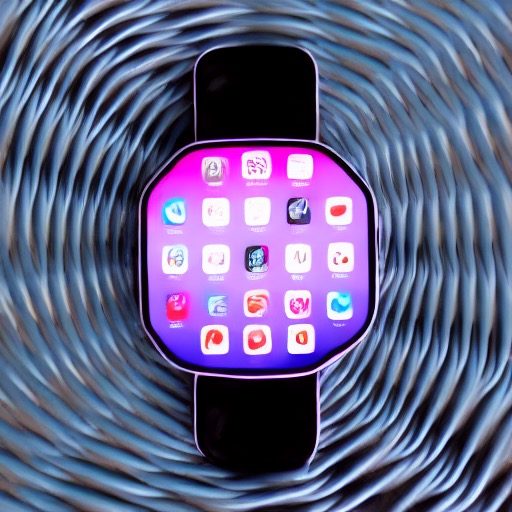} &
        \includegraphics[width=0.09\textwidth]{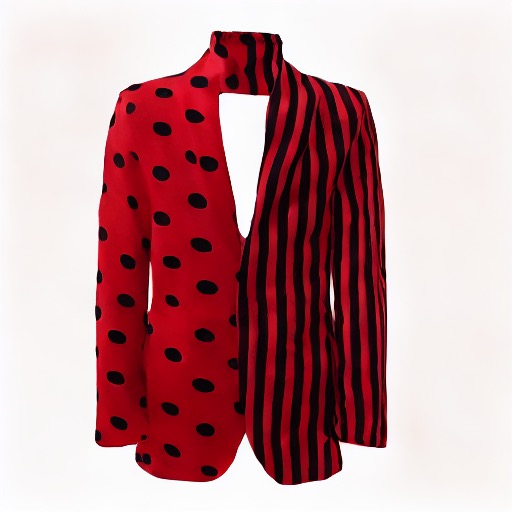} &
        \includegraphics[width=0.09\textwidth]{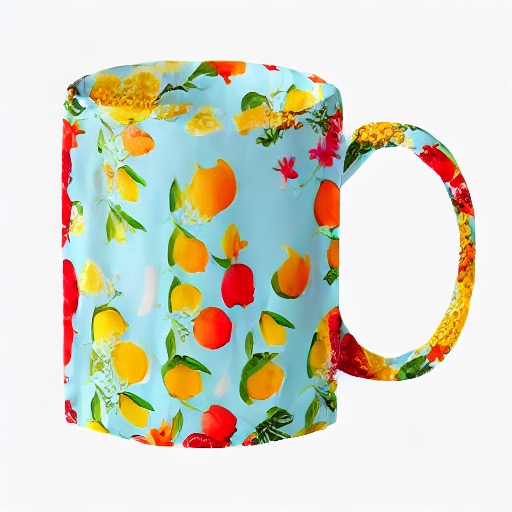} \\
        
    \end{tabular}

    }
    \vspace{-0.2cm}
    \caption{Cross-domain appearance transfer. Our approach can transfer appearance between cross-domain objects. This transfer is possible even in a zero-shot setting thanks to the strong correspondences already captured by the diffusion model itself.}
    \label{fig:cross_domain_results}
\end{figure}

\paragraph{Evaluation Setup}
We evaluate the performance of our cross-image attention mechanism in comparison to state-of-the-art semantic-based appearance transfer methods. These works range from methods that require training a generator for each target domain (Swapping Autoencoder~\cite{park2020swapping}) or each input image pair (SpliceVIT~\cite{tumanyan2022splicing}) to those relying on external models to guide an inference-time optimization process (e.g., DiffuseIT~\cite{kwon2022diffusion}). Results for all methods are produced using their official implementations and default parameters. Additional details can be found in~\Cref{sec:additional_details}.

\begin{figure}
    \centering
    \setlength{\tabcolsep}{0.3pt}
    \renewcommand{\arraystretch}{0.3}
    \addtolength{\belowcaptionskip}{-5pt}
    {\small
    \begin{tabular}{c c c c c c}

        &
        \includegraphics[width=0.09\textwidth]{images/struct_app.png} &
        \includegraphics[width=0.09\textwidth]{images/inputs/buildings/saint_basil.jpg} &
        \includegraphics[width=0.09\textwidth]{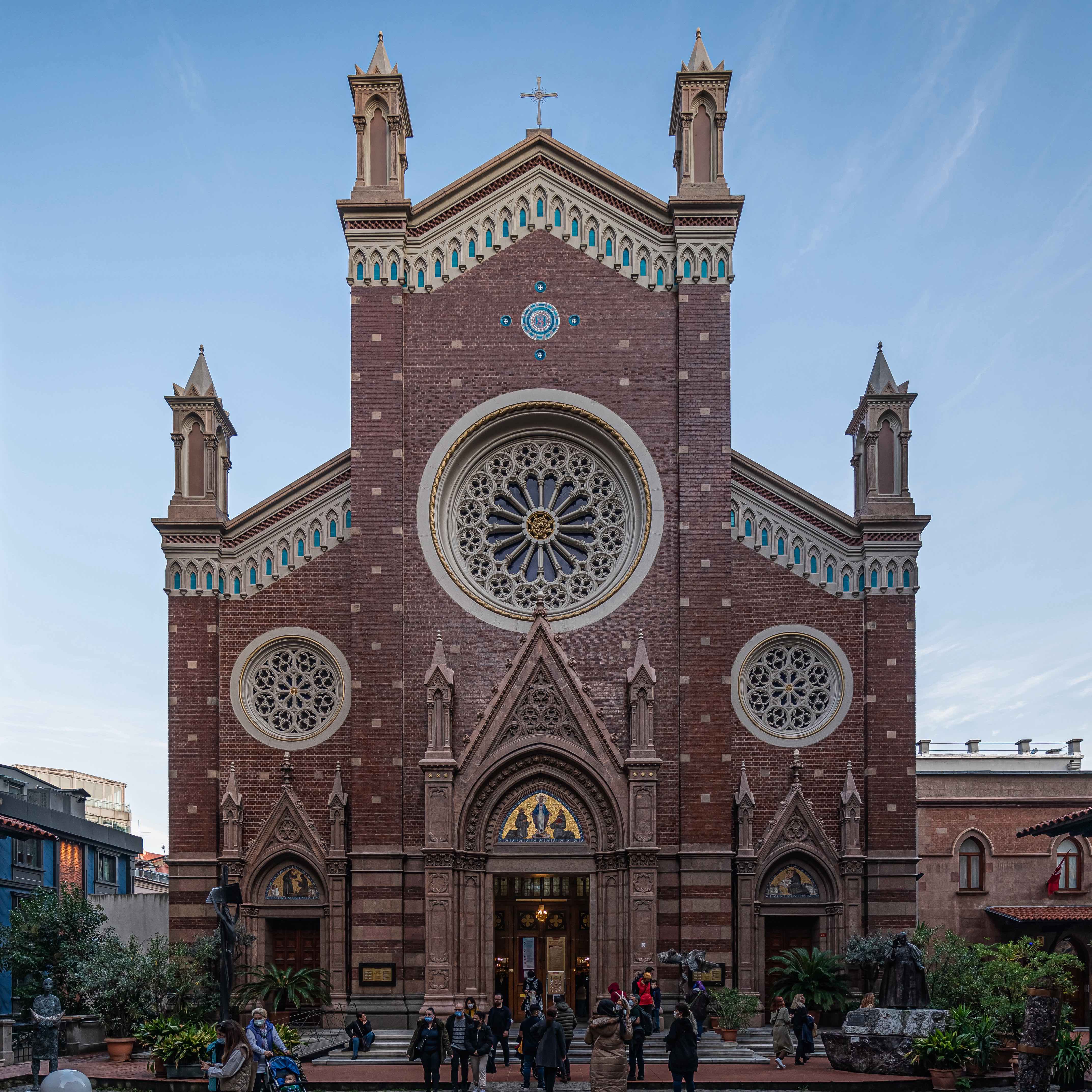} &
        \includegraphics[width=0.09\textwidth]{images/inputs/buildings/chile-church.jpg} &
        \includegraphics[width=0.09\textwidth]{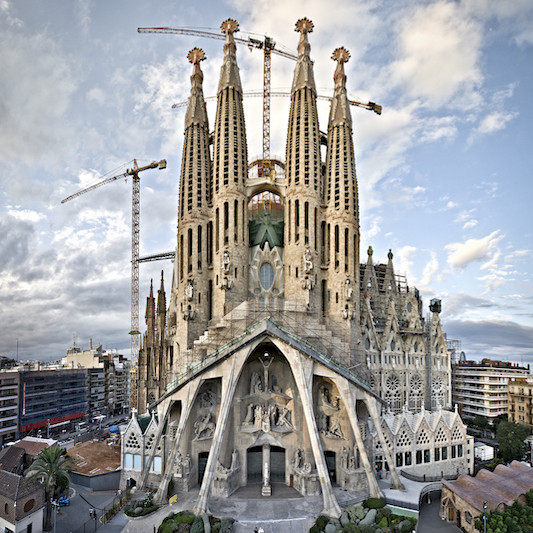} \\

        \raisebox{0.15in}{\rotatebox{90}{ Ours }} &
        \includegraphics[width=0.09\textwidth]{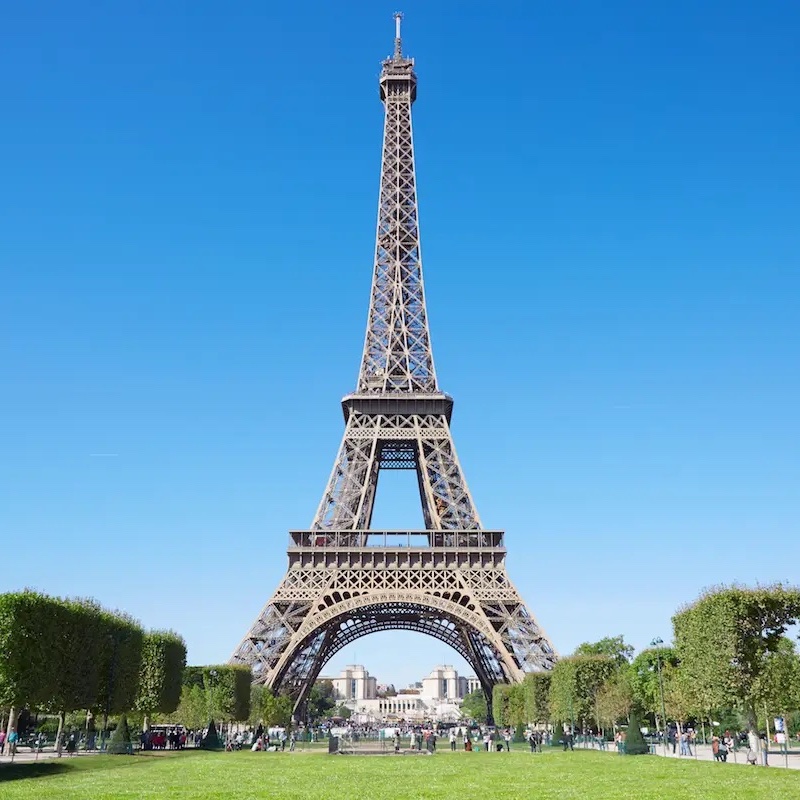} &
        \includegraphics[width=0.09\textwidth]{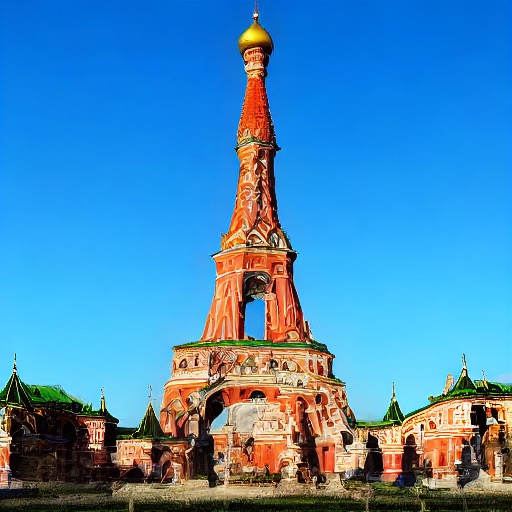} &
        \includegraphics[width=0.09\textwidth]{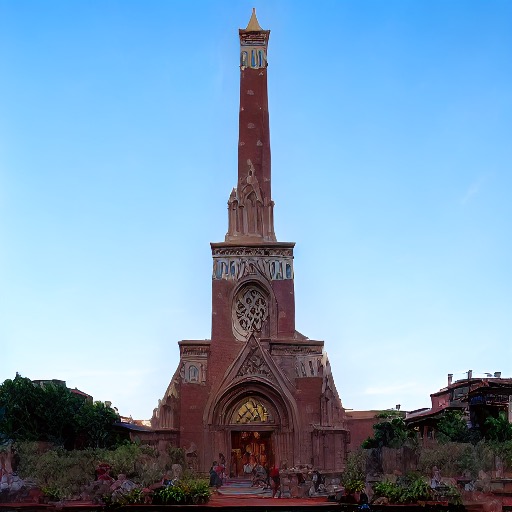} &
        \includegraphics[width=0.09\textwidth]{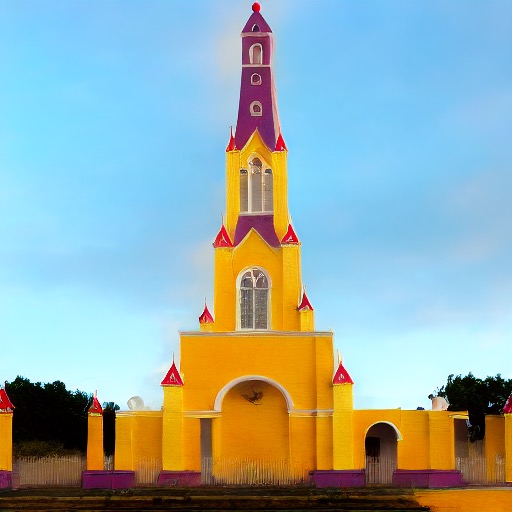} &
        \includegraphics[width=0.09\textwidth]{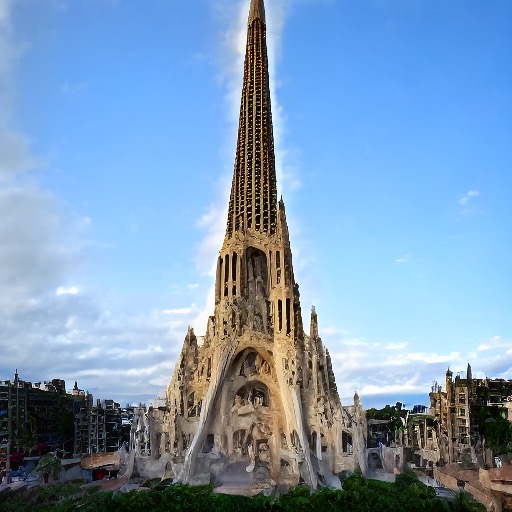} \\

        \vspace{0.1cm}

        \raisebox{0.025in}{\rotatebox{90}{ Swap. AE }} &
        \includegraphics[width=0.09\textwidth]{images/inputs/buildings/eiffel_tower.jpg} &
        \includegraphics[width=0.09\textwidth]{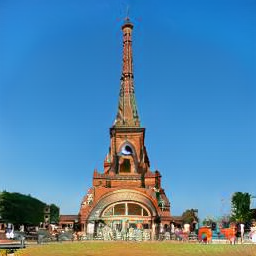} &
        \includegraphics[width=0.09\textwidth]{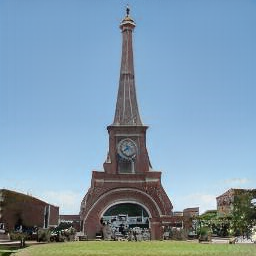} &
        \includegraphics[width=0.09\textwidth]{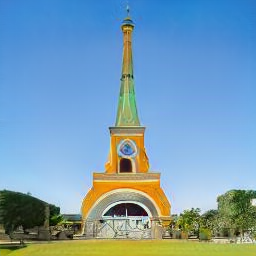} &
        \includegraphics[width=0.09\textwidth]{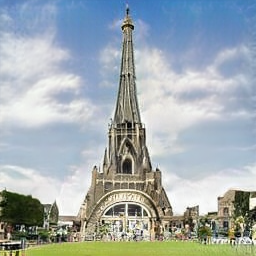} \\

        &
        \includegraphics[width=0.09\textwidth]{images/struct_app.png} &
        \includegraphics[width=0.09\textwidth]{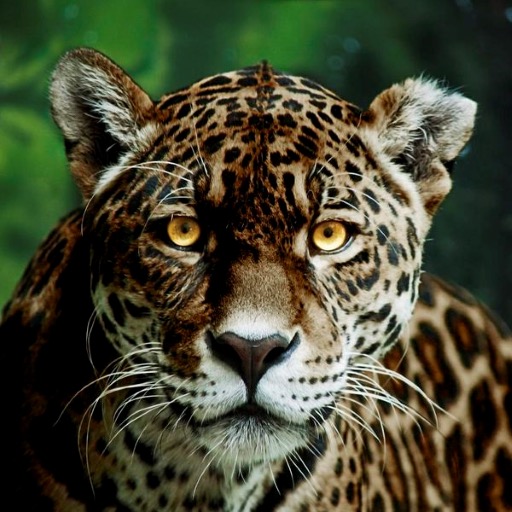} &
        \includegraphics[width=0.09\textwidth]{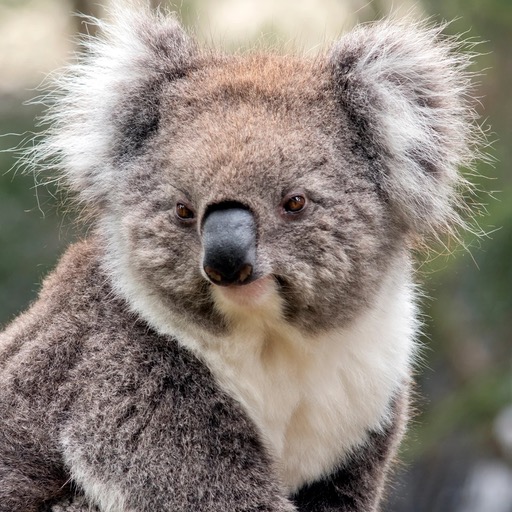} &
        \includegraphics[width=0.09\textwidth]{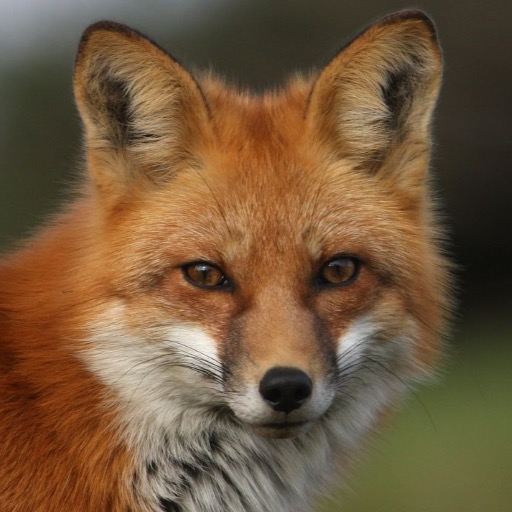} &
        \includegraphics[width=0.09\textwidth]{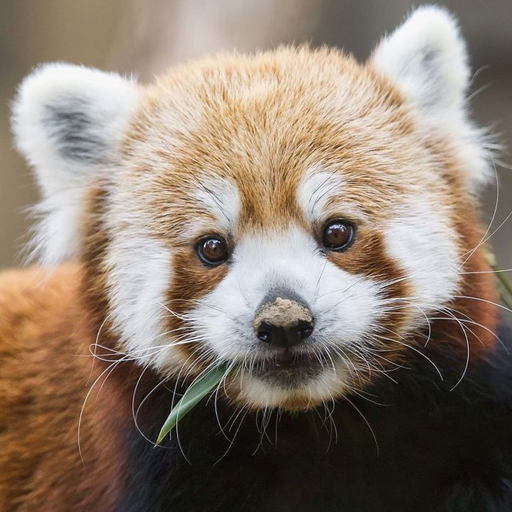} \\
    
        \raisebox{0.15in}{\rotatebox{90}{ Ours }} &
        \includegraphics[width=0.09\textwidth]{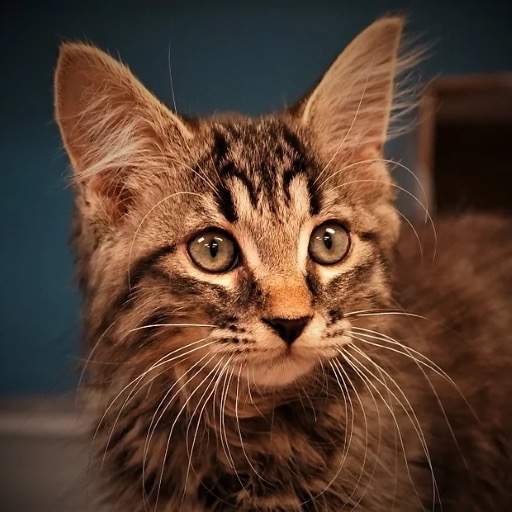} &
        \includegraphics[width=0.09\textwidth]{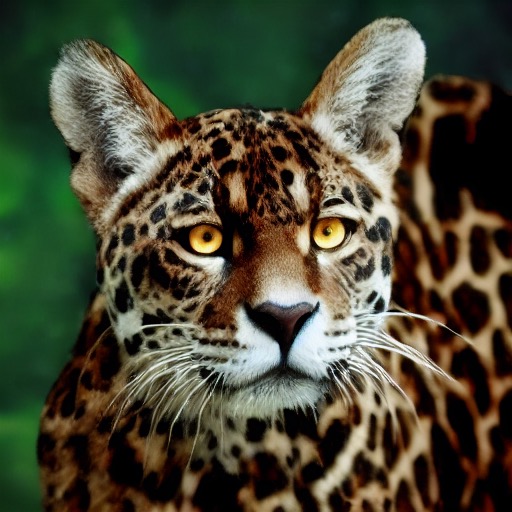} &
        \includegraphics[width=0.09\textwidth]{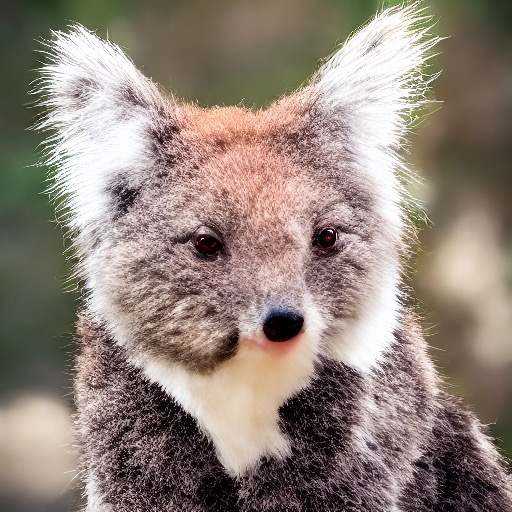} &
        \includegraphics[width=0.09\textwidth]{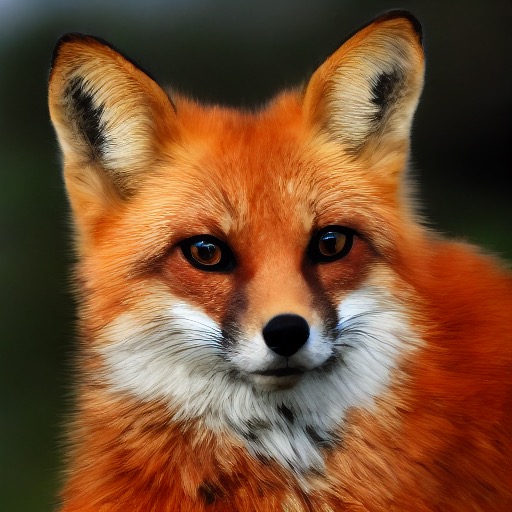} &
        \includegraphics[width=0.09\textwidth]{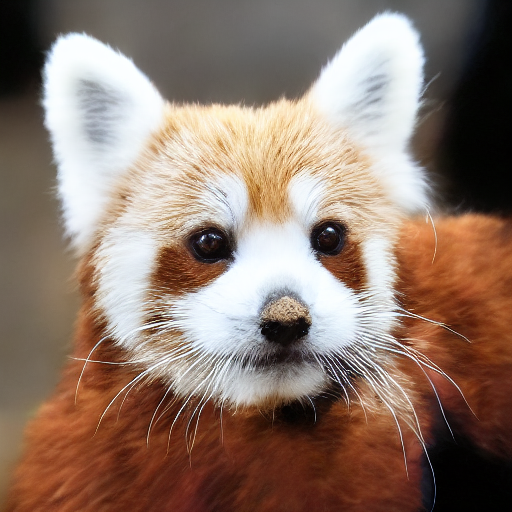} \\

        \raisebox{0.025in}{\rotatebox{90}{ Swap. AE }} &
        \includegraphics[width=0.09\textwidth]{images/compare_swapping_ae/animal_faces/origin/cat.jpg} &
        \includegraphics[width=0.09\textwidth]{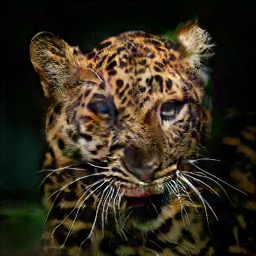} &
        \includegraphics[width=0.09\textwidth]{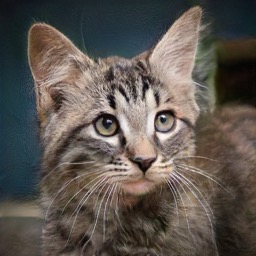} &
        \includegraphics[width=0.09\textwidth]{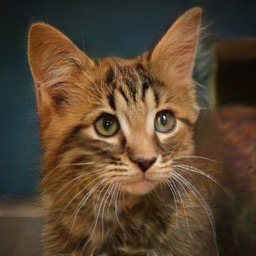} &
        \includegraphics[width=0.09\textwidth]{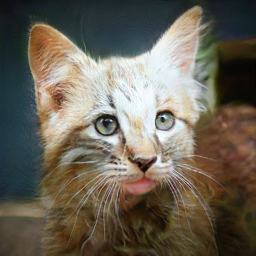} \\

    \end{tabular}

    }
    \vspace{-0.25cm}
    \caption{
    Comparison to Swapping Autoencoder (SA)~\cite{park2020swapping}. We provide a comparison to SA using their pretrained churches and animal faces models. For each input structure image (shown to the left), we show transfer results obtained through four different appearance images (shown in the top row). 
    }
    \vspace{-0.1cm}
    \label{fig:compare_swapping_ae}
\end{figure}

\vspace{-0.15cm}
\paragraph{Qualitative Evaluation}
In~\Cref{fig:our_results_grid} we illustrate appearance transfer results obtained by our method across three object domains. As can be seen, our method is effective in transferring the visual appearance in a semantically faithful manner. 
This holds even for challenging image pairs that may contain variations in object shape. %
For example, in the leftmost grid, our method successfully transfers prominent features between the buildings such as the dome tops in the first building or the columns of the Taj Mahal in the second row. Moreover, in the middle example, we successfully transfer key visual features between the cars such as the headlights of the blue beetle (leftmost column) or the front grill of the red car (middle column). 

Next, in~\Cref{fig:cross_domain_results} we present more challenging cross-domain results where the structure and appearance images come from different object categories. Our method can still generate semantically plausible images, such as between the airplane and the hummingbird in the leftmost column. This transfer also works surprisingly well between objects with less shared semantics such as a watch and a phone or a shirt and a coffee mug. We do observe, however, that transfer between cross-domain images is generally more challenging due to the less accurate correspondences typically established by the model. For instance, in the fourth column, the tie of the tuxedo is not transferred to the output image.

\begin{figure}
    \centering
    \setlength{\tabcolsep}{0.4pt}
    \addtolength{\belowcaptionskip}{-10pt}

    {\small

    \begin{tabular}{c c@{\hspace{0.2cm}} c c c}

        \includegraphics[width=0.0915\textwidth]{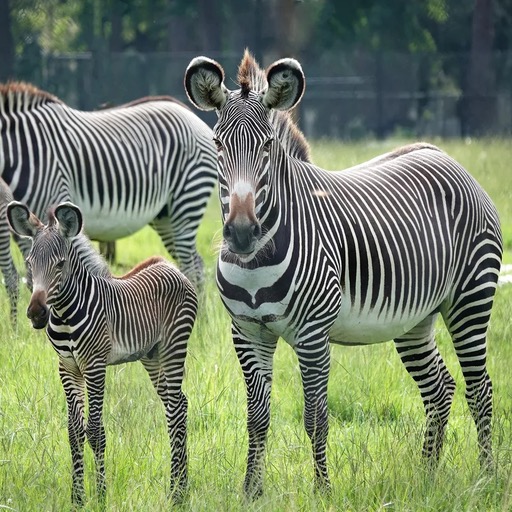} &
        \includegraphics[width=0.0915\textwidth]{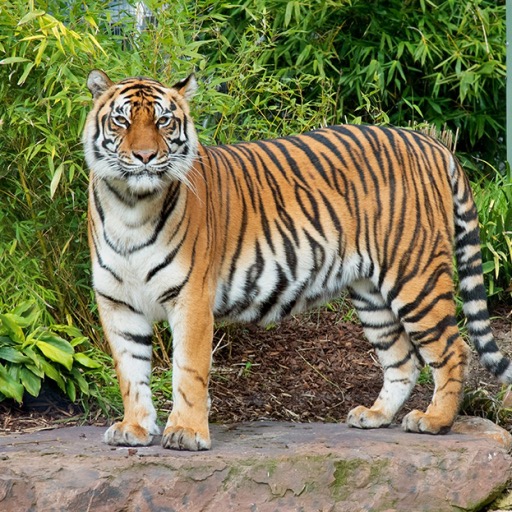} &
        \includegraphics[width=0.0915\textwidth]{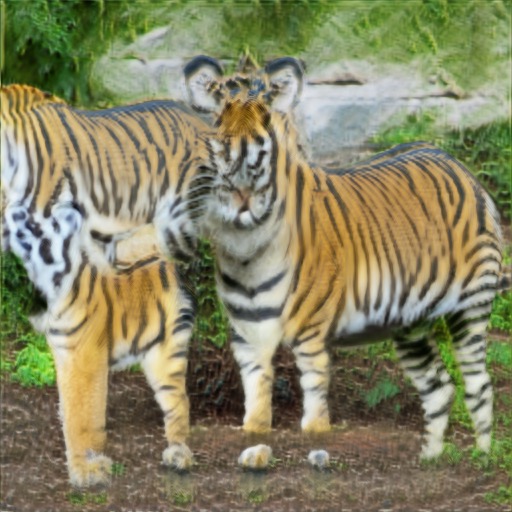} &
        \includegraphics[width=0.0915\textwidth]{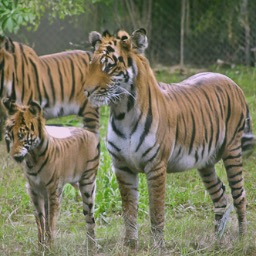} &
        \includegraphics[width=0.0915\textwidth]{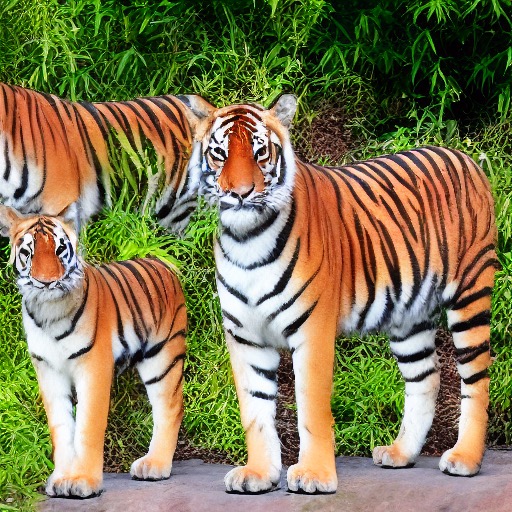} \\

        \includegraphics[width=0.0915\textwidth]{images/inputs/cars/red_vintage.jpg} &
        \includegraphics[width=0.0915\textwidth]{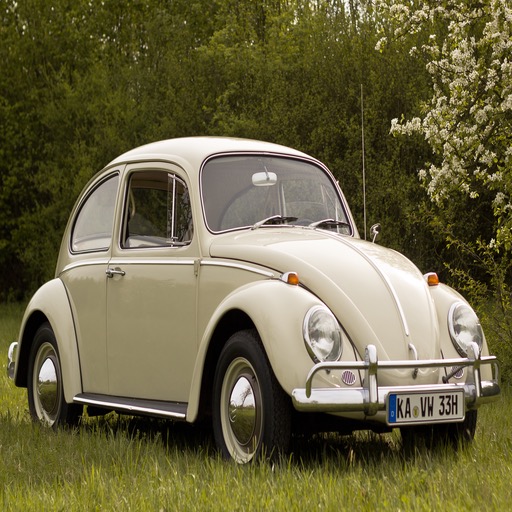} &
        \includegraphics[width=0.0915\textwidth]{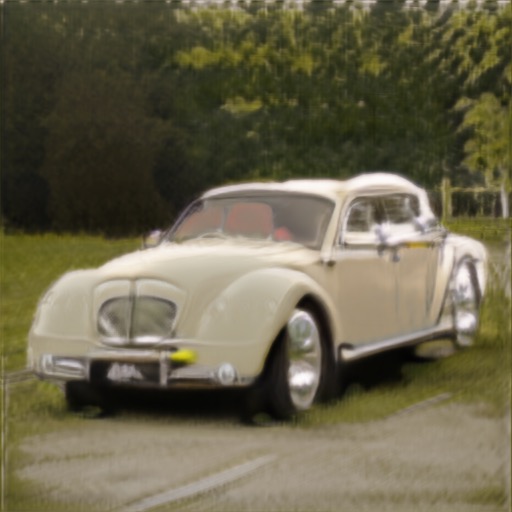} &
        \includegraphics[width=0.0915\textwidth]{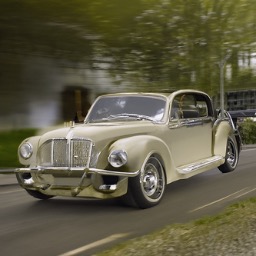} &
        \includegraphics[width=0.0915\textwidth]{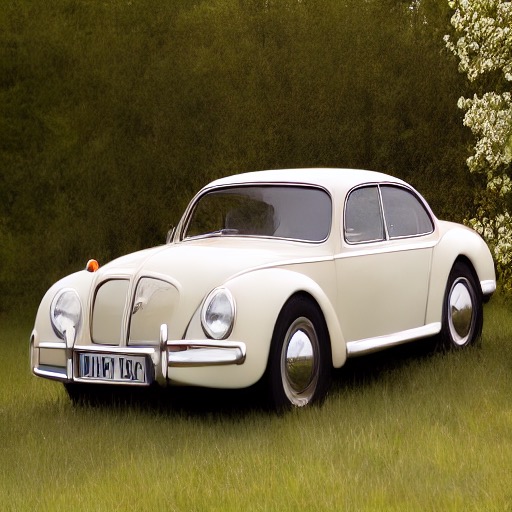} \\

        \includegraphics[width=0.0915\textwidth]{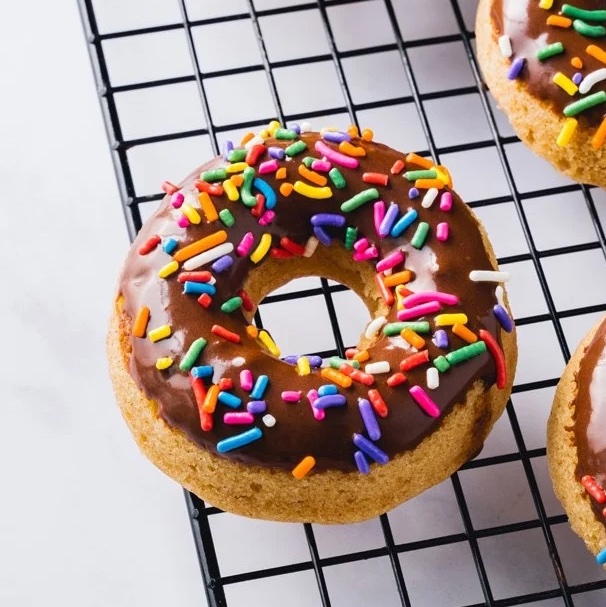} &
        \includegraphics[width=0.0915\textwidth]{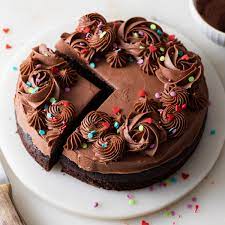} &
        \includegraphics[width=0.0915\textwidth]{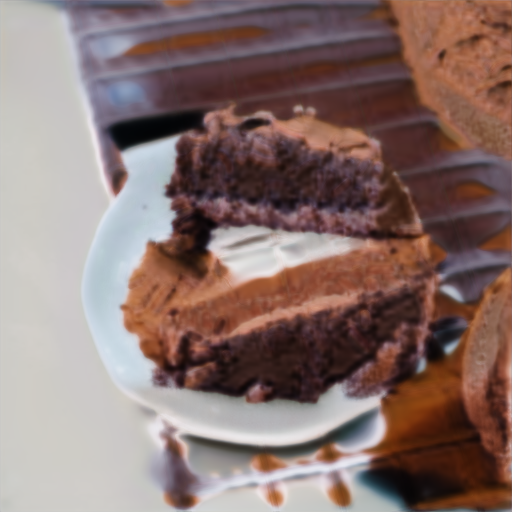} &
        \includegraphics[width=0.0915\textwidth]{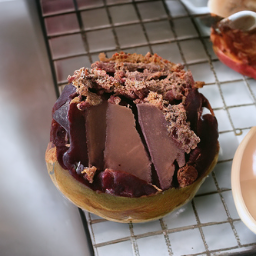} &
        \includegraphics[width=0.0915\textwidth]{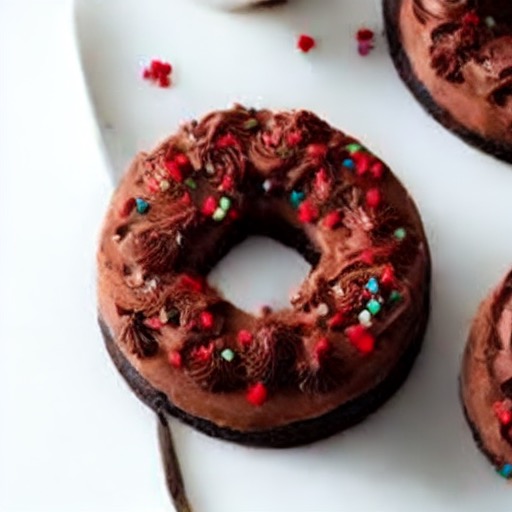} \\
        
        \includegraphics[width=0.0915\textwidth]{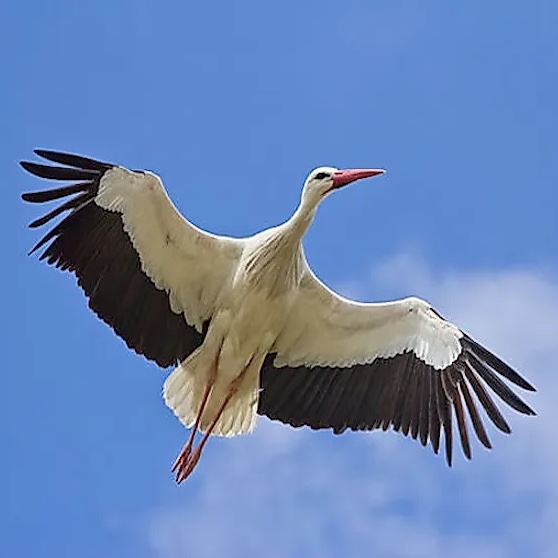} &
        \includegraphics[width=0.0915\textwidth]{images/inputs/birds/hummingbird.jpg} &
        \includegraphics[width=0.0915\textwidth]{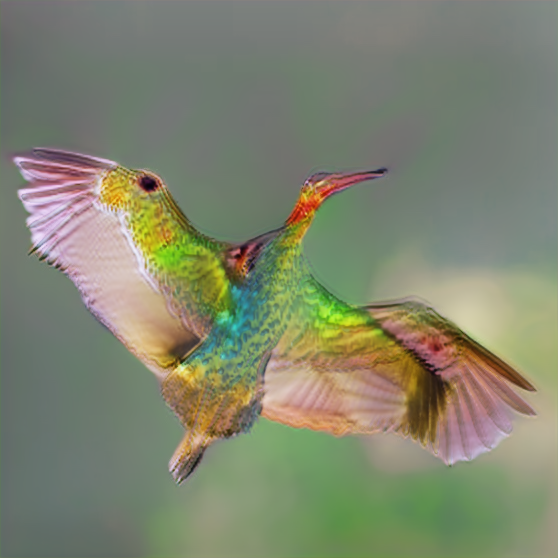} &
        \includegraphics[width=0.0915\textwidth]{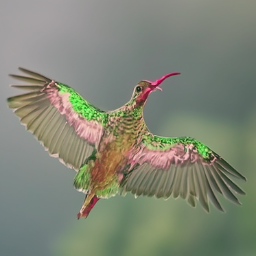} &
        \includegraphics[width=0.0915\textwidth]{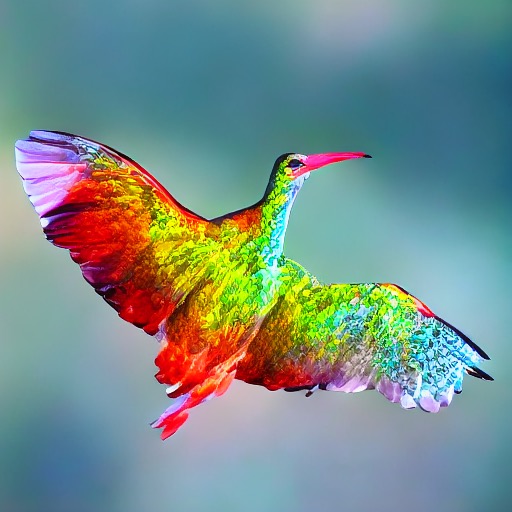} \\

        \includegraphics[width=0.0915\textwidth]{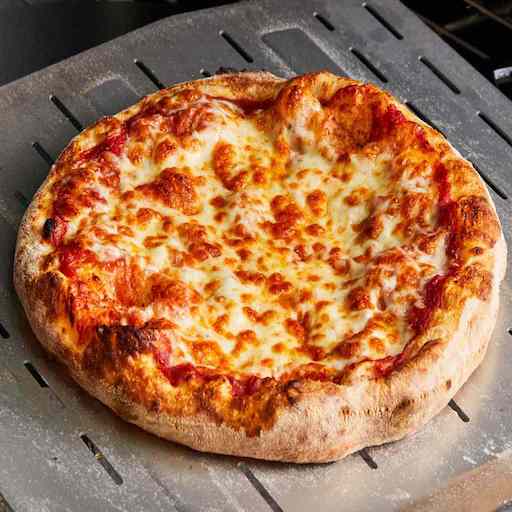} &
        \includegraphics[width=0.0915\textwidth]{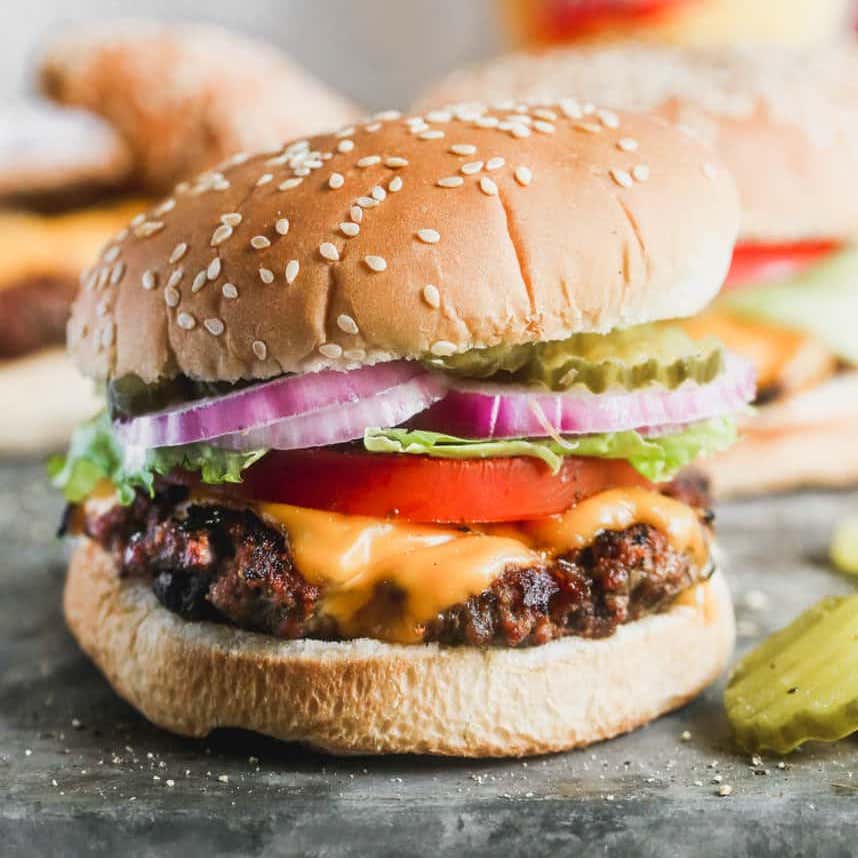} &
        \includegraphics[width=0.0915\textwidth]{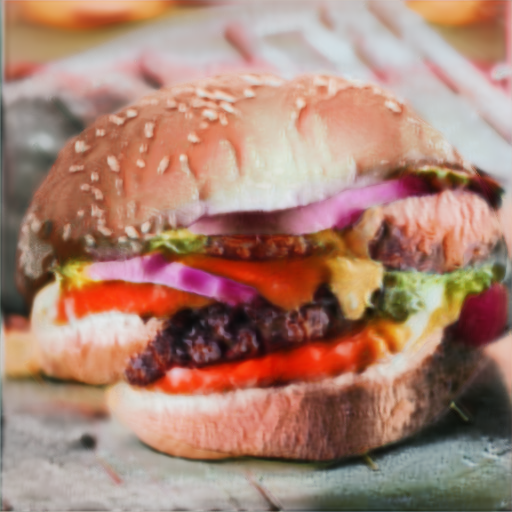} &
        \includegraphics[width=0.0915\textwidth]{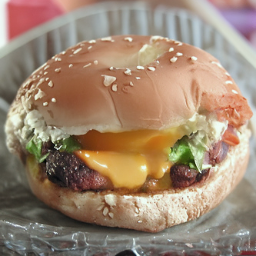}  &
        \includegraphics[width=0.0915\textwidth]{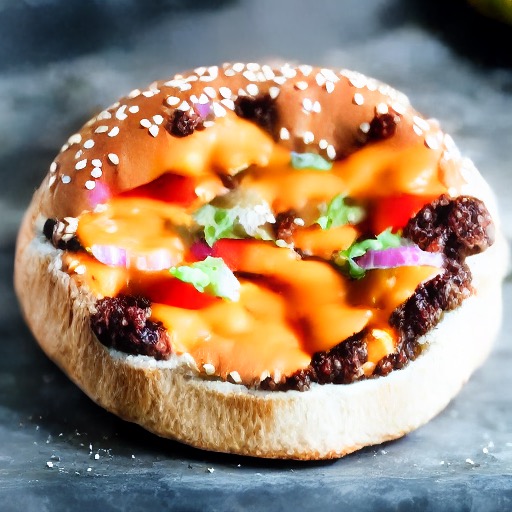} \\

        \includegraphics[width=0.0915\textwidth]{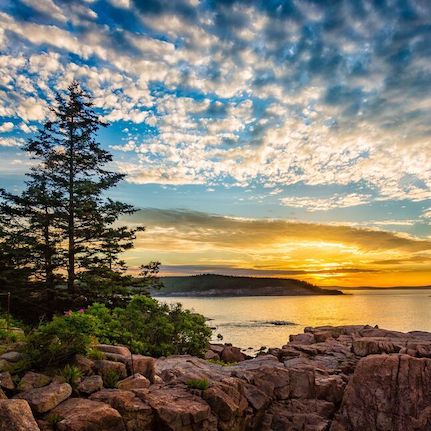} &
        \includegraphics[width=0.0915\textwidth]{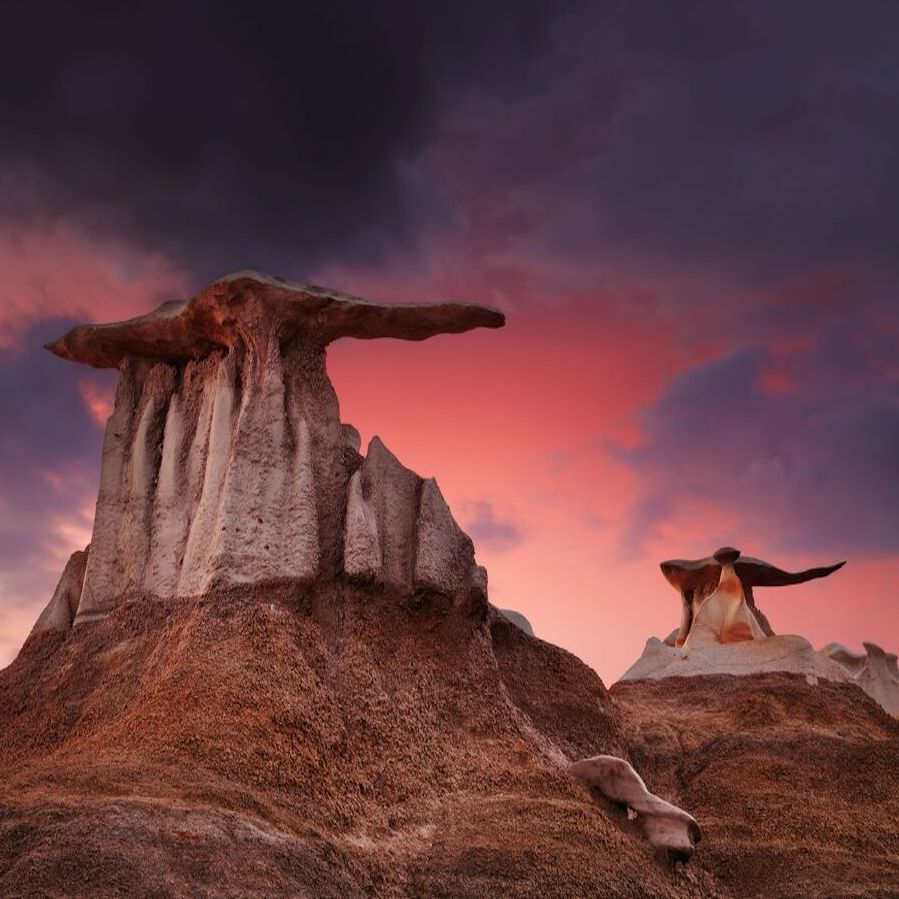} &
        \includegraphics[width=0.0915\textwidth]{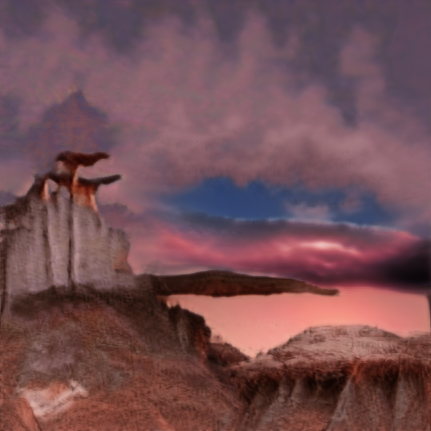} &
        \includegraphics[width=0.0915\textwidth]{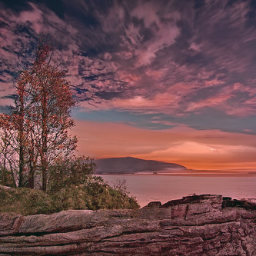} &
        \includegraphics[width=0.0915\textwidth]{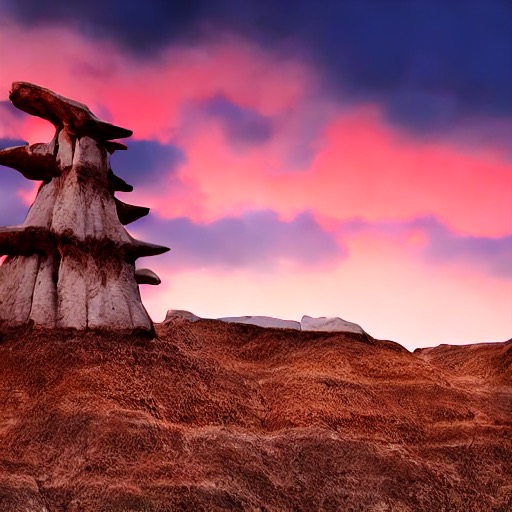} \\

        \includegraphics[width=0.0915\textwidth]{images/inputs/buildings/eiffel_tower.jpg} &
        \includegraphics[width=0.0915\textwidth]{images/inputs/buildings/saint_basil.jpg} &
        \includegraphics[width=0.0915\textwidth]{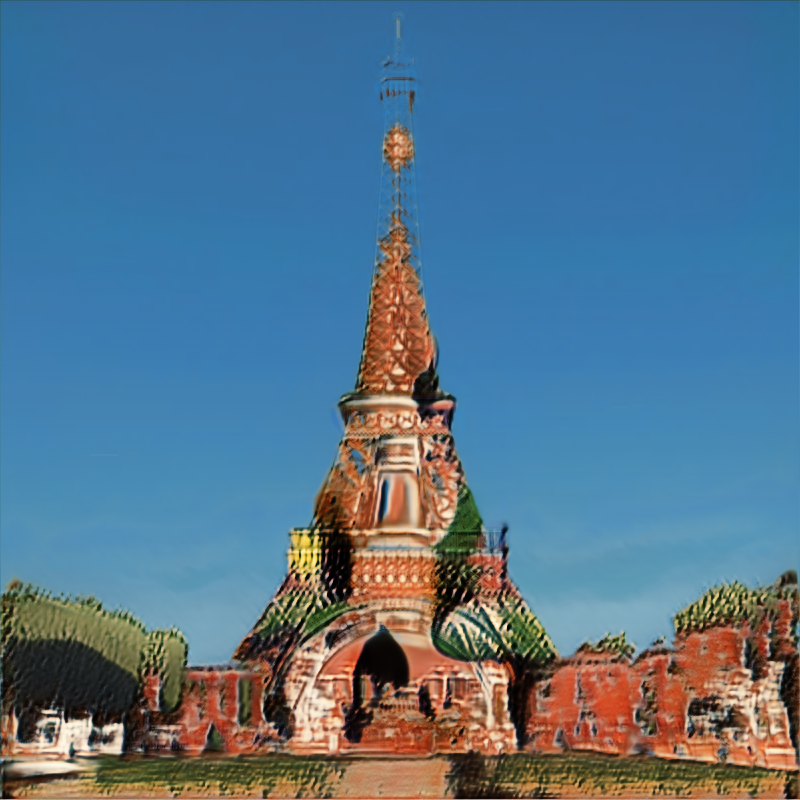} &
        \includegraphics[width=0.0915\textwidth]{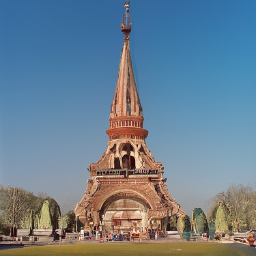} &
        \includegraphics[width=0.0915\textwidth]{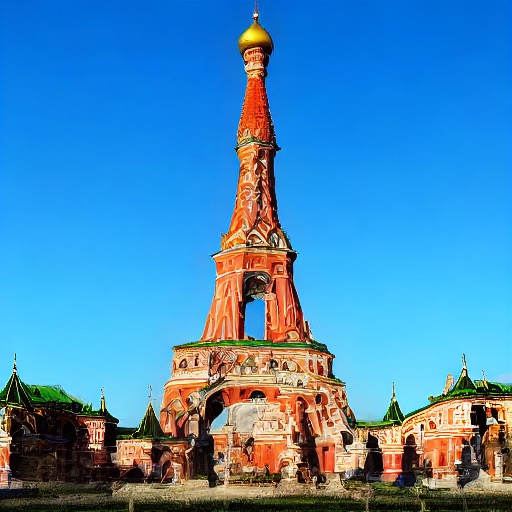} \\

        \includegraphics[width=0.0915\textwidth]{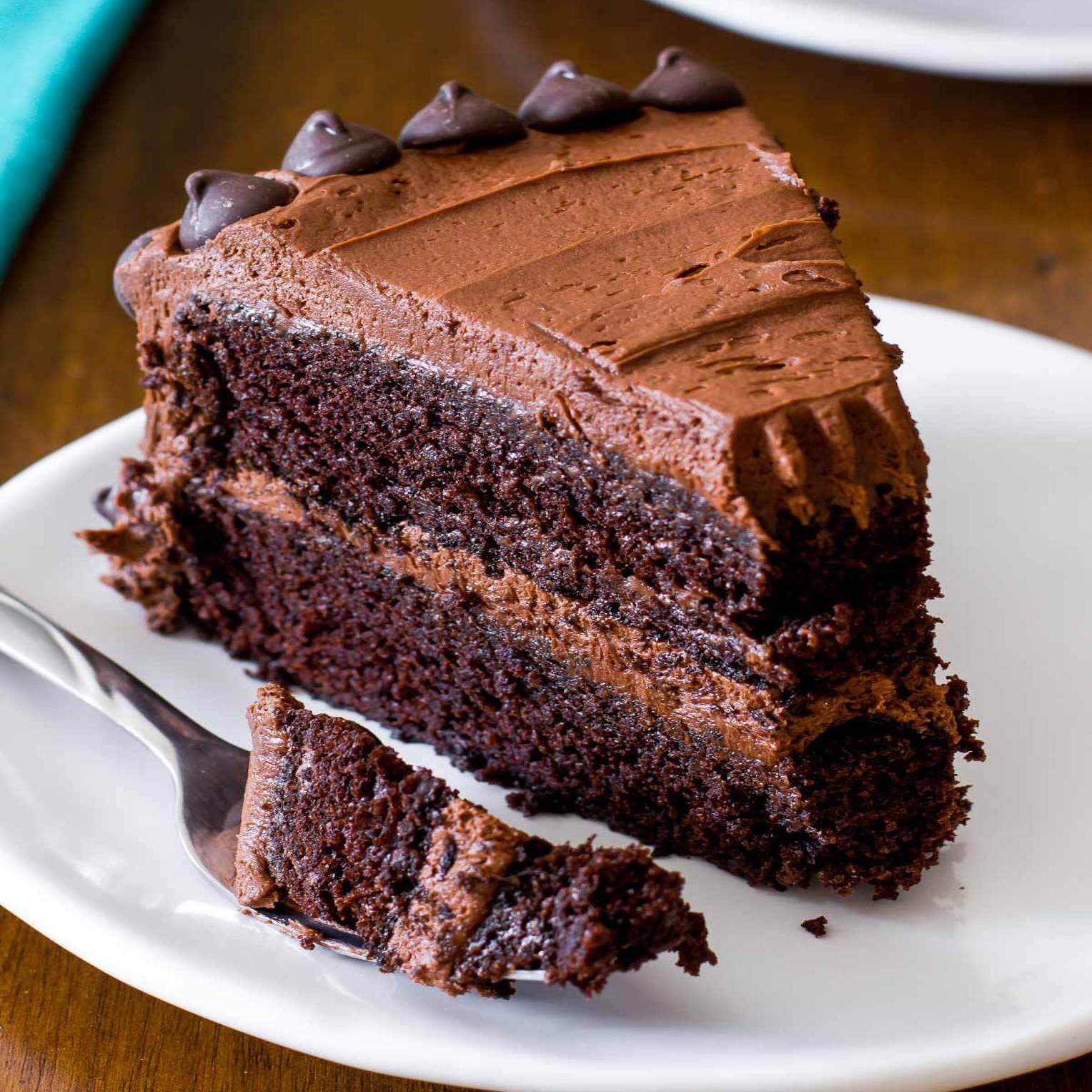} &
        \includegraphics[width=0.0915\textwidth]{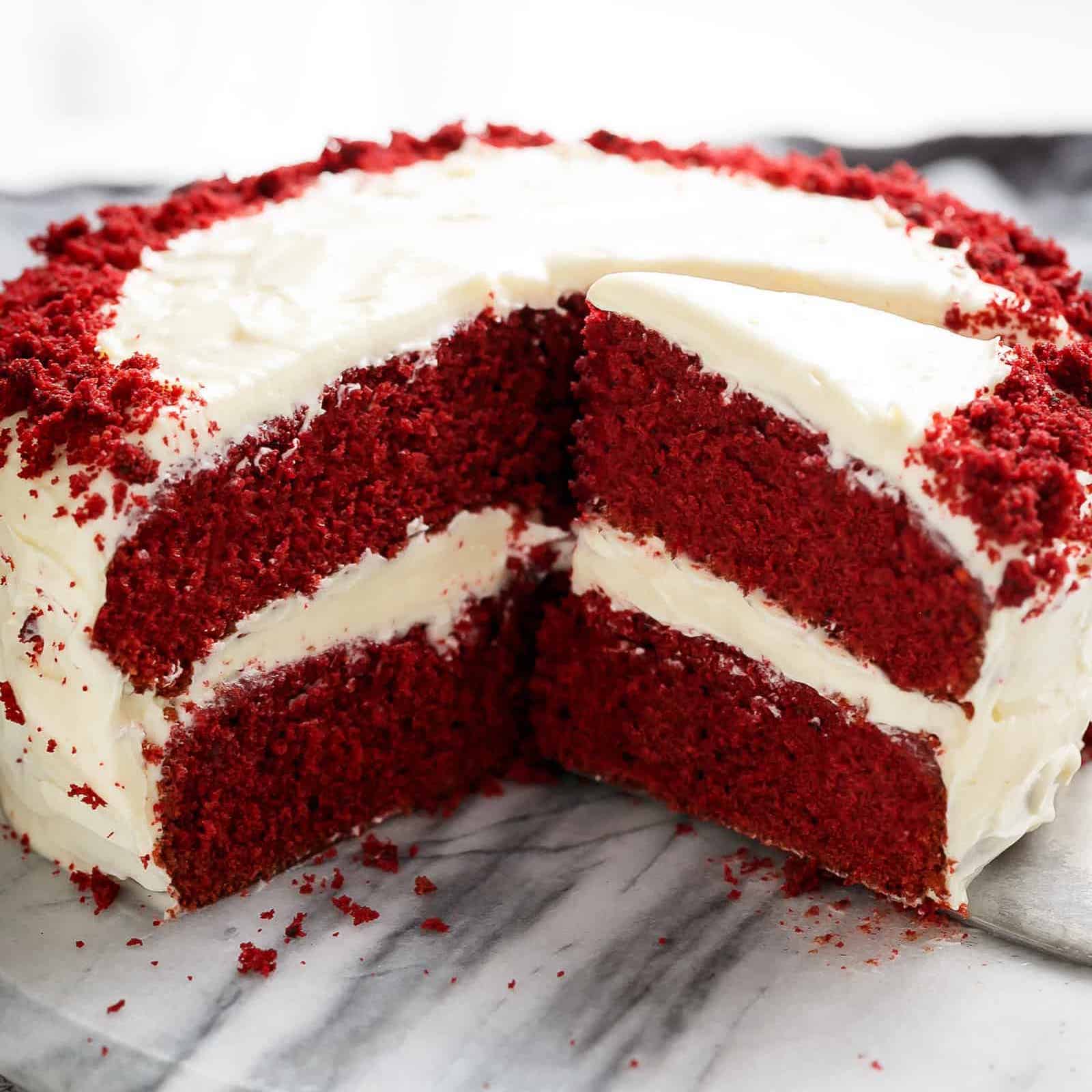} &
        \includegraphics[width=0.0915\textwidth]{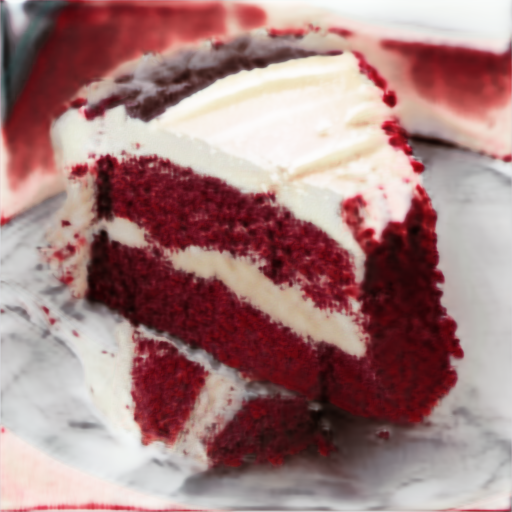} &
        \includegraphics[width=0.0915\textwidth]{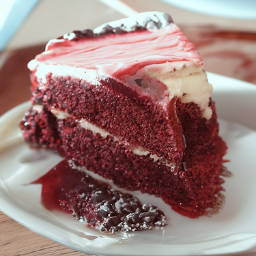} &
        \includegraphics[width=0.0915\textwidth]{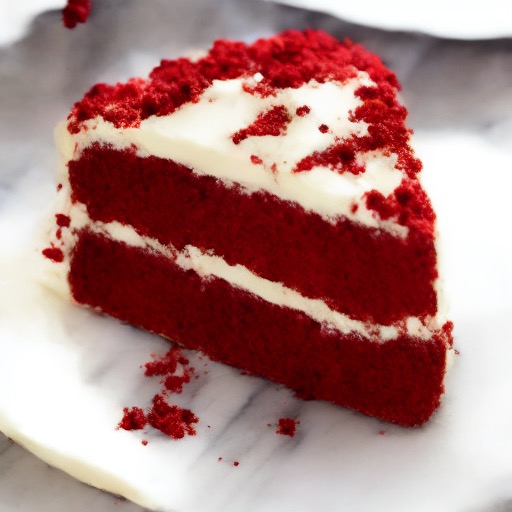} \\

        \includegraphics[width=0.0915\textwidth]{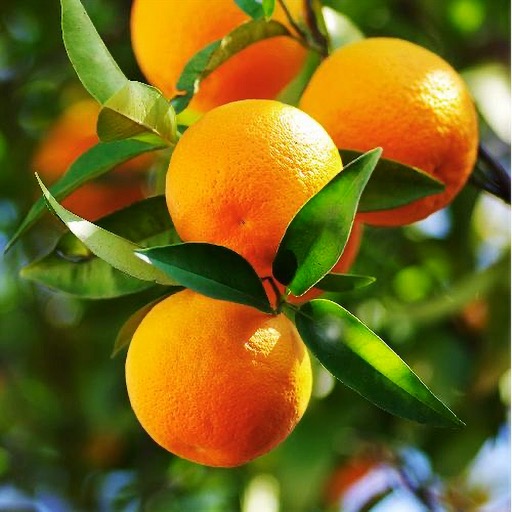} &
        \includegraphics[width=0.0915\textwidth]{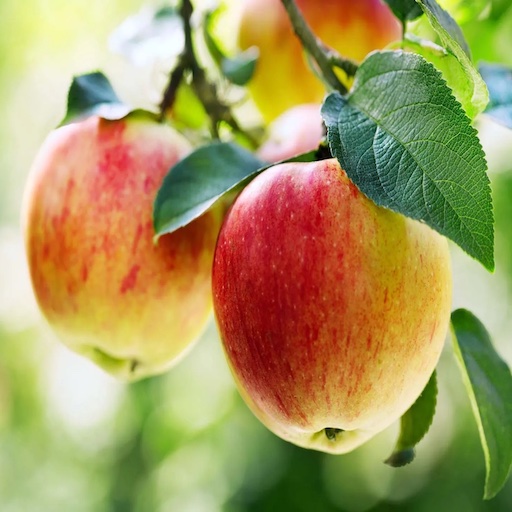} &
        \includegraphics[width=0.0915\textwidth]{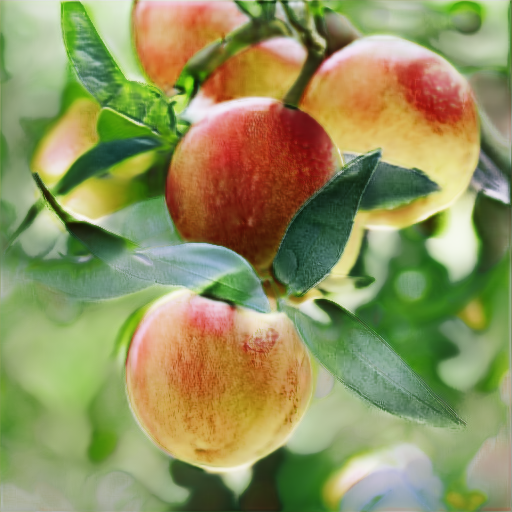} &
        \includegraphics[width=0.0915\textwidth]{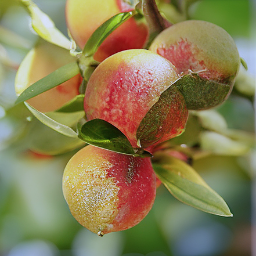} &
        \includegraphics[width=0.0915\textwidth]{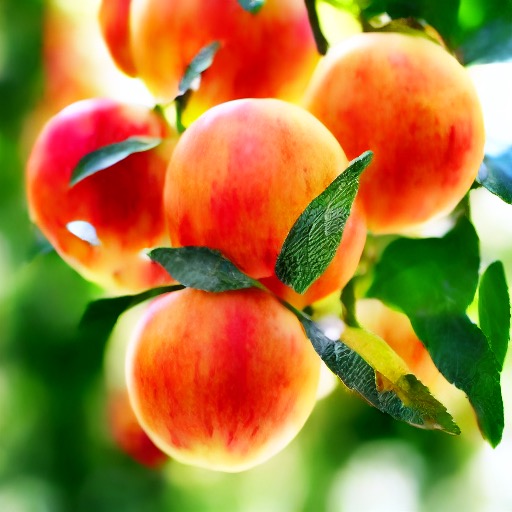} \\

        Structure & Appearance & SpliceViT & DiffuseIT & Ours
        
    \end{tabular}
    
    }
    \vspace{-0.2cm}
    \caption{
    Qualitative comparison to additional appearance transfer techniques. In each row, we provide the input structure and appearance images, followed by the results obtained by each method.
    }
    \label{fig:compare_figure}
\end{figure}

\vspace{-0.2cm}
\paragraph{Qualitative Comparison}
We now turn to qualitatively compare our cross-image attention mechanism to existing appearance transfer techniques. Since Swapping Autoencoder (SA) requires a dedicated generator for each domain, we begin with a comparison to SA using their pretrained church and AFHQ~\cite{choi2020stargan} models. 
Results are presented in~\Cref{fig:compare_swapping_ae}. 
SA effectively maintains the source structure while transferring the general color scheme of the target appearance. However, it often falls short of capturing the semantic details. For instance, it struggles to transfer the gold dome and colorful patterns in the leftmost image or the distinct church entrance in the rightmost image.
For the AFHQ dataset, SA can transfer the general color from the appearance image, but the resulting images strongly resemble the original structure image with minimal semantic changes. 
In contrast, our method preserves the general shape of the target structure while adapting its precise geometry to better capture the visual characteristics of the appearance image. 
For example, in the buildings domain, our method integrates the distinctive purple towers from the appearance image in the third column into the structure of the Eiffel Tower
Moreover, for the AFHQ dataset, our method adapts the target appearance and semantics to the shape of the cat, e.g., preserving the cat's pointed ears.

\begin{table}
\small
\centering
\setlength{\tabcolsep}{2.5pt}
\caption{Quantitative Comparison. We measure the level of structure preservation and appearance fidelity across all methods and various domains. To measure structure preservation, we calculate the mean IoU between binary masks extracted from the input structure image and the generated image. For appearance fidelity, we compute the distances between the Gram matrices of the input appearance image and the generated image.
\\[-0.65cm]} 
\begin{tabular}{l c c c c} 
    \toprule
    & \multicolumn{4}{c}{Structure Preservation $\uparrow$} \\
    \midrule
    Domain & Swapping AE & SpliceViT & DiffuseIT & \textbf{Ours} \\
    \midrule
    Buildings & $0.82$ & $0.56$ & $0.79$ & $0.76$ \\
    Animal Faces & $0.59$ & $0.71$ & $0.96$ & $0.68$ \\
    Animals & N/A & $0.71$ & $0.80$ & $0.75$ \\
    Cars & N/A & $0.93$ & $0.94$ & $0.88$ \\
    Birds & N/A & $0.66$ & $0.77$ & $0.70$ \\
    Cakes & N/A & $0.64$ & $0.66$ & $0.61$ \\
    \bottomrule
    Average & $0.71$ & $0.70$ & $0.82$ & $0.73$ \\
    \toprule
    & \multicolumn{4}{c}{Appearance Fidelity $\downarrow$} \\
    \midrule
    Domain & Swapping AE & SpliceViT & DiffuseIT & \textbf{Ours} \\
    \midrule
    Buildings & $0.88$ & $0.61$ & $1.05$ & $1.24$ \\
    Animal Faces & $0.58$ & $1.11$ & $1.74$ & $0.27$ \\
    Animals & N/A & $2.53$ & $2.89$ & $1.41$ \\
    Cars & N/A & $0.56$ & $0.50$ & $0.21$ \\
    Birds & N/A & $0.14$ & $0.15$ & $0.41$ \\
    Cakes & N/A & $0.41$ & $0.49$ & $0.21$ \\
    \bottomrule
    Average & $0.73$ & $0.89$ & $1.14$ & $0.62$ \\
    \bottomrule
\end{tabular}
\label{tb:quantitative}
\vspace{-0.1cm}
\end{table}

In~\Cref{fig:compare_figure} we present a comparison with techniques supporting appearance transfer between objects found in natural images. 
First, SpliceViT can transfer appearance between objects with similar shapes and viewpoints such as between the zebra and tiger in the first row or oranges and apples in the bottom row. 
However, when the two objects differ significantly in their visual characteristics, SpliceViT fails to find meaningful semantic correspondences. This results in heavy artifacts in the outputs, as seen in the second and third rows. Notably, SpliceViT requires a per-image generator tuning spanning dozens of minutes on a commercial GPU. 
While DiffuseIT attains results comparable to SpliceViT without the need for per-image model training, it still struggles to achieve high-quality transfer results in natural images. 
In contrast, our method can accurately transfer appearance between objects that vary in the number of instances (first and last rows) and between objects differing in shape (third and second-to-last row) and viewpoint (second and fourth rows). Moreover, our approach operates in a zero-shot setting while requiring no external models to guide the disentanglement process. Instead, we rely on the rich internal representations already captured by the model.

\vspace{-0.15cm}
\paragraph{Quantitative Comparison}
We quantitatively evaluate each considered method in two aspects: (1) how well they preserve the source structure, and (2) how well the generated images depict the target appearance. To measure structure preservation, we first extract binary masks over the input structure images and corresponding output images using SAM~\cite{kirillov2023segany}. We then measure the mean IoU of the output images with respect to the input structure images. As there is no standard automatic metric for assessing semantic-based appearance fidelity, we turn to the neural style transfer literature which has demonstrated that images with similar styles tend to have similar Gram matrices~\cite{gatys2015neural}. As such, we measure the $L_2$ distance between the Gram matrices of the input style and output images computed along five intermediate layers of a pretrained VGG19~\cite{simonyan2014very} network. 

As Swapping Autoencoder~\cite{park2020swapping} is limited in its supported domains, we compute the above metrics across six domains (buildings, animal faces, animals, cars, birds, and cakes). For each domain, we selected $20$ structure-appearance input pairs. Results are displayed in~\Cref{tb:quantitative}. 
As shown, our method demonstrates comparable performance to the alternative methods across all domains in both structure preservation and appearance fidelity. It is worth noting that achieving faithful semantic transfer often necessitates minor structure modifications. For example, merely transferring the general color scheme between images would lead to a high mean IoU, but would fail to capture the true semantics of the appearance image. Our method offers a favorable balance between preserving the precise input geometry and capturing the prominent semantics of the target appearance, as also supported by our qualitative evaluations. 

\begin{table}
\small
\centering
\caption{User Study. We asked respondents to select which set of images they most preferred based on their faithfulness to the input structure and appearance as well as the overall quality of the generated images. Results are averaged across all responses.
\\[-0.65cm]} 
\begin{tabular}{l c c c c} 
    \toprule
    & \multicolumn{3}{c}{Buildings} \\
    \midrule
    Method & Structure & Appearance & Overall Quality \\
    \midrule
    Swapping AE & $44.3\%$ & $3.1\%$ & $20.9\%$ \\
    SpliceViT & $2\%$ & $17.7\%$ & $2.3\%$ \\
    DiffuseIT & $16.4\%$ & $2.9\%$ & $10.4\%$ \\
    \textbf{Ours} & $37.3\%$ & $76.3\%$ & $66.4\%$ \\
    \toprule
    & \multicolumn{3}{c}{Animals, Cars, Cakes, Birds} \\
    \midrule
    Method & Structure & Appearance & Overall Quality \\
    \midrule
    SpliceViT & $11.0\%$ & $21.4\%$ & $9.6\%$ \\
    DiffuseIT & $44.8\%$ & $8.5\%$ & $30.3\%$ \\
    \textbf{Ours} & $44.2\%$ & $70.1\%$ & $60.1\%$ \\
    \bottomrule
\end{tabular}
\label{tb:user_study}
\vspace{-0.1cm}
\end{table}

\vspace{-0.2cm}
\paragraph{User Study}
Finally, we conduct a user study to analyze all techniques across five object domains (buildings, animals, cars, cakes, and birds). For each domain, we selected multiple structure-appearance input pairs and generated transfer results using each of the four considered methods. Additional details on the evaluation setup are provided in~\Cref{sec:additional_details}.
For each pair, participants were tasked with evaluating the results based on three key aspects: (1) how well the source structure was preserved, (2) how well the output depicted the target appearance, and (3) the overall quality of the generated image. 
Participants were presented with the outputs from all relevant methods and were asked to select the most favorable result for each aspect.

Results are presented in~\Cref{tb:user_study} where the final score for each method is calculated by averaging the number of times participants selected that approach across all questions. In the buildings domain, Swapping Autoencoder outperforms all methods, which is likely due to its per-domain training. However, our method achieves a comparable level of structure preservation while significantly surpassing all other methods in the ability to capture the target appearance and generate high-quality images. In the remaining four domains, our method consistently outperforms both SpliceViT and DiffuseIT in appearance preservation and quality while achieving better or comparable structure preservation.

\begin{figure}
    \centering
    \setlength{\tabcolsep}{0.4pt}
    \addtolength{\belowcaptionskip}{-10pt}

    {\small

    \begin{tabular}{c c c c c c}

        \raisebox{0.025in}{\rotatebox{90}{ Structure }} &
        \includegraphics[width=0.0915\textwidth]{images/inputs/buildings/duomo.jpg} &
        \includegraphics[width=0.0915\textwidth]{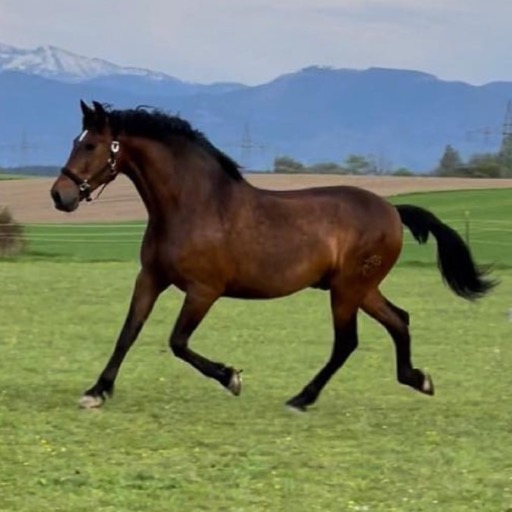} &
        \includegraphics[width=0.0915\textwidth]{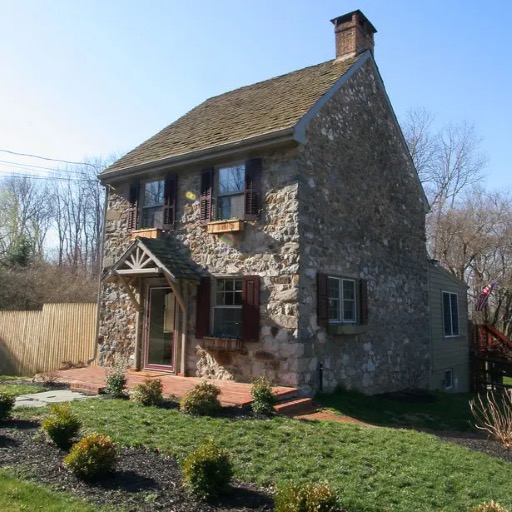} &
        \includegraphics[width=0.0915\textwidth]{images/inputs/cake/chocolate_cake.jpg} &
        \includegraphics[width=0.0915\textwidth]{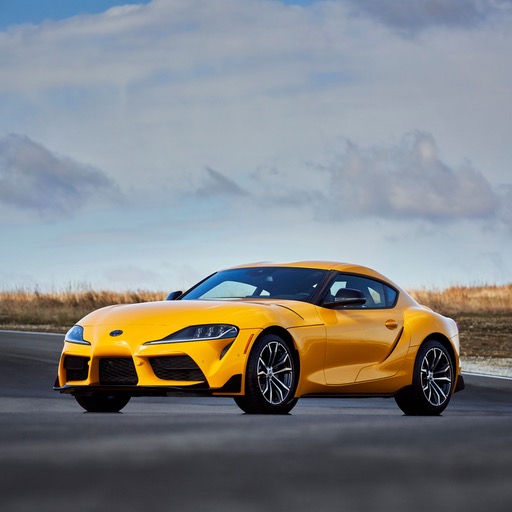} \\

        \raisebox{-0.025in}{\rotatebox{90}{ Appearance }} &
        \includegraphics[width=0.0915\textwidth]{images/inputs/buildings/chile-church.jpg} &
        \includegraphics[width=0.0915\textwidth]{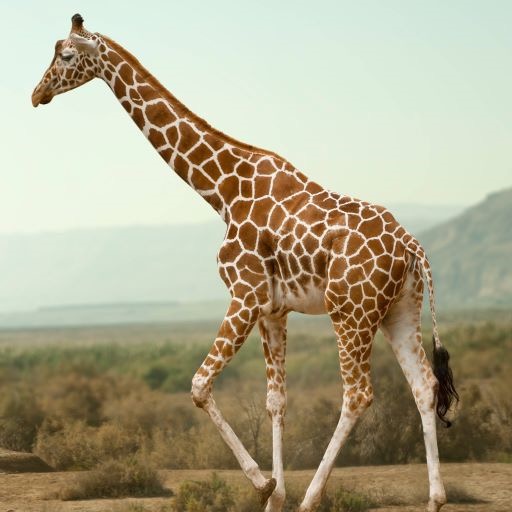} &
        \includegraphics[width=0.0915\textwidth]{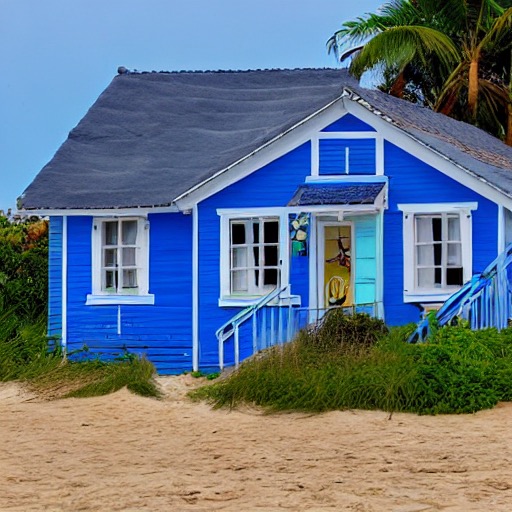} &
        \includegraphics[width=0.0915\textwidth]{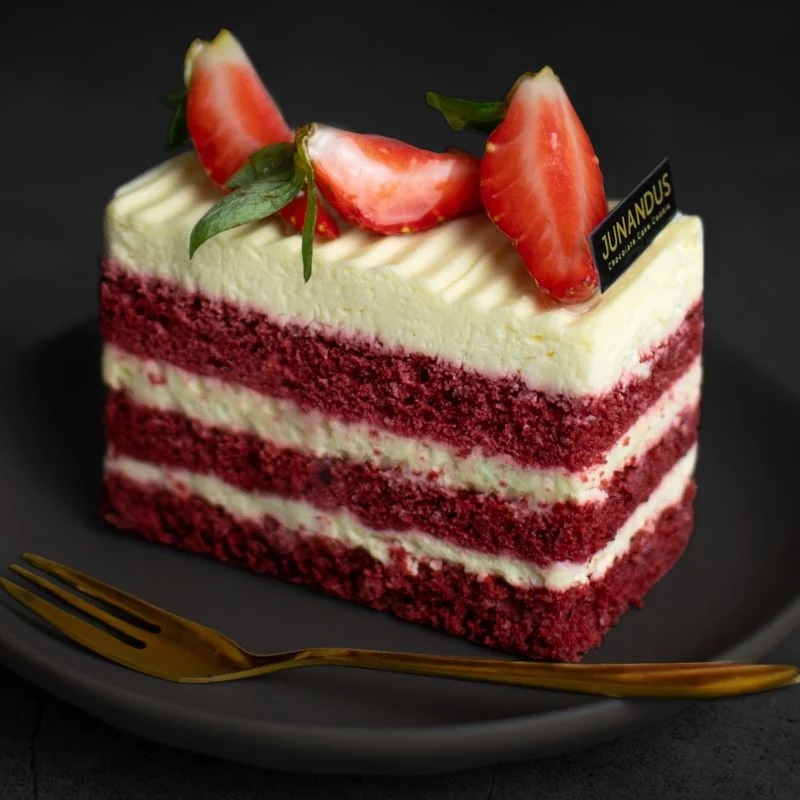} &
        \includegraphics[width=0.0915\textwidth]{images/inputs/cars/vintage.jpg} \\

        \raisebox{0.05in}{\rotatebox{90}{ Baseline }} &
        \includegraphics[width=0.0915\textwidth]{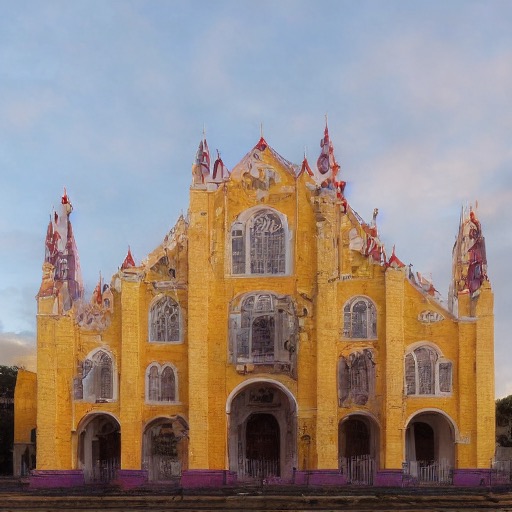} &
        \includegraphics[width=0.0915\textwidth]{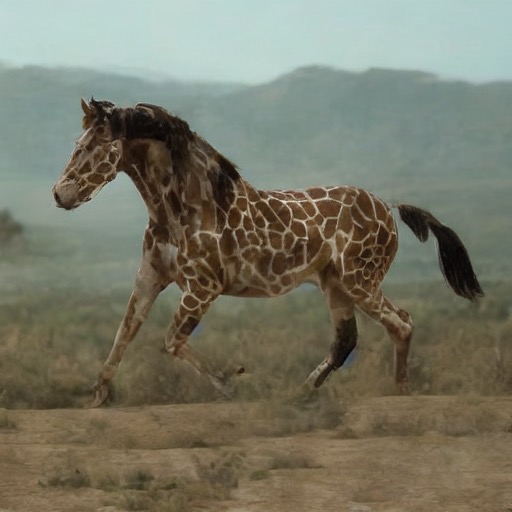} &
        \includegraphics[width=0.0915\textwidth]{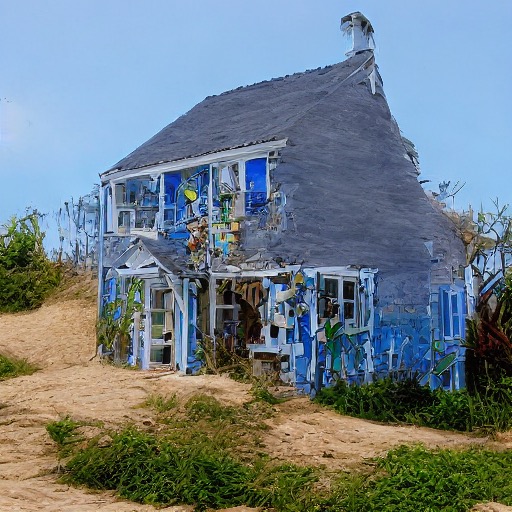} &
        \includegraphics[width=0.0915\textwidth]{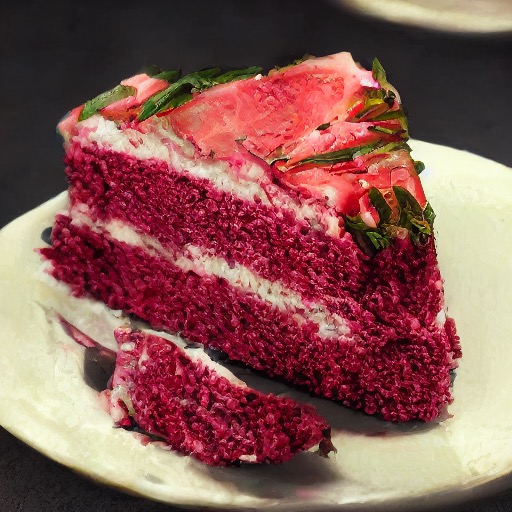} &
        \includegraphics[width=0.0915\textwidth]{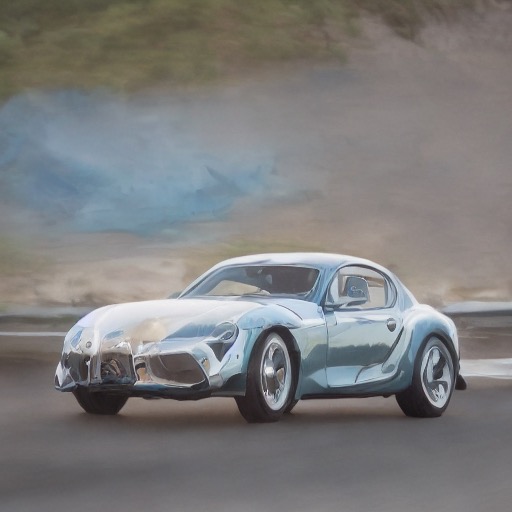} \\

        \raisebox{-0.025in}{\rotatebox{90}{ $+$ Contrast }} &
        \includegraphics[width=0.0915\textwidth]{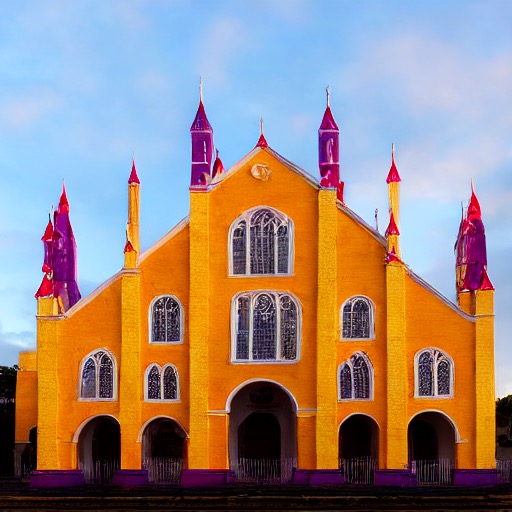} &
        \includegraphics[width=0.0915\textwidth]{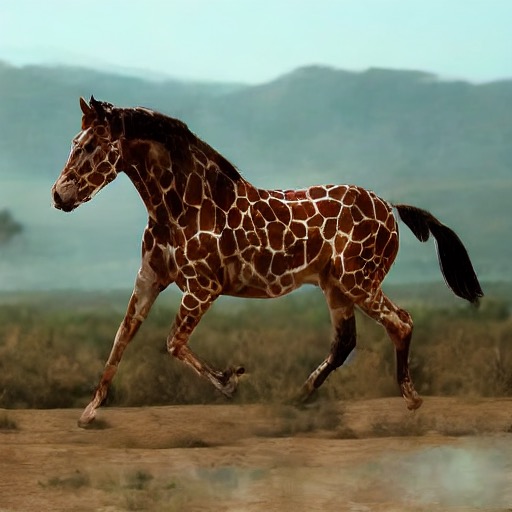} &
        \includegraphics[width=0.0915\textwidth]{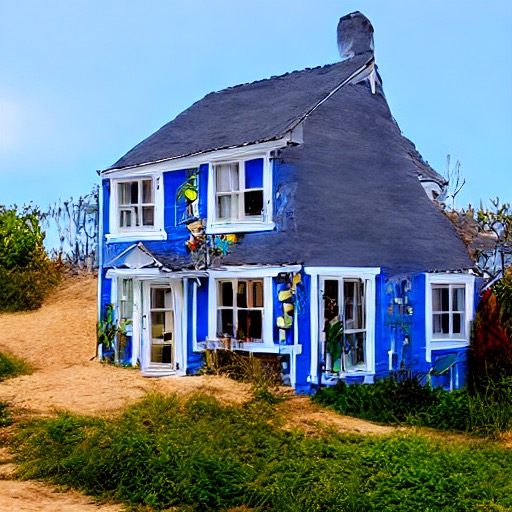} &
        \includegraphics[width=0.0915\textwidth]{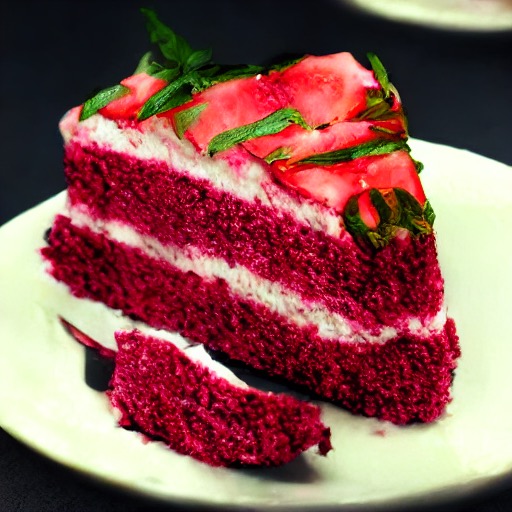} &
        \includegraphics[width=0.0915\textwidth]{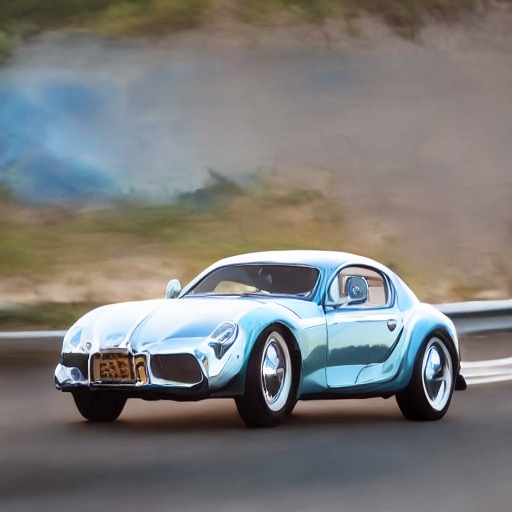} \\

        \raisebox{0.025in}{\rotatebox{90}{ $+$ AdaIN }} &
        \includegraphics[width=0.0915\textwidth]{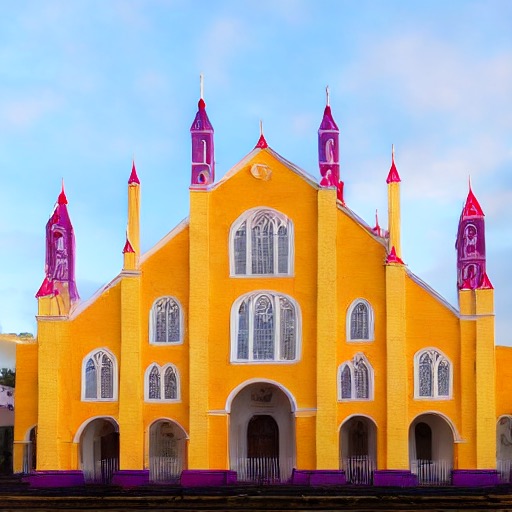} &
        \includegraphics[width=0.0915\textwidth]{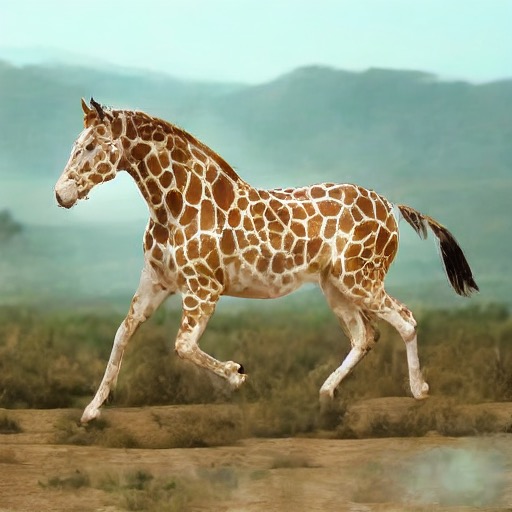} &
        \includegraphics[width=0.0915\textwidth]{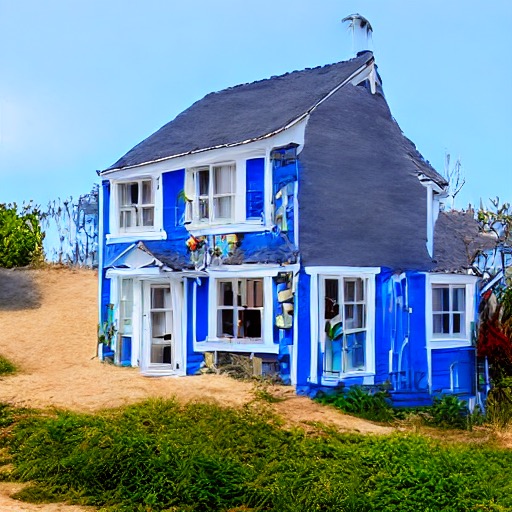} &
        \includegraphics[width=0.0915\textwidth]{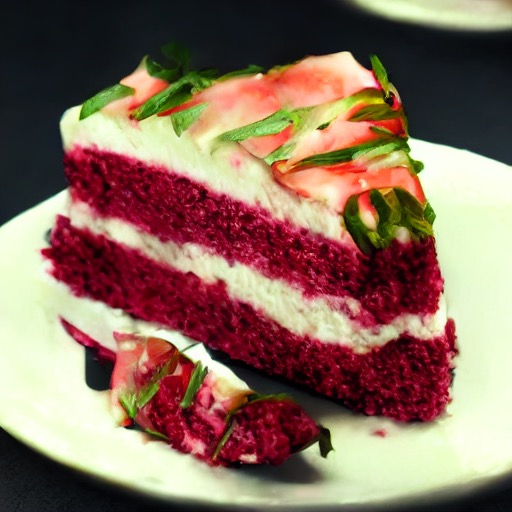} &
        \includegraphics[width=0.0915\textwidth]{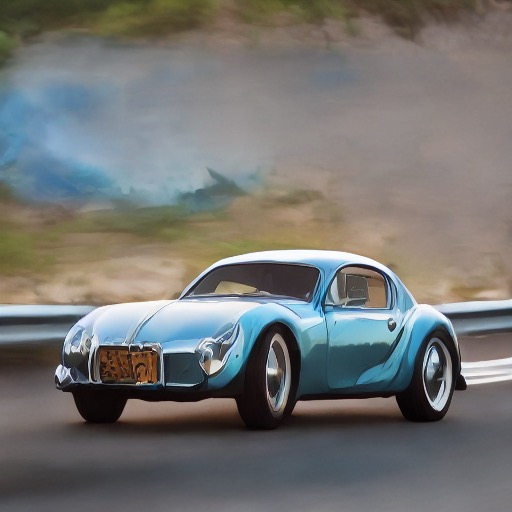} \\

        \raisebox{-0.025in}{\rotatebox{90}{ $+$ Guidance }} &
        \includegraphics[width=0.0915\textwidth]{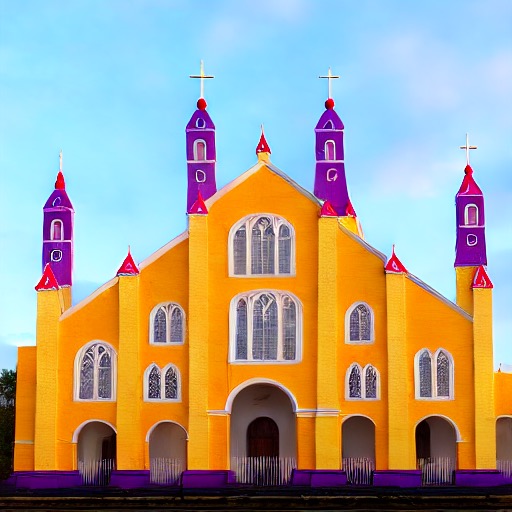} &
        \includegraphics[width=0.0915\textwidth]{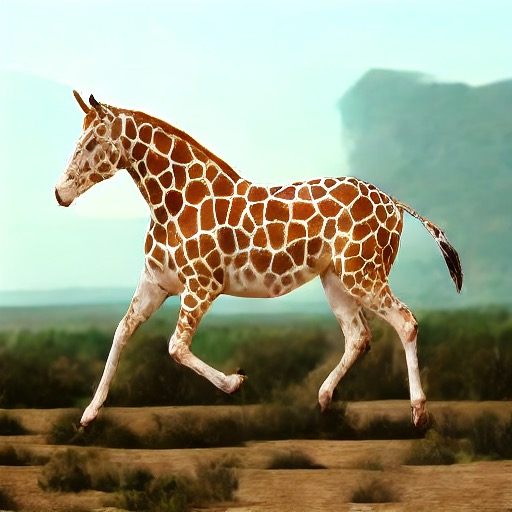} &
        \includegraphics[width=0.0915\textwidth]{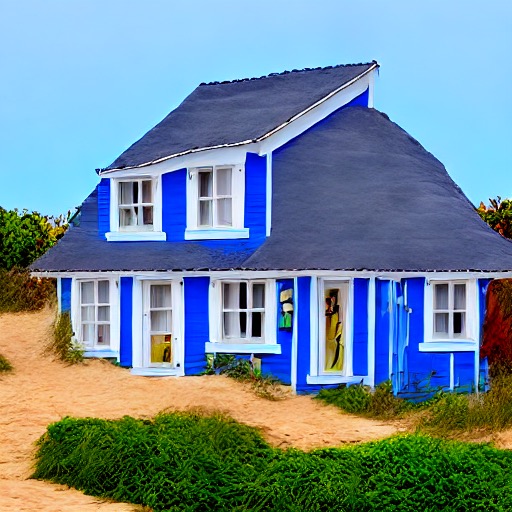} &
        \includegraphics[width=0.0915\textwidth]{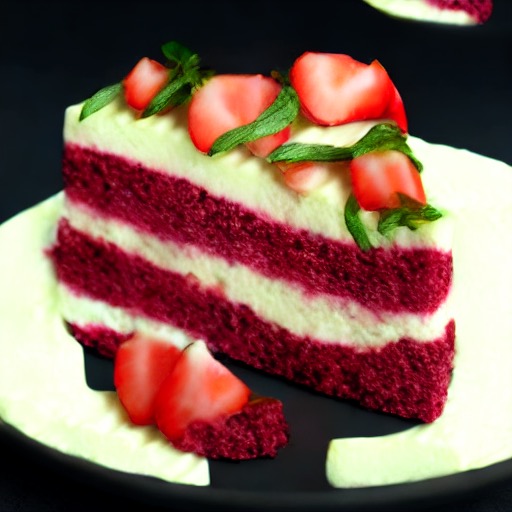} &
        \includegraphics[width=0.0915\textwidth]{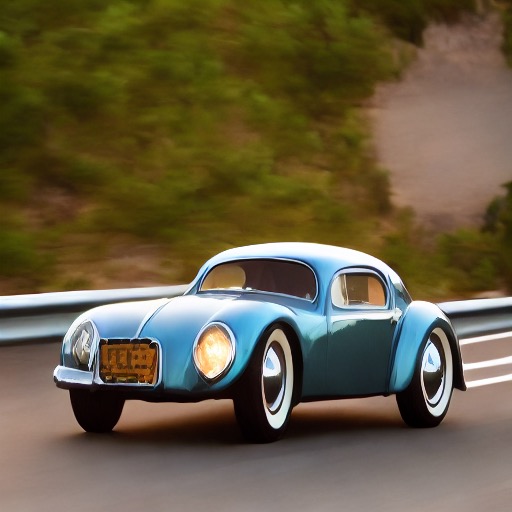} \\
 
    \end{tabular}

    }
    \vspace{-0.2cm}
    \caption{Ablation Study. In each row, we add an additional component of our appearance transfer scheme. Images in the bottom row represent results obtained by our complete method.}
    \label{fig:ablation_study}
\end{figure}

\subsection{Ablation Study}~\label{sec:ablations}
Finally, we perform an ablation study to validate the key design choices of our method. Specifically, we assess the contribution of (1) the attention map contrasting operation, (2) the AdaIN normalization over the noised latent code, and (3) our appearance guidance technique performed over the noise estimates of the denoising network. 
The results are presented in~\Cref{fig:ablation_study}. For our baseline, we simply swapping the standard self-attention layer with our cross-image attention layer. As shown, while the general semantics are transferred from the appearance images to the structure images, many artifacts are present in the outputs. In each subsequent row, we add an additional component to our technique, with the final row representing our complete method. 
As shown, applying the contrasting operation significantly reduces the artifacts present in the baselines. By employing the AdaIN operation, we can better refine the general color distribution of the output as can be seen in the second column. Finally, incorporating the appearance guidance throughout the denoising process significantly improves the overall quality of results by refining finer-level details in the image. For example, observe the purple towers present in the leftmost column or the strawberries on the cake in the fourth column.  
\section{Limitations and Discussion}
\begin{figure}
    \centering
    \setlength{\tabcolsep}{0.5pt}
    \addtolength{\belowcaptionskip}{-10pt}
    {\scriptsize
    \begin{tabular}{c c c c c}

        \includegraphics[width=0.09\textwidth]{images/inputs/birds/airplane.jpeg} &
        \includegraphics[width=0.09\textwidth]{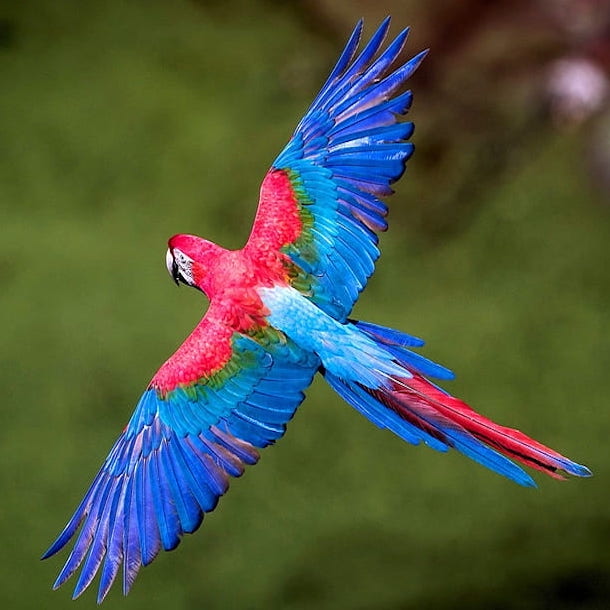} &
        \includegraphics[width=0.09\textwidth]{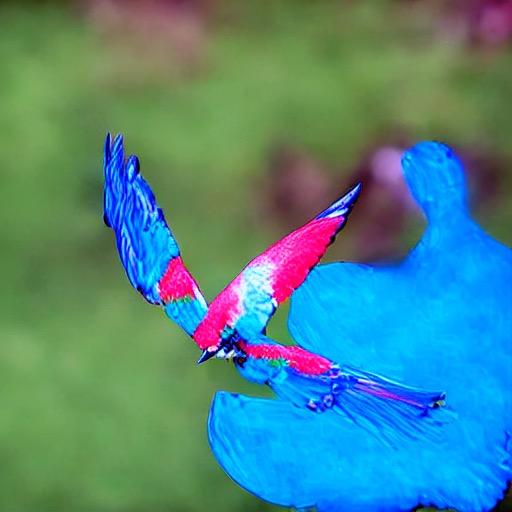} &
        \includegraphics[width=0.09\textwidth]{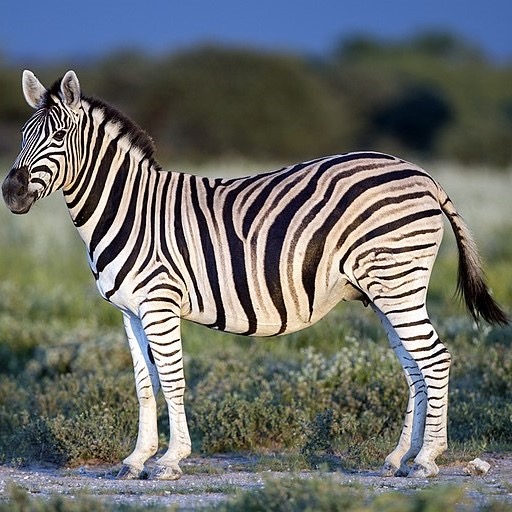} &
        \includegraphics[width=0.09\textwidth]{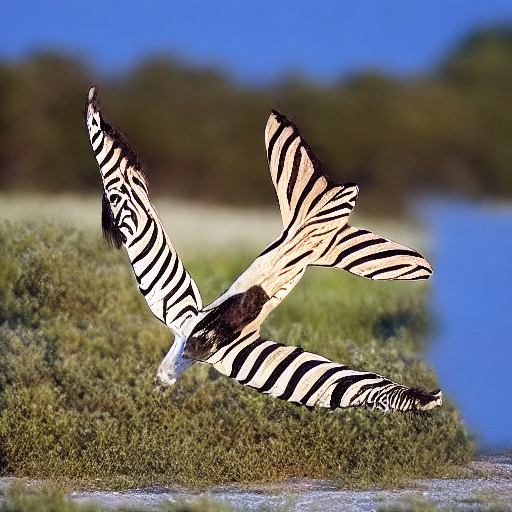} \\

        \includegraphics[width=0.09\textwidth]{images/inputs/vehicles/train.jpg} &
        \includegraphics[width=0.09\textwidth]{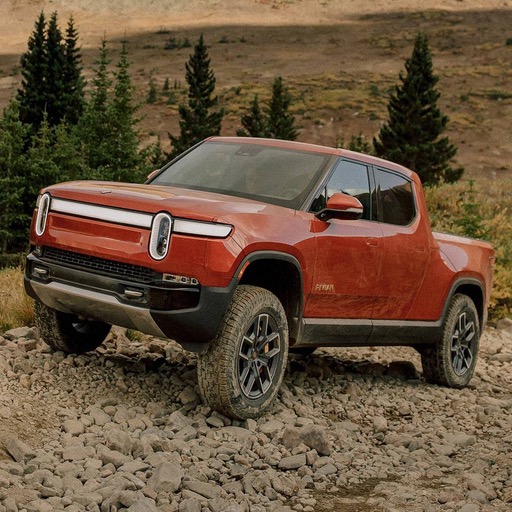} &
        \includegraphics[width=0.09\textwidth]{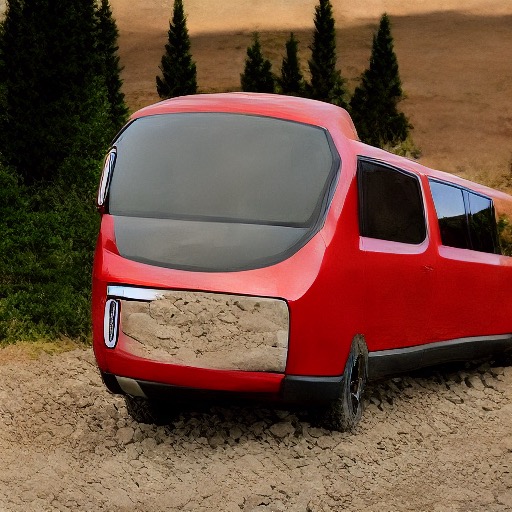} &
        \includegraphics[width=0.09\textwidth]{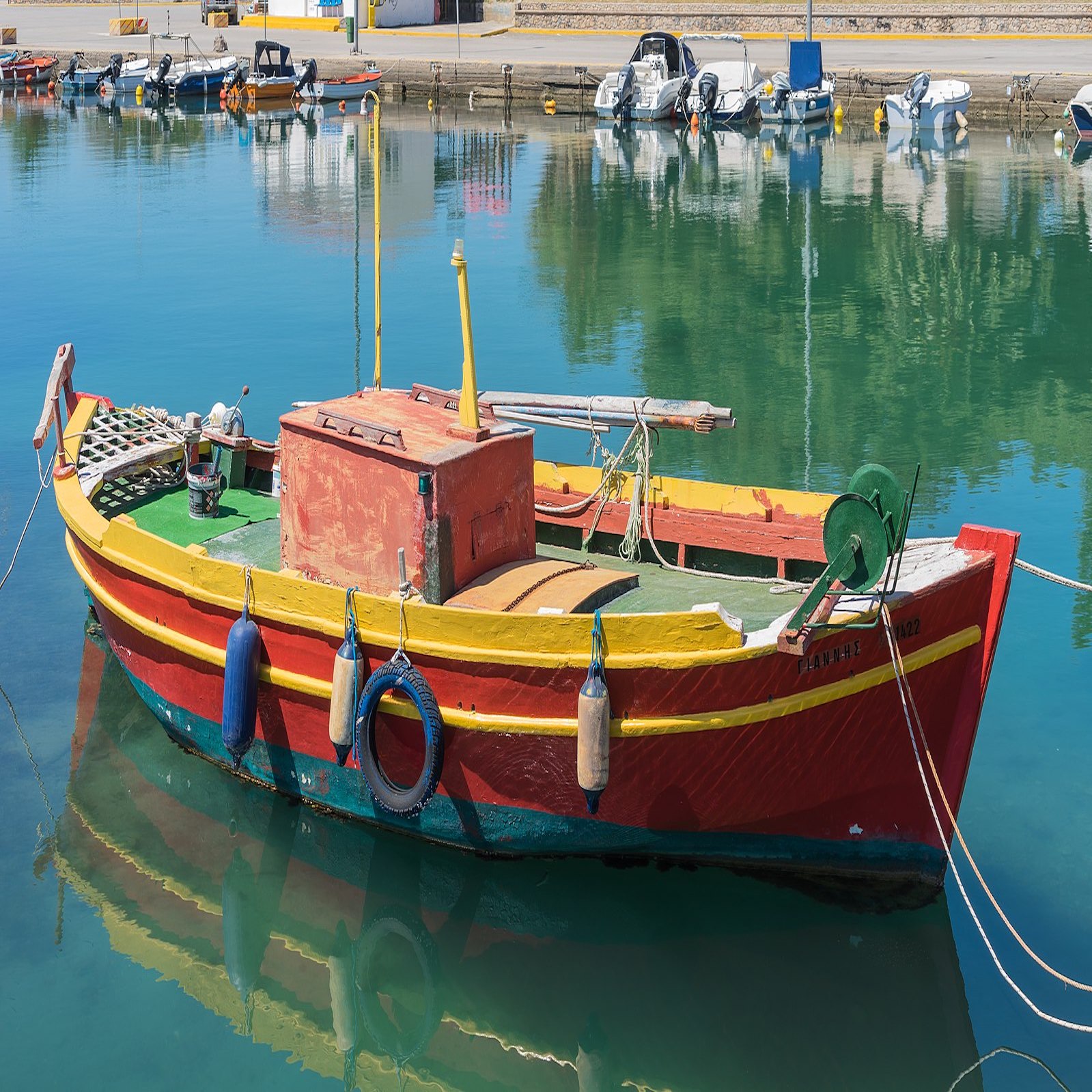} &
        \includegraphics[width=0.09\textwidth]{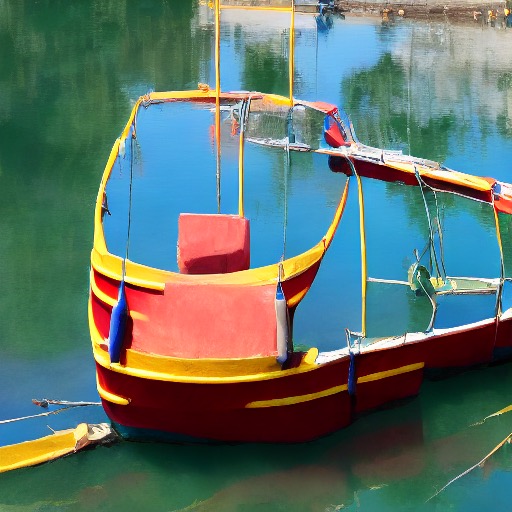} \\

        Structure & Appearance & Output & Appearance & Output \\ \\

        \includegraphics[width=0.09\textwidth]{images/inputs/animals/horse.jpg} &
        \includegraphics[width=0.09\textwidth]{images/inputs/animals/zebra.jpg} &
        \includegraphics[width=0.09\textwidth]{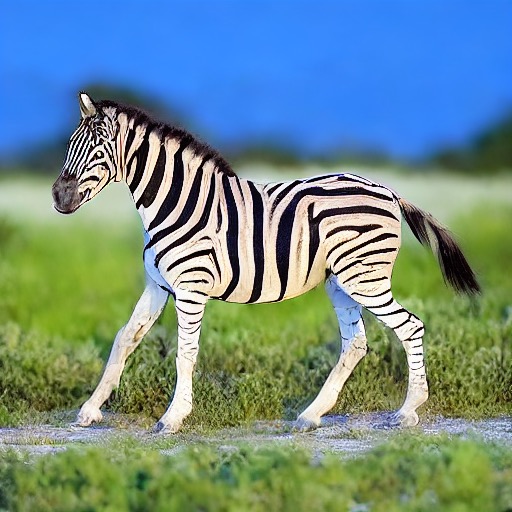} &
        \includegraphics[width=0.09\textwidth]{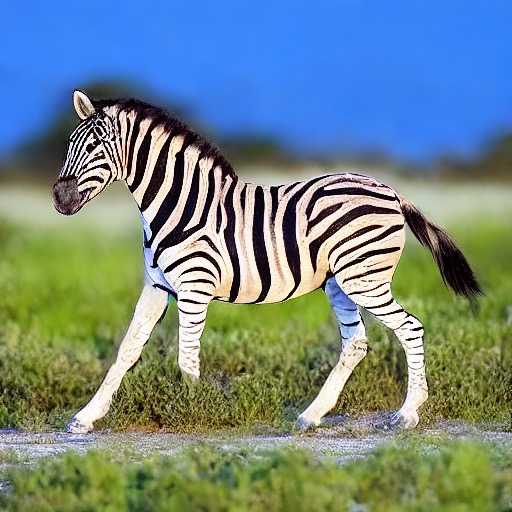} &
        \includegraphics[width=0.09\textwidth]{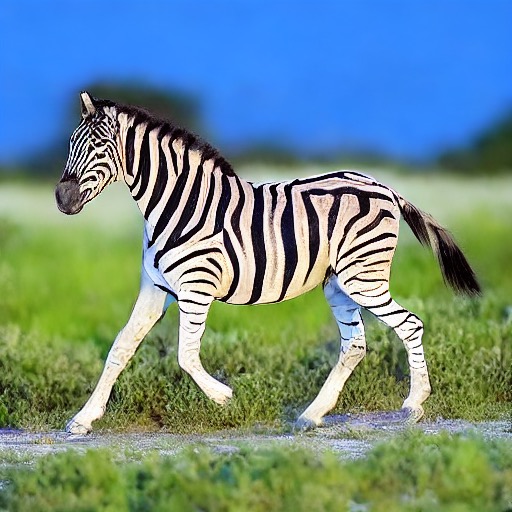} \\

        Structure & Appearance & \multicolumn{3}{c}{$\longleftarrow$ Outputs $\longrightarrow$}
        
    \end{tabular}
    }
    \vspace{-0.2cm}
    \caption{
    Limitations. Our method may struggle to transfer appearance between objects that do not share strong semantics (first two rows). Moreover, the quality of our transfer relies on the quality and editability of the inversion and may vary depending on the random seed used for the DDPM inversion (bottom row).}
    \label{fig:limitations}
\end{figure}

While we have demonstrated the effectiveness of our cross-image attention mechanism for zero-shot appearance transfer, several limitations should be considered. 
First, our method relies on the ability of the generative model to establish accurate correspondences between subjects in the two input images. As a result, transferring appearance between subjects in the images that do not share semantics (e.g., belong to different domains) can be more challenging, see the first two rows of~\Cref{fig:limitations}.
Next, our method relies on inverting the input structure and appearance images into the latent space of the image diffusion model. In cases where the inversion fails to reconstruct the input or inverts the images into less editable latent codes, our transfer introduces unwanted artifacts. Specifically, the inversion method used in our approach may exhibit sensitivity to the random seed employed for the inversion, as evident in the bottom row of~\Cref{fig:limitations} where the leg of the output may vary between random seeds. Achieving accurate, yet highly-editable inversions within the context of diffusion models remains an open problem. We believe that additional progress in this area will contribute to improved performance in downstream tasks such as appearance transfer. 

\section{Conclusions}
We have introduced a novel zero-shot approach that enables semantic-based appearance transfer between objects found in natural images. Importantly, our method demonstrates that this transfer is possible without requiring any model training or user-provided conditioning. Furthermore, this transfer can be achieved even when the objects vary in shape, size, or viewpoint. 
After examining the components of the self-attention layers --- the queries, keys, and values --- we introduced the Cross-Image Attention layer. This layer implicitly establishes semantic correspondences between objects by mixing the queries, keys, and values corresponding to two \textit{different} images. 
We then introduced three extensions to reduce the domain gap caused by our mixing operation, accomplished through the manipulation of the noised latent codes and the internal representations of the denoising model.
By leveraging the iterative denoising process, our method attains a \textit{gradual} appearance transfer, encouraging the generation of more realistic, high-quality images. 

We hope that our work encourages further exploration into the semantics of the internal representations within these powerful generative models. We believe that a deeper understanding of these representations can enable their utilization in addressing a diverse set of generative tasks with minimal user intervention while functioning in a zero-shot manner.

\section*{Acknowledgements}
We would like to thank Michael Cohen and Lior Shapiro for their support and Amir Hertz, Dani Lischinski, Gal Metzer, Rinon Gal, and Yael Vinker for their insightful feedback. This work was funded by a research gift from Meta.

{
    \small
    \bibliographystyle{ieeenat_fullname}
    \bibliography{main}
}

\clearpage
\appendix
\appendixpage

\section{Additional Details}~\label{sec:additional_details}

\paragraph{Implementation Details}
We operate over Stable Diffusion v1.5 text-to-image model~\cite{rombach2022high}. To invert the two input images, we apply the DDPM inversion technique introduced in~\cite{huberman2023edit} using their default hyperparameters and using the prompt ``A photo of a \textit{domain}'' where \textit{domain} denotes the domain of the object we wish to transfer (e.g., animal or building). For the denoising process, we employ the standard DDIM scheduler introduced by Song~\etal~\cite{song2021denoising} for $100$ denoising steps. 

For appearance transfer, we replace the conventional self-attention layers within the denoising network's decoder at resolutions of $32\times 32$ and $64\times 64$ with our cross-image attention layers. 
However, we inject the keys and values only at a subset of the denoising timesteps. Specifically, for layers with a resolution of $32\times 32$, the injection is performed between timesteps $10$ and $70$, while for layers at a resolution of $64\times 64$ the injection is applied between timesteps $10$ and $90$.
At all other timesteps, our cross-image attention layer functions identically to the standard self-attention layer.

Additionally, we apply a contrast strength of $\beta=1.67$ over the intermediate cross-image attention maps. For our appearance guidance, we set the guidance scale to $\alpha=3.5$ and apply the AdaIN operation between the style and output noise latents between timesteps $20$ and $100$.

To compute the object masks used for the AdaIN operation, we use the unsupervised self-segmentation technique introduced in Patashnik~\etal~\cite{patashnik2023localizing} using the domain name as the guiding noun. 
Finally, we apply the FreeU technique~\cite{si2023freeu} over Stable Diffusion and find that doing so leads to fewer artifacts in the generated images.

\paragraph{Structure Injection}
Lastly, we explore a simple technique that we find helps to better preserve the original structure in $I^{struct}$ for certain object domains. Instead of replacing the keys and values corresponding to $z_{t}^{out}$ with those of $z_{t}^{app}$, we choose specific intervals where we replace $K_{out}$ and $V_{out}$ with the keys and values derived from $z_{t}^{struct}$. That is, the feature output at these timesteps is now defined as
\begin{equation}
    \text{softmax} \left ( \frac{Q_{out} \cdot K^T_{struct}}{\sqrt{d}} \right ) \cdot V_{struct}.
\end{equation}
We observe that this approach is effective for object categories containing finer-level structural details such as the ear of an animal. 
We find that performing this structure injection every five timesteps provides a favorable balance between faithfully transferring the target appearance to the output image while maintaining its original structure.

\paragraph{User Study}
Since Swapping Autoencoder~\cite{park2020swapping} is limited to the buildings domain, we select eight structure-appearance pairs from the building domain and two pairs for the four other domains (animals, cars, cakes, and birds). This results in a total of $16$ input pairs in total. Note that Swapping Autoencoder was not evaluated with respect to the four other domains as no trained models exist for these domains.

\begin{figure*}
    \centering
    \setlength{\tabcolsep}{0.3pt}
    \renewcommand{\arraystretch}{0.3}
    \addtolength{\belowcaptionskip}{-5pt}
    {

    \begin{minipage}{0.5\textwidth}
        \centering
        \begin{tabular}{c c c c}
    
            \includegraphics[width=0.22\textwidth]{images/struct_app.png} &
            \includegraphics[width=0.22\textwidth]{images/inputs/buildings/chile-church.jpg} &
            \includegraphics[width=0.22\textwidth]{images/inputs/buildings/st_anthony_of_padua.jpg} &
            \includegraphics[width=0.22\textwidth]{images/inputs/buildings/sagrada_familia.jpg} \\
    
            \includegraphics[width=0.22\textwidth]{images/inputs/buildings/taj_mahal.jpg} &
            \includegraphics[width=0.22\textwidth]{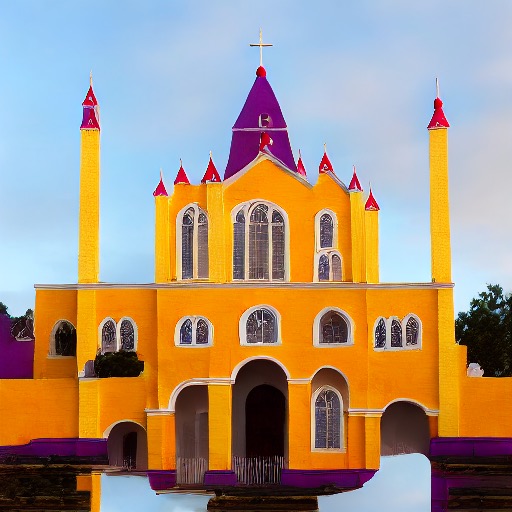} &
            \includegraphics[width=0.22\textwidth]{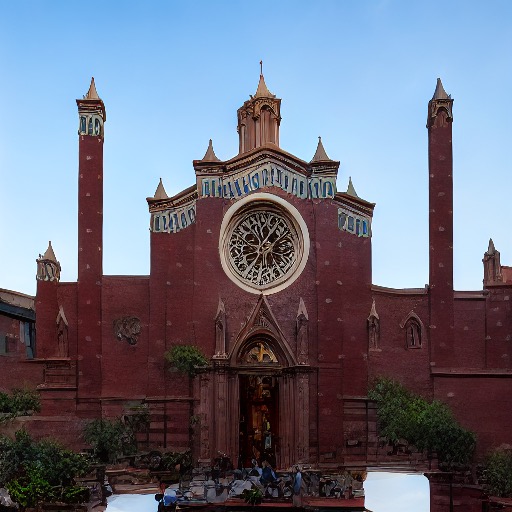} &
            \includegraphics[width=0.22\textwidth]{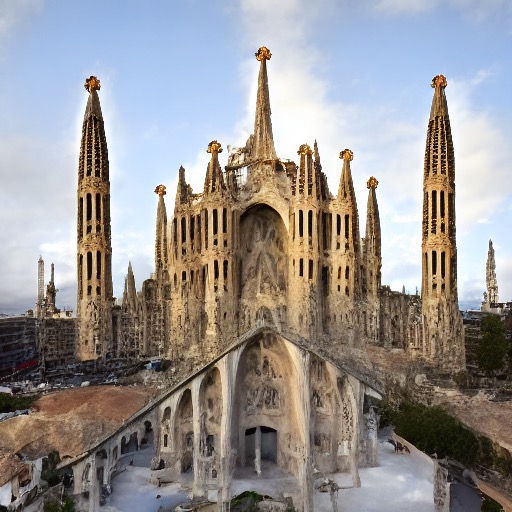} \\

        \end{tabular}
    \end{minipage}%
    \begin{minipage}{0.5\textwidth}
        \centering
        \begin{tabular}{c c c c}
    
            \includegraphics[width=0.22\textwidth]{images/struct_app.png} &
            \includegraphics[width=0.22\textwidth]{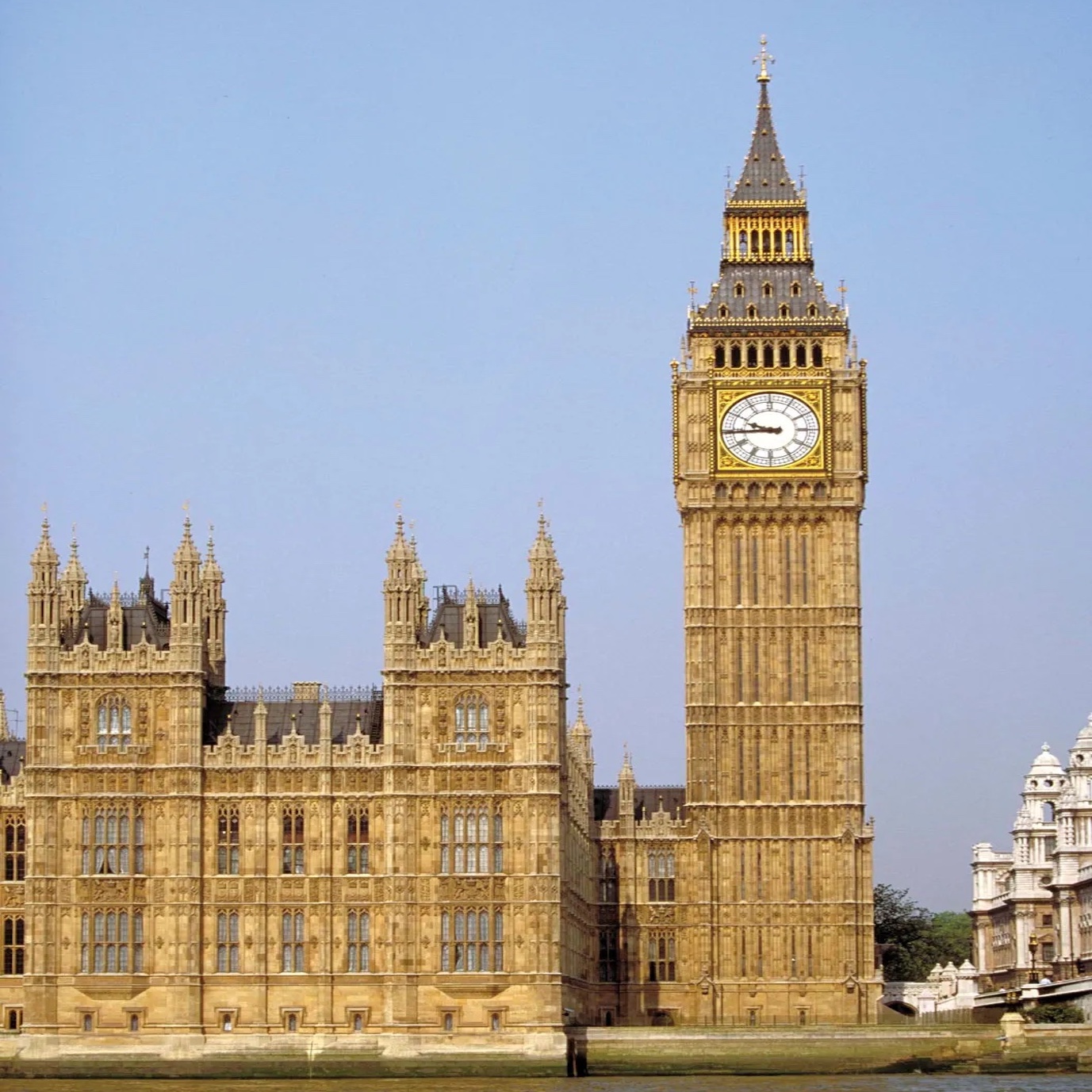} &
            \includegraphics[width=0.22\textwidth]{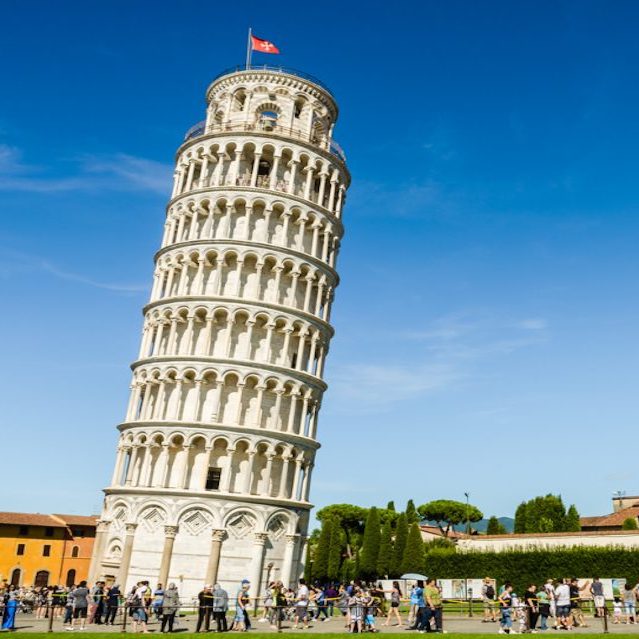} &
            \includegraphics[width=0.22\textwidth]{images/inputs/buildings/sagrada_familia.jpg} \\
    
            \includegraphics[width=0.22\textwidth]{images/inputs/buildings/saint_basil.jpg} &
            \includegraphics[width=0.22\textwidth]{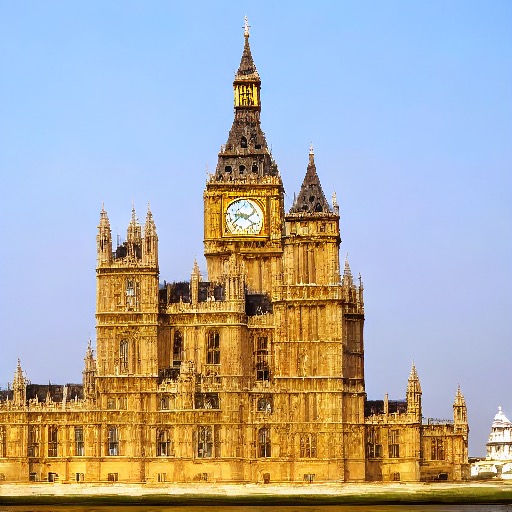} &
            \includegraphics[width=0.22\textwidth]{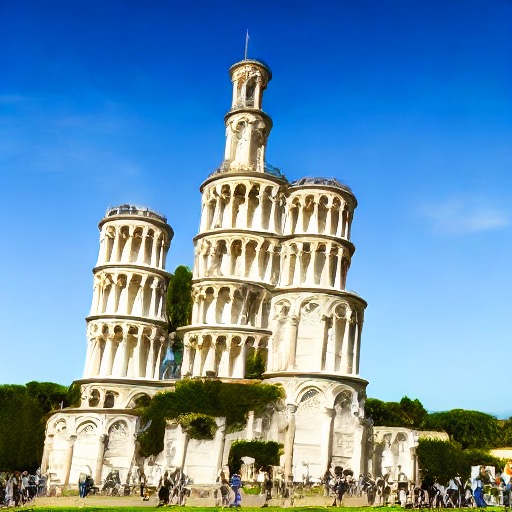} &
            \includegraphics[width=0.22\textwidth]{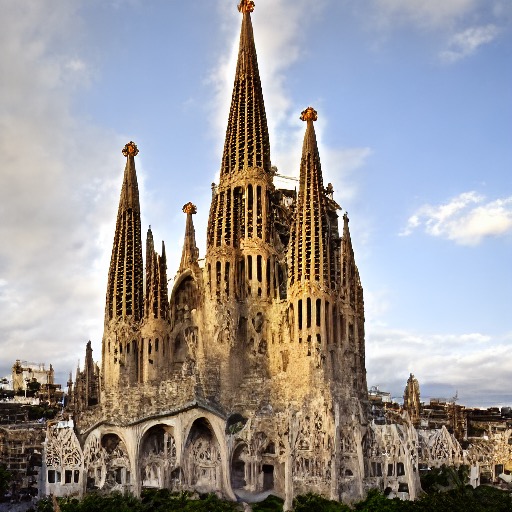} \\

        \\ \\
    
        \end{tabular}
        
    \end{minipage}%

    \begin{minipage}{0.5\textwidth}
        \centering
        \begin{tabular}{c c c c}
    
            \includegraphics[width=0.22\textwidth]{images/struct_app.png} &
            \includegraphics[width=0.22\textwidth]{images/inputs/buildings/big_ben.jpg} &
            \includegraphics[width=0.22\textwidth]{images/inputs/buildings/st_anthony_of_padua.jpg} &
            \includegraphics[width=0.22\textwidth]{images/inputs/buildings/sagrada_familia.jpg} \\
    
            \includegraphics[width=0.22\textwidth]{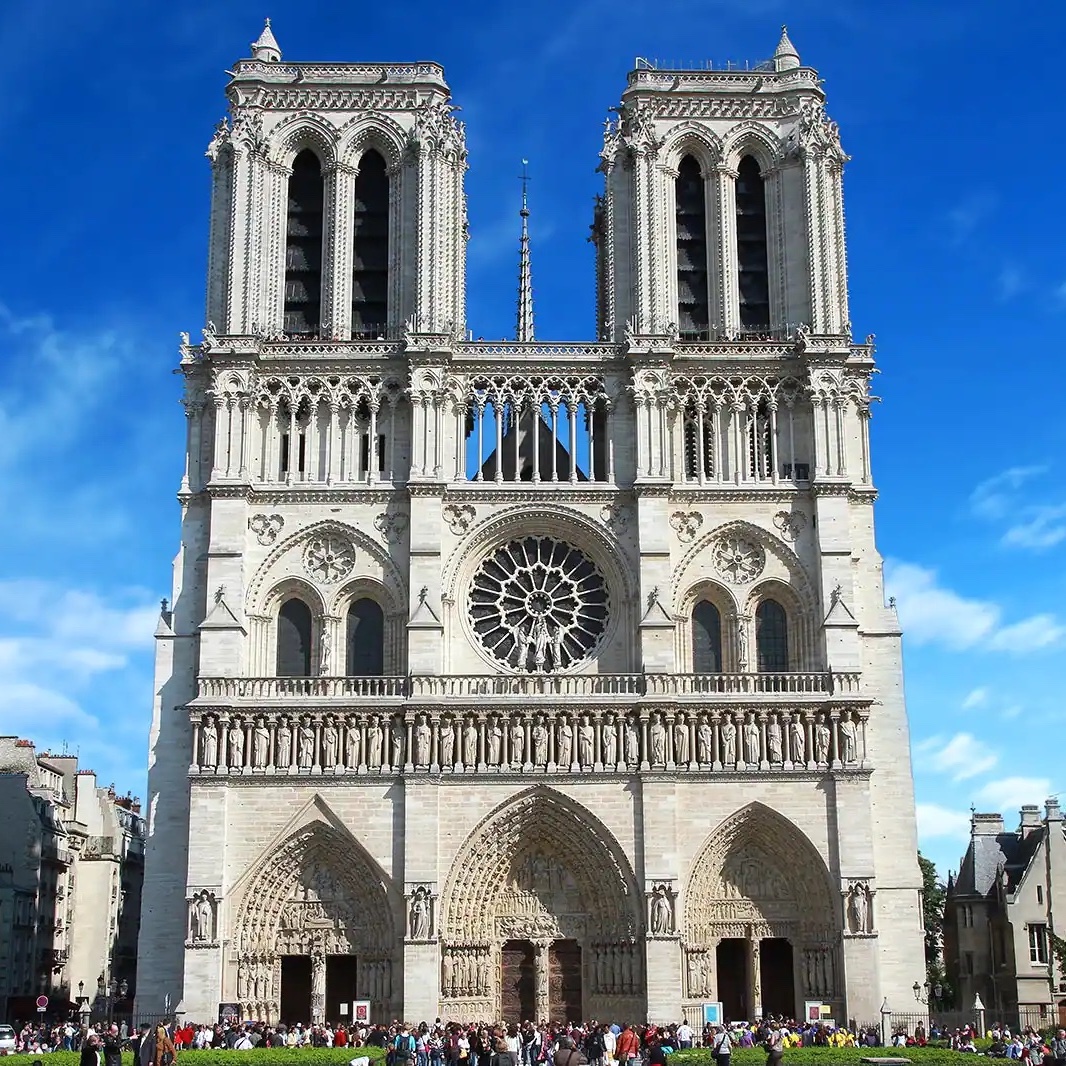} &
            \includegraphics[width=0.22\textwidth]{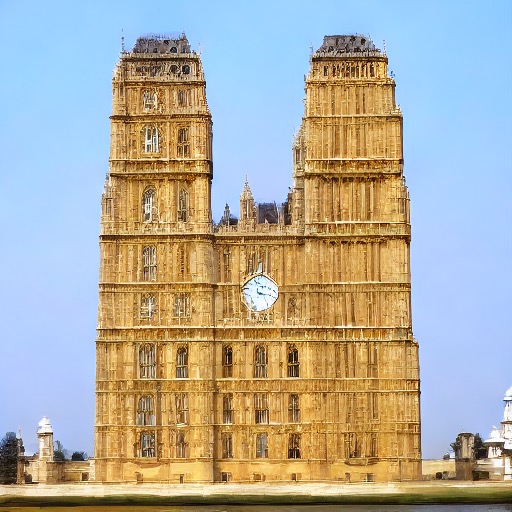} &
            \includegraphics[width=0.22\textwidth]{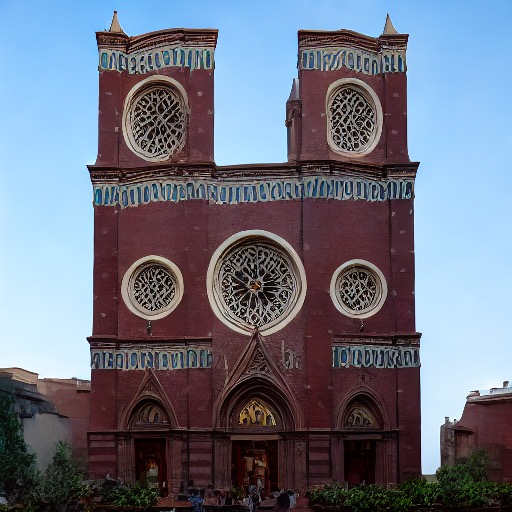} &
            \includegraphics[width=0.22\textwidth]{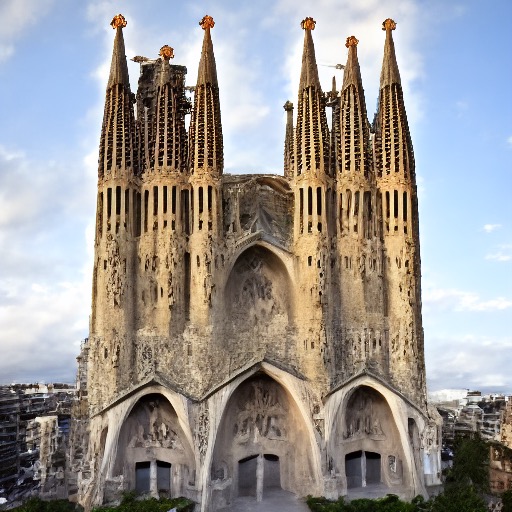} \\

        \end{tabular}
    \end{minipage}%
    \begin{minipage}{0.5\textwidth}
        \centering
        \begin{tabular}{c c c c}
    
            \includegraphics[width=0.22\textwidth]{images/struct_app.png} &
            \includegraphics[width=0.22\textwidth]{images/inputs/buildings/big_ben.jpg} &
            \includegraphics[width=0.22\textwidth]{images/inputs/buildings/taj_mahal.jpg} &
            \includegraphics[width=0.22\textwidth]{images/inputs/buildings/sagrada_familia.jpg} \\
    
            \includegraphics[width=0.22\textwidth]{images/inputs/buildings/chile-church.jpg} &
            \includegraphics[width=0.22\textwidth]{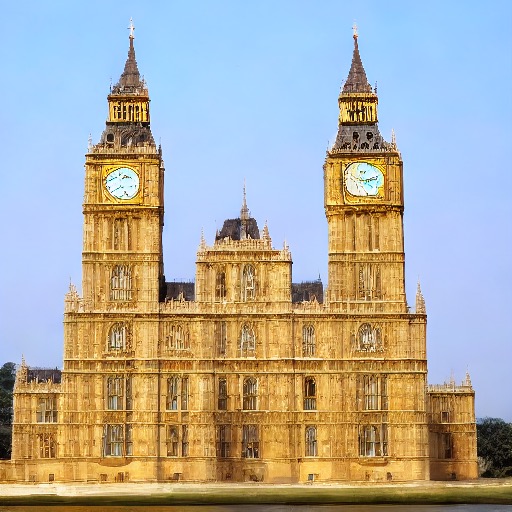} &
            \includegraphics[width=0.22\textwidth]{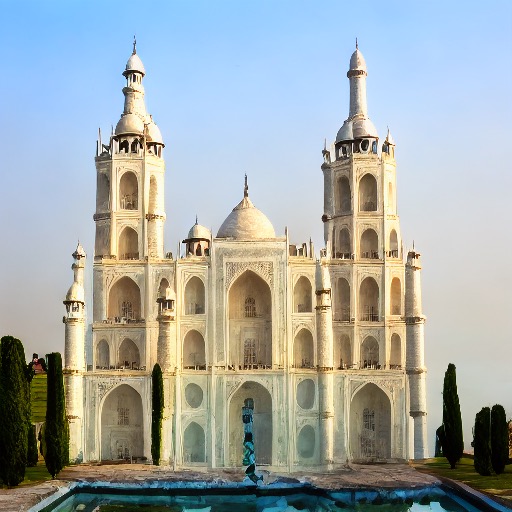} &
            \includegraphics[width=0.22\textwidth]{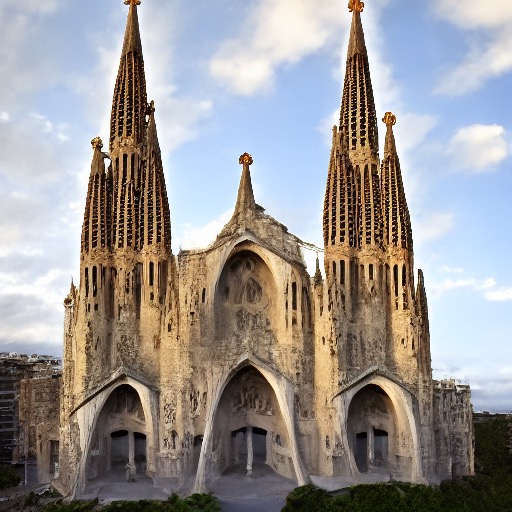} \\

        \\ \\
    
        \end{tabular}
        
    \end{minipage}%

    \begin{minipage}{0.5\textwidth}
        \centering
        \begin{tabular}{c c c c}
    
            \includegraphics[width=0.22\textwidth]{images/struct_app.png} &
            \includegraphics[width=0.22\textwidth]{images/inputs/cars/vintage.jpg} &
            \includegraphics[width=0.22\textwidth]{images/inputs/cars/red_vintage.jpg} &
            \includegraphics[width=0.22\textwidth]{images/inputs/cars/wolksvagen_beetle_mirror_mirror.jpg} \\
    
            \includegraphics[width=0.22\textwidth]{images/inputs/vehicles/red_truck.jpg} &
            \includegraphics[width=0.22\textwidth]{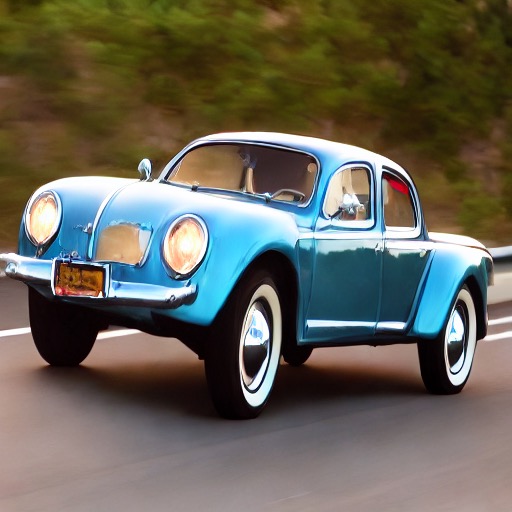} &
            \includegraphics[width=0.22\textwidth]{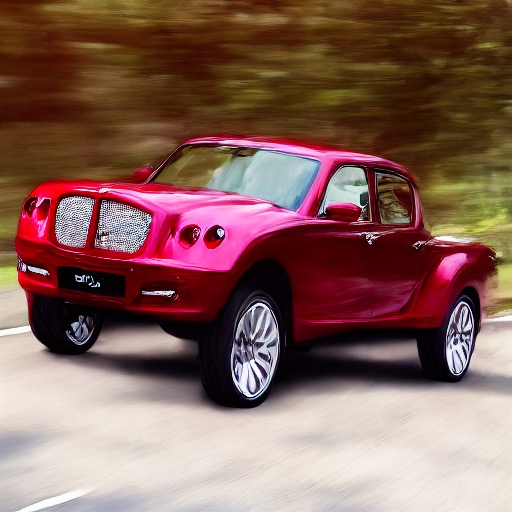} &
            \includegraphics[width=0.22\textwidth]{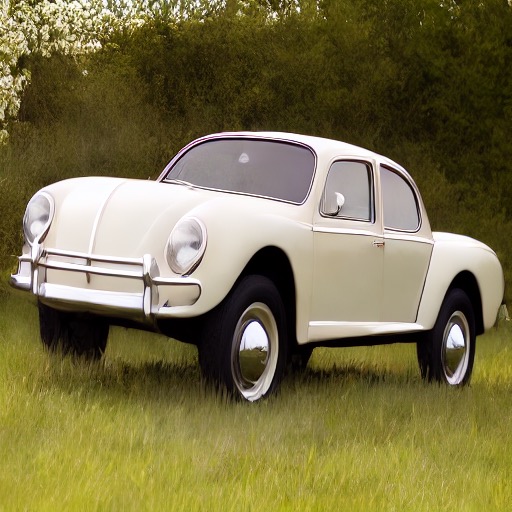} \\

        \end{tabular}
    \end{minipage}%
    \begin{minipage}{0.5\textwidth}
        \centering
        \begin{tabular}{c c c c}
    
            \includegraphics[width=0.22\textwidth]{images/struct_app.png} &
            \includegraphics[width=0.22\textwidth]{images/inputs/cars/vintage.jpg} &
            \includegraphics[width=0.22\textwidth]{images/inputs/cars/red_vintage.jpg} &
            \includegraphics[width=0.22\textwidth]{images/inputs/cars/wolksvagen_beetle_mirror_mirror.jpg} \\
    
            \includegraphics[width=0.22\textwidth]{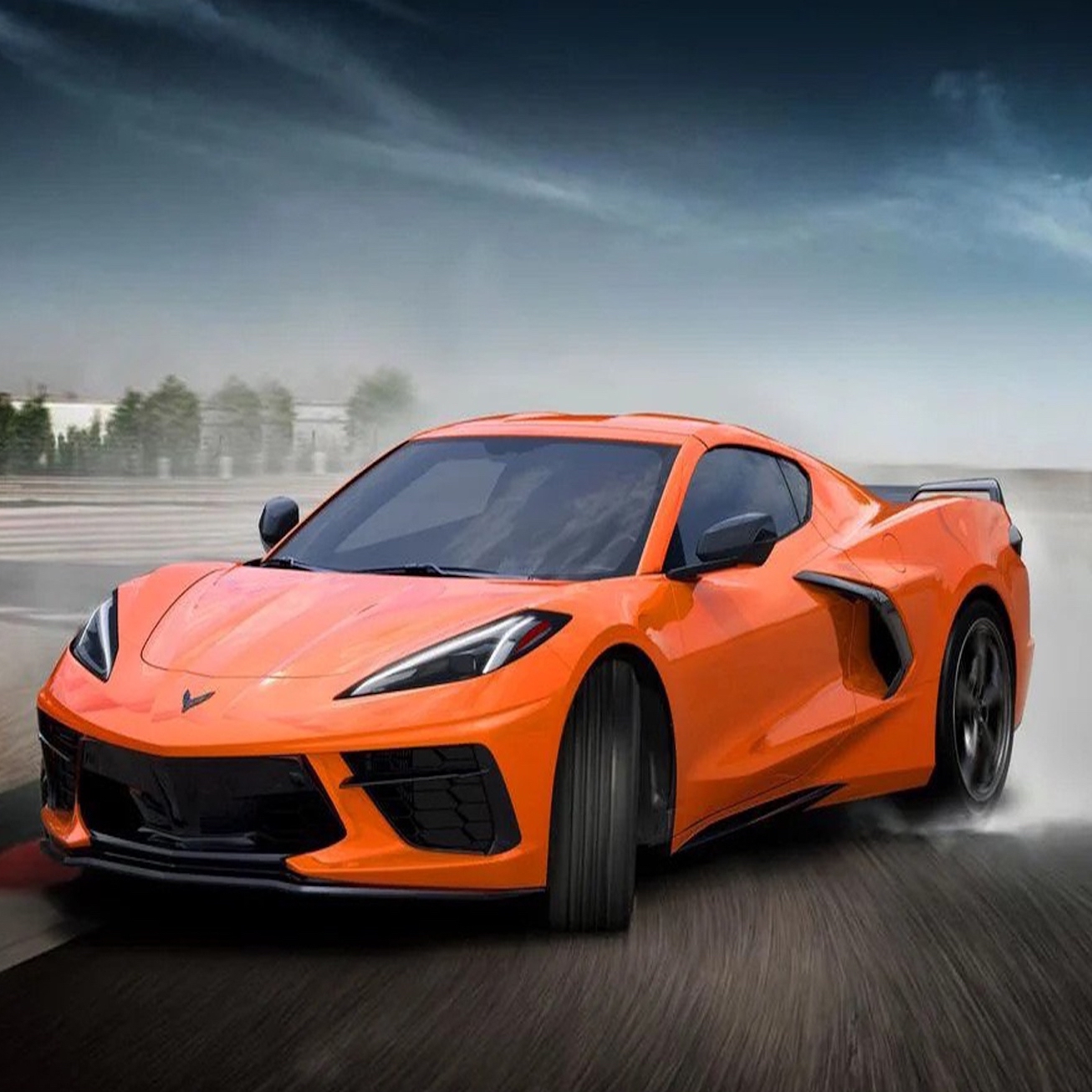} &
            \includegraphics[width=0.22\textwidth]{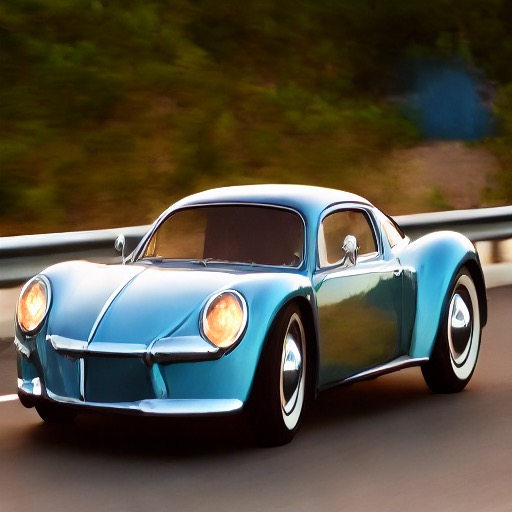} &
            \includegraphics[width=0.22\textwidth]{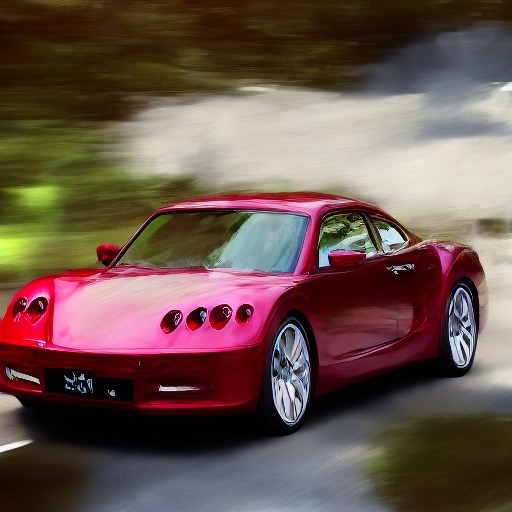} &
            \includegraphics[width=0.22\textwidth]{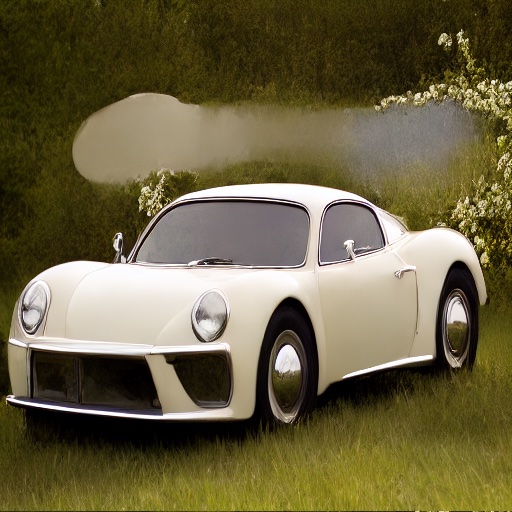} \\
    
        \end{tabular}
        
    \end{minipage}%

    \begin{minipage}{0.5\textwidth}
        \centering
        \begin{tabular}{c c c c}
    
            \includegraphics[width=0.22\textwidth]{images/struct_app.png} &
            \includegraphics[width=0.22\textwidth]{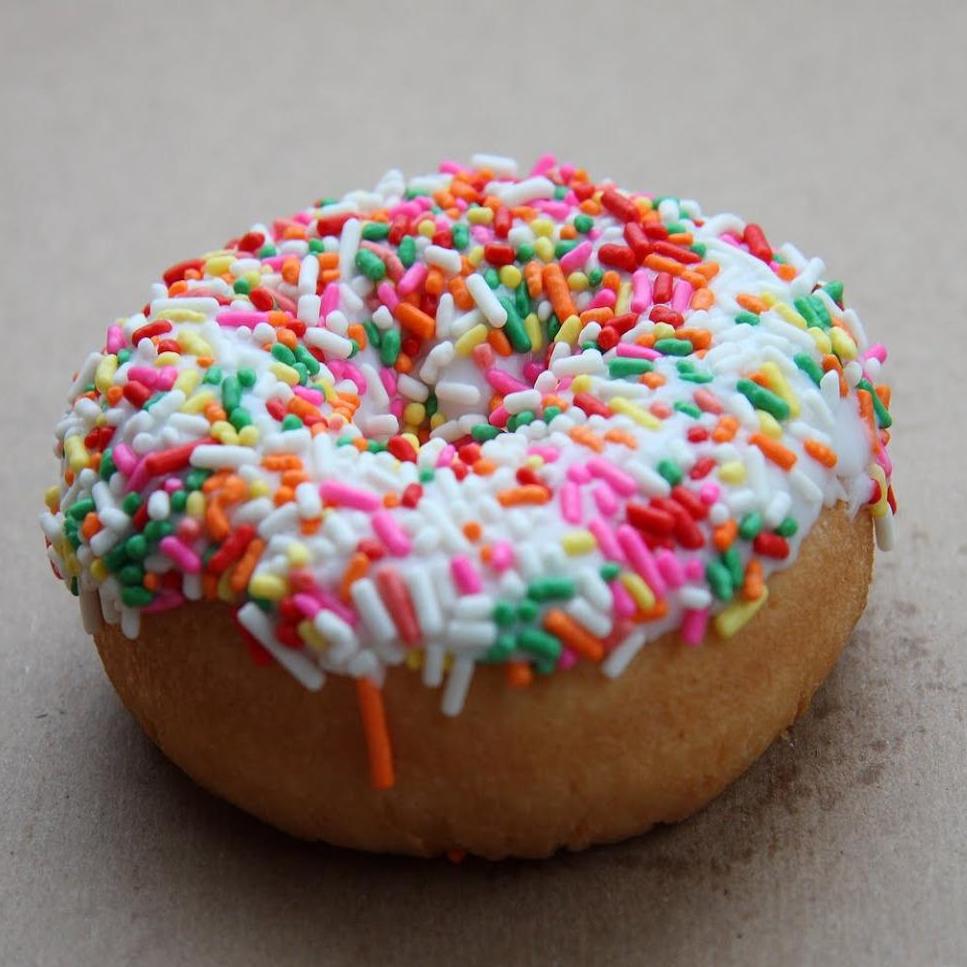} &
            \includegraphics[width=0.22\textwidth]{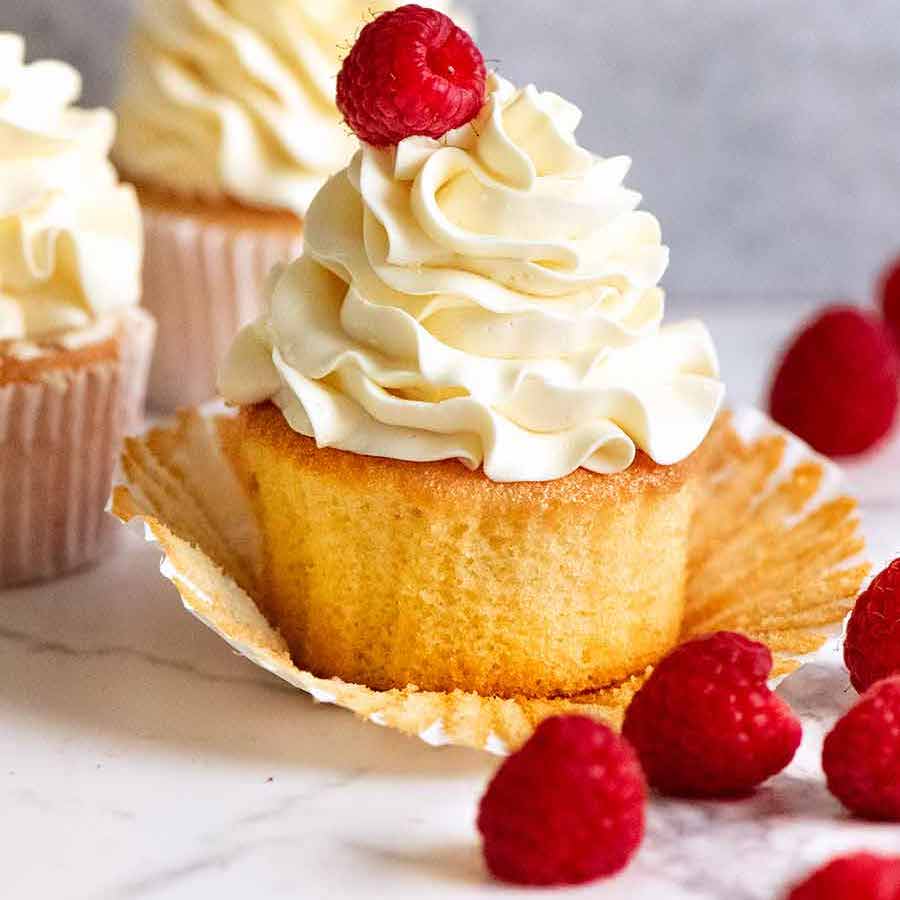} &
            \includegraphics[width=0.22\textwidth]{images/inputs/cake/strawberry.jpg} \\
    
            \includegraphics[width=0.22\textwidth]{images/inputs/cake/red_velvet.jpg} &
            \includegraphics[width=0.22\textwidth]{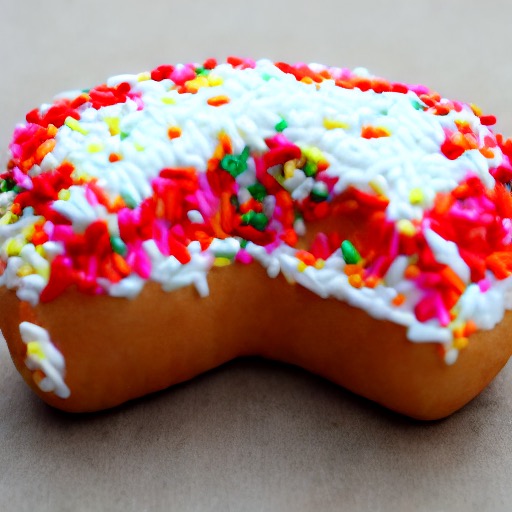} &
            \includegraphics[width=0.22\textwidth]{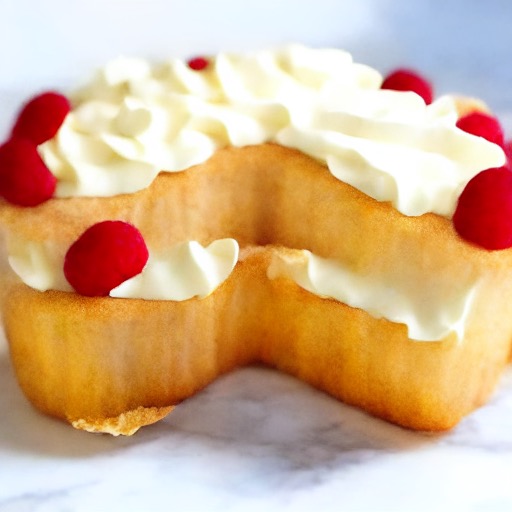} &
            \includegraphics[width=0.22\textwidth]{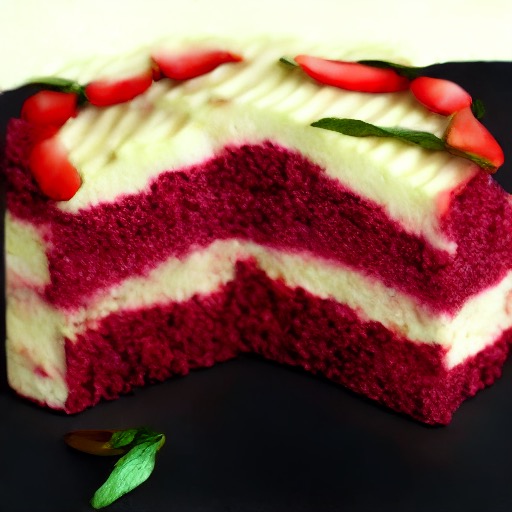} \\

        \end{tabular}
    \end{minipage}%
    \begin{minipage}{0.5\textwidth}
        \centering
        \begin{tabular}{c c c c}
    
            \includegraphics[width=0.22\textwidth]{images/struct_app.png} &
            \includegraphics[width=0.22\textwidth]{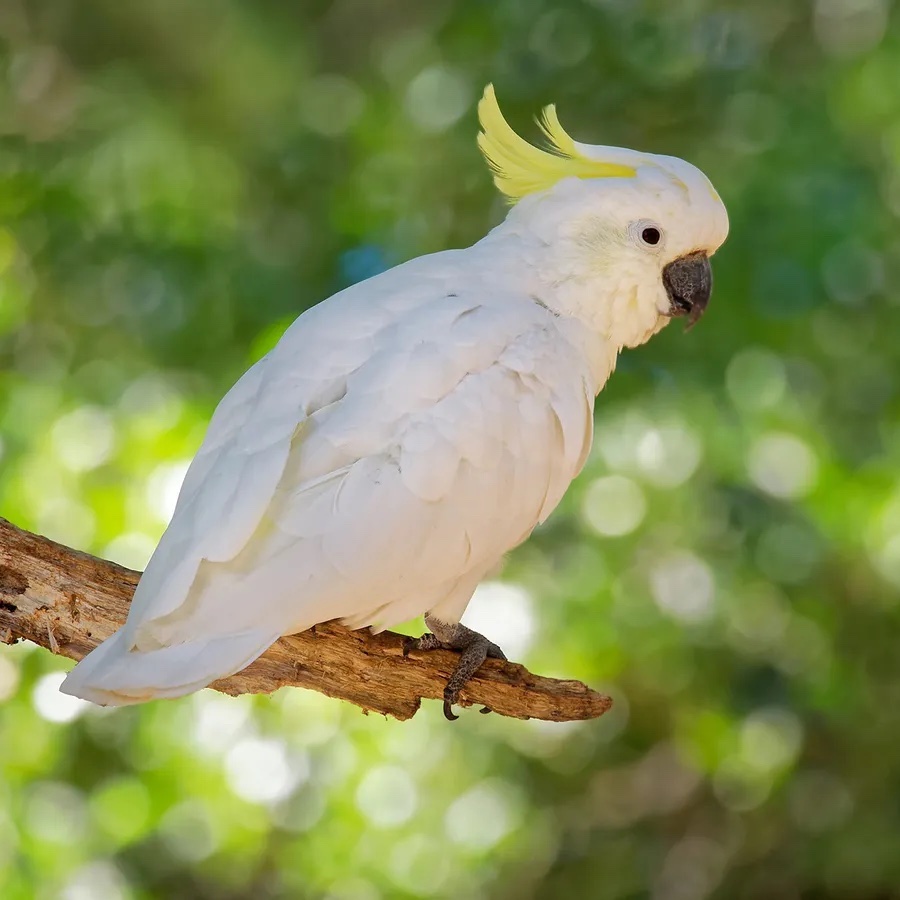} &
            \includegraphics[width=0.22\textwidth]{images/inputs/birds/hummingbird.jpg} &
            \includegraphics[width=0.22\textwidth]{images/inputs/birds/lilac_roller.jpg} \\
    
            \includegraphics[width=0.22\textwidth]{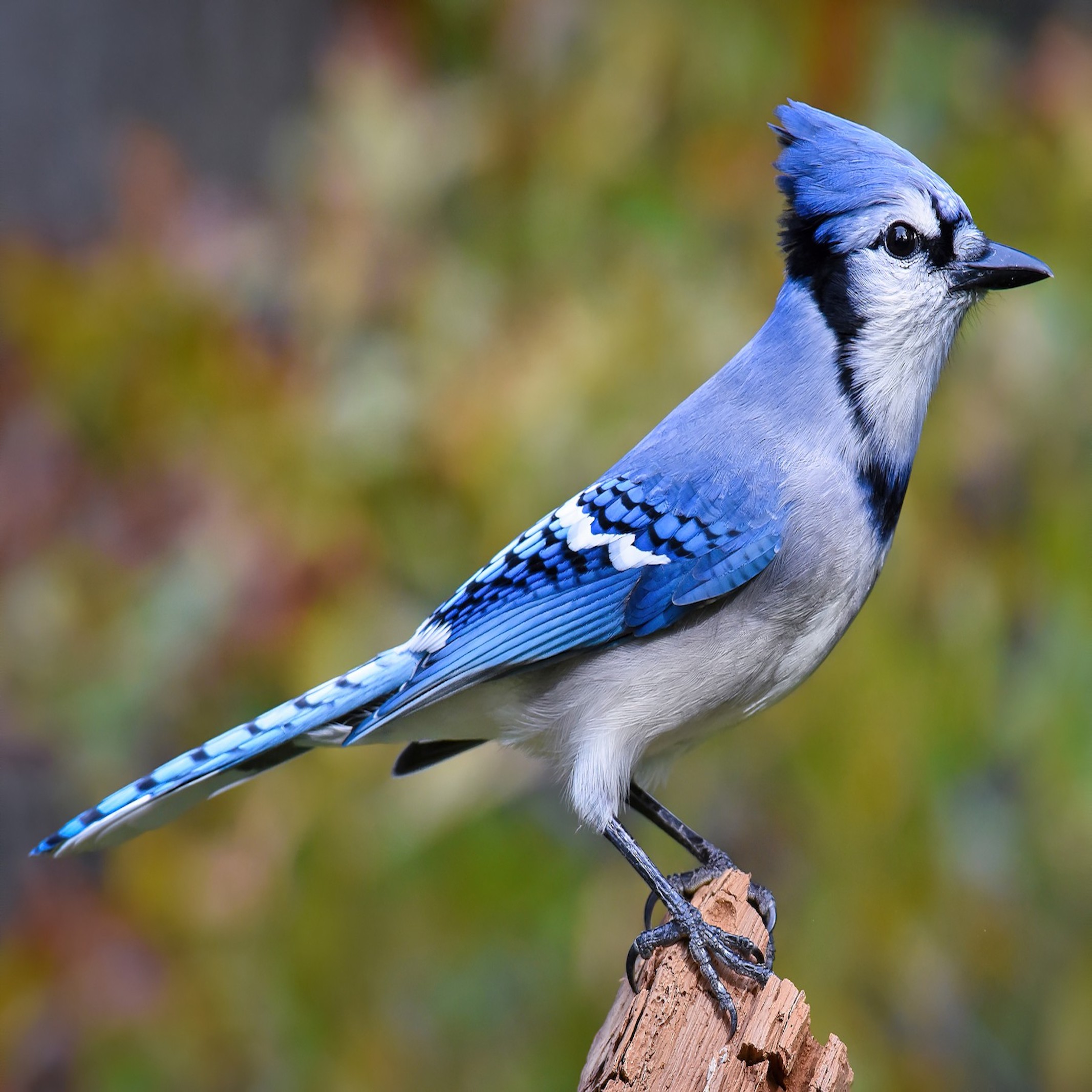} &
            \includegraphics[width=0.22\textwidth]{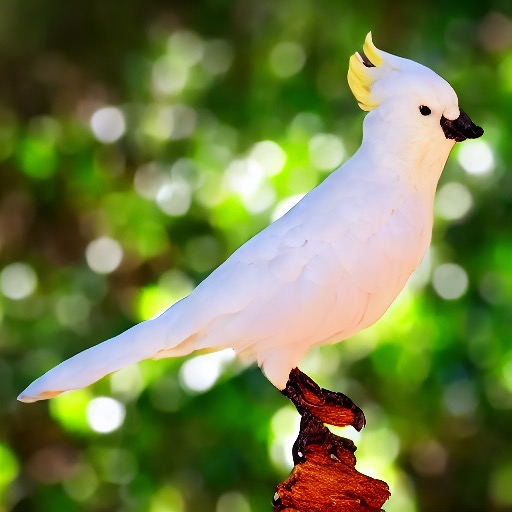} &
            \includegraphics[width=0.22\textwidth]{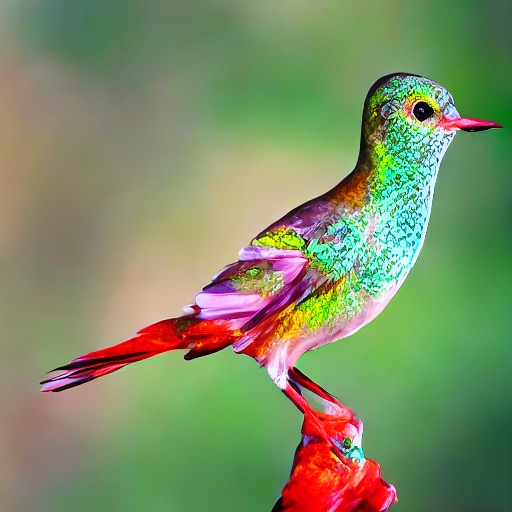} &
            \includegraphics[width=0.22\textwidth]{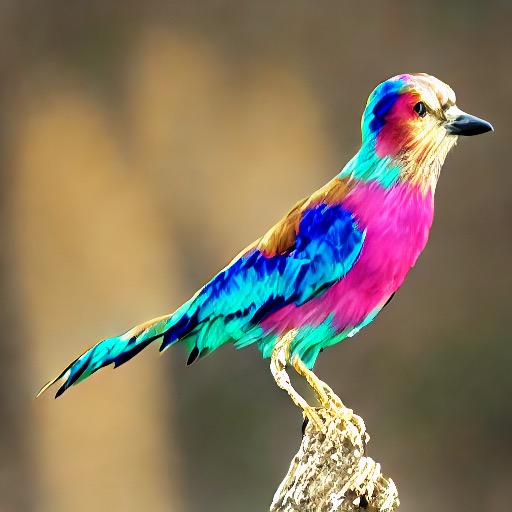} \\

        \\ \\
    
        \end{tabular}
    \end{minipage}

    \begin{minipage}{0.5\textwidth}
        \centering
        \begin{tabular}{c c c c}
    
            \includegraphics[width=0.22\textwidth]{images/struct_app.png} &
            \includegraphics[width=0.22\textwidth]{images/inputs/house/stone_house.jpg} &
            \includegraphics[width=0.22\textwidth]{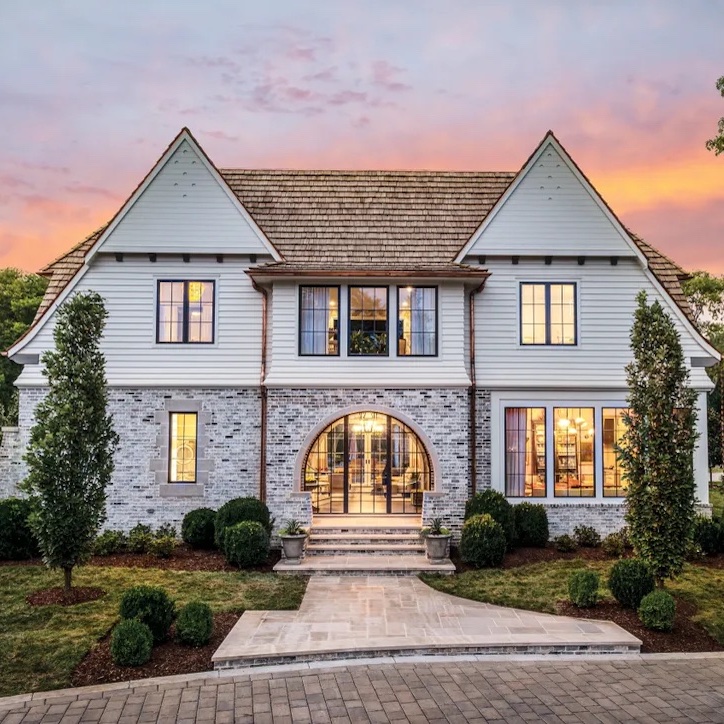} &
            \includegraphics[width=0.22\textwidth]{images/inputs/house/blue_beach_house.jpg} \\
    
            \includegraphics[width=0.22\textwidth]{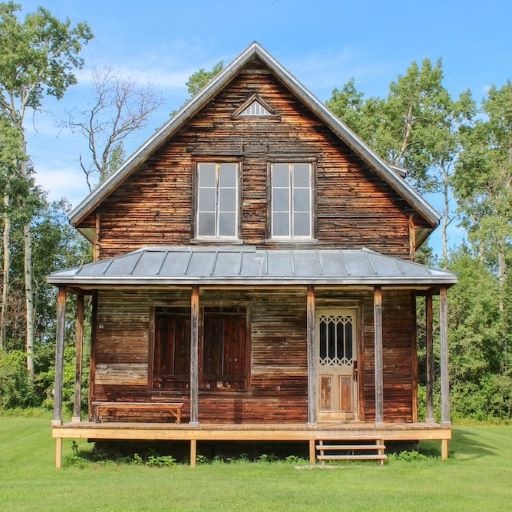} &
            \includegraphics[width=0.22\textwidth]{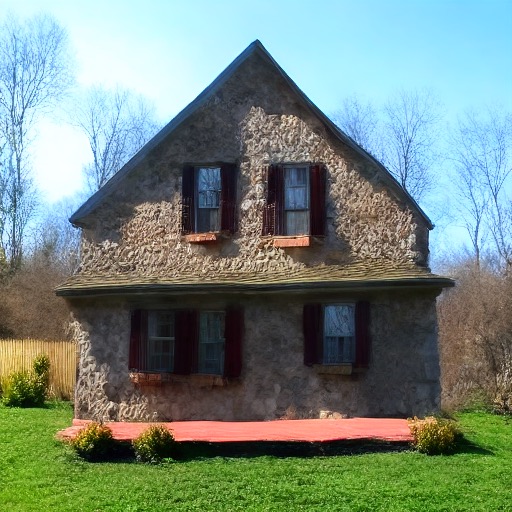} &
            \includegraphics[width=0.22\textwidth]{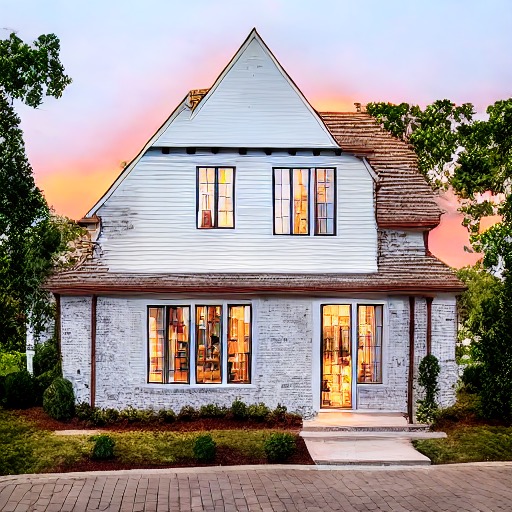} &
            \includegraphics[width=0.22\textwidth]{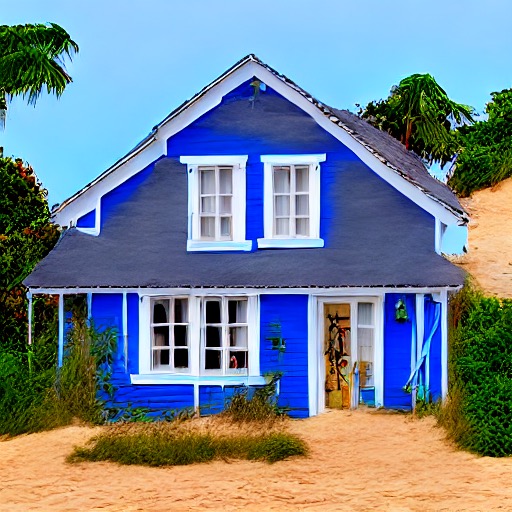} \\
    
        \end{tabular}
    \end{minipage}%
    \begin{minipage}{0.5\textwidth}
        \centering
        \begin{tabular}{c c c c}
    
            \includegraphics[width=0.22\textwidth]{images/struct_app.png} &
            \includegraphics[width=0.22\textwidth]{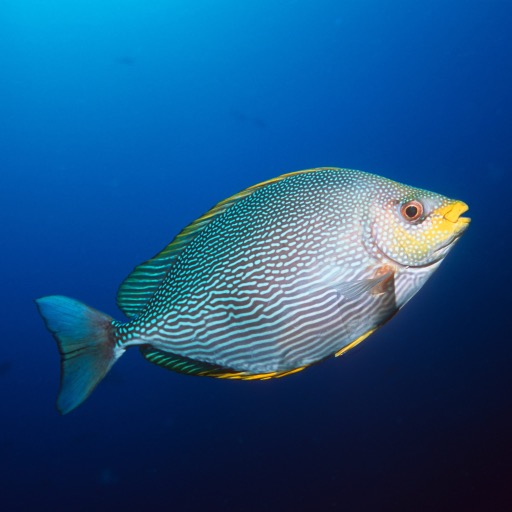} &
            \includegraphics[width=0.22\textwidth]{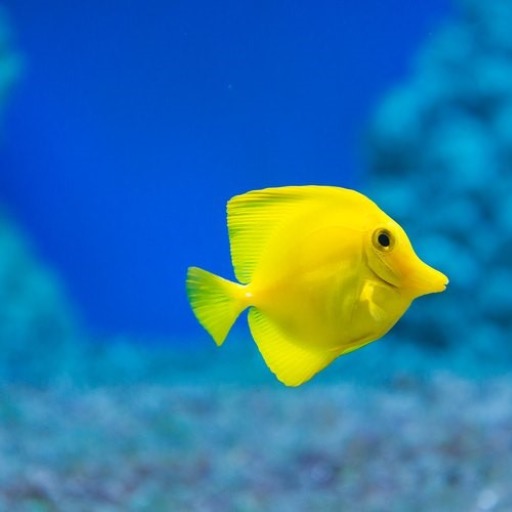} &
            \includegraphics[width=0.22\textwidth]{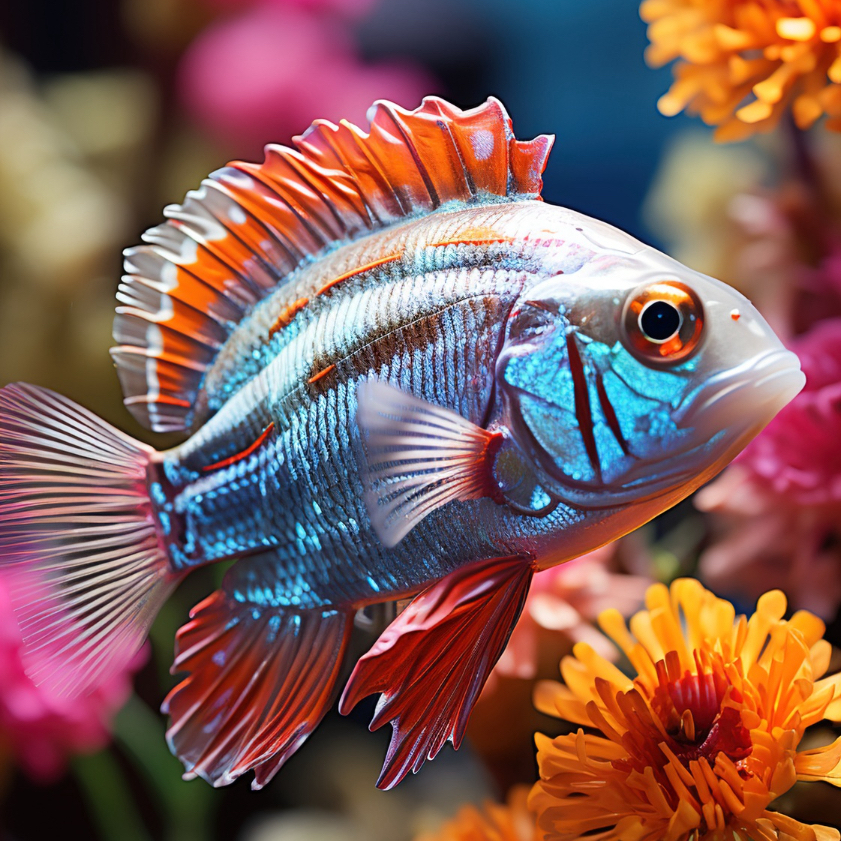} \\
    
            \includegraphics[width=0.22\textwidth]{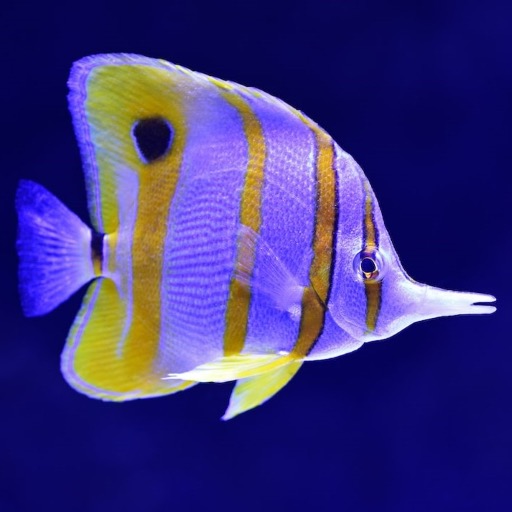} &
            \includegraphics[width=0.22\textwidth]{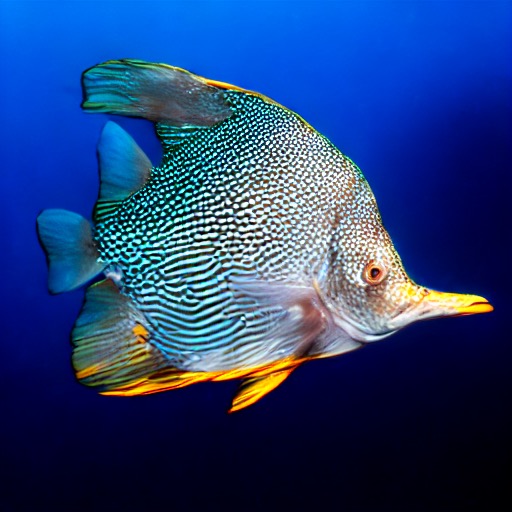} &
            \includegraphics[width=0.22\textwidth]{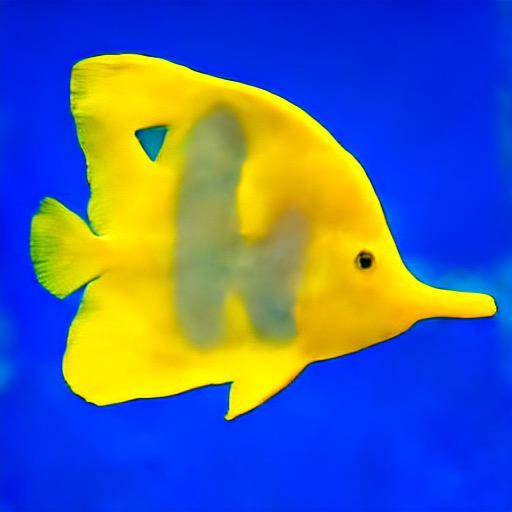} &
            \includegraphics[width=0.22\textwidth]{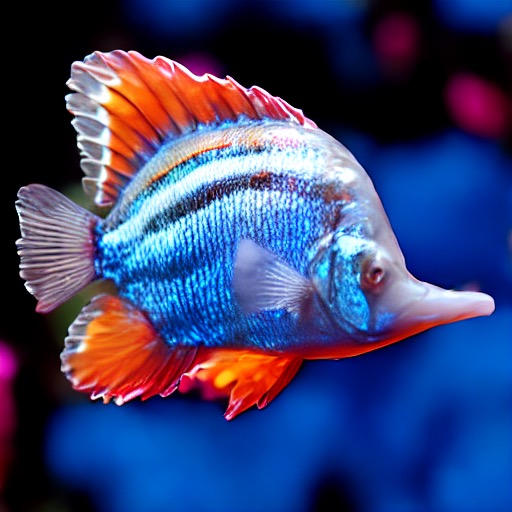} 

        \\ \\
    
        \end{tabular}
    \end{minipage}

    }
    \vspace{-0.2cm}
    \caption{
    Additional appearance transfer results obtained by our method. For each set of images, we show transfer results between a single structure image (shown to the left) and three different appearance images (shown to the top).
    }
    \label{fig:additional_multi_style}
\end{figure*}

\begin{figure*}
    \centering
    \setlength{\tabcolsep}{0.5pt}
    \addtolength{\belowcaptionskip}{-10pt}
    {
    \begin{tabular}{c c c@{\hspace{0.4cm}} c c c@{\hspace{0.4cm}} c c c}

        \includegraphics[width=0.15\textwidth]{images/inputs/buildings/pisa.jpg} &
        \includegraphics[width=0.15\textwidth]{images/inputs/buildings/saint_basil.jpg} &
        \includegraphics[width=0.15\textwidth]{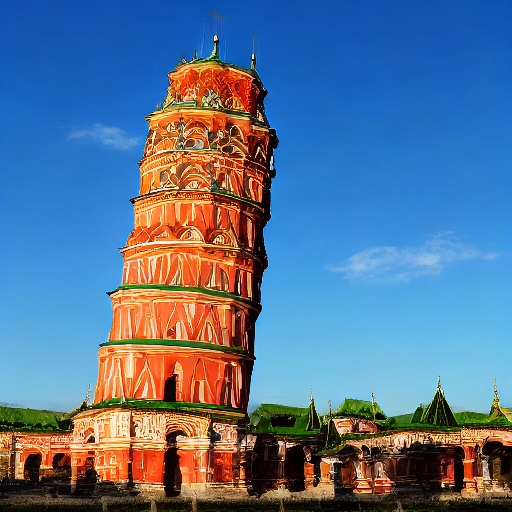} &

        \includegraphics[width=0.15\textwidth]{images/inputs/buildings/duomo.jpg} &
        \includegraphics[width=0.15\textwidth]{images/inputs/buildings/norte_dame.jpg} &
        \includegraphics[width=0.15\textwidth]{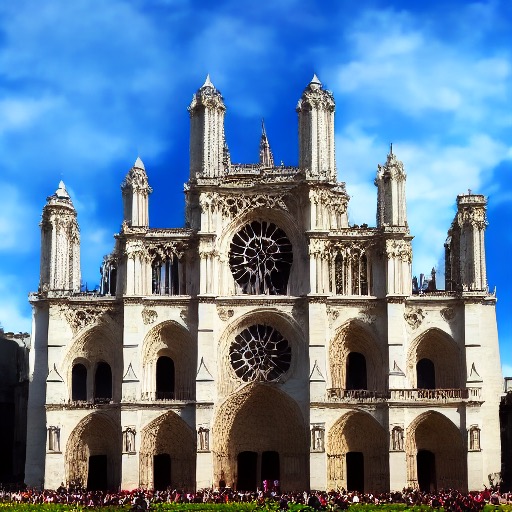} \\

        Structure & Appearance & Output & Structure & Appearance & Output \\ \\

        \includegraphics[width=0.15\textwidth]{images/inputs/buildings/eiffel_tower.jpg} &
        \includegraphics[width=0.15\textwidth]{images/inputs/buildings/big_ben.jpg} &
        \includegraphics[width=0.15\textwidth]{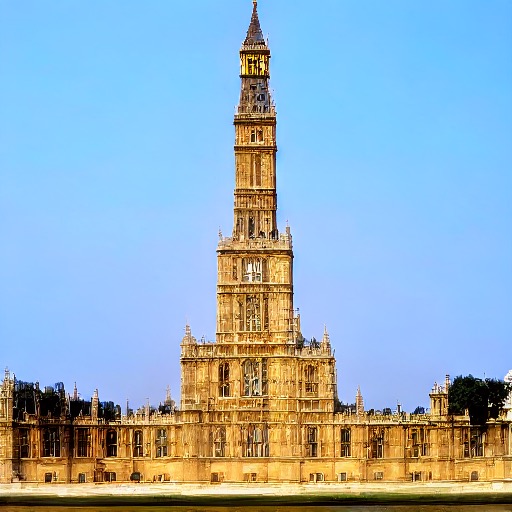} &

        \includegraphics[width=0.15\textwidth]{images/inputs/buildings/big_ben.jpg} &
        \includegraphics[width=0.15\textwidth]{images/inputs/buildings/norte_dame.jpg} &
        \includegraphics[width=0.15\textwidth]{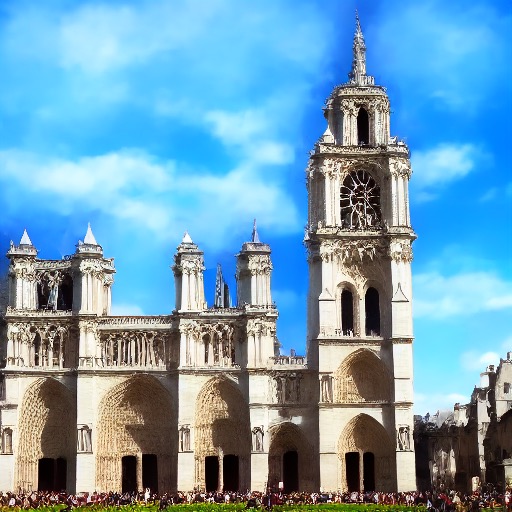} \\

        Structure & Appearance & Output & Structure & Appearance & Output \\ \\

        \includegraphics[width=0.15\textwidth]{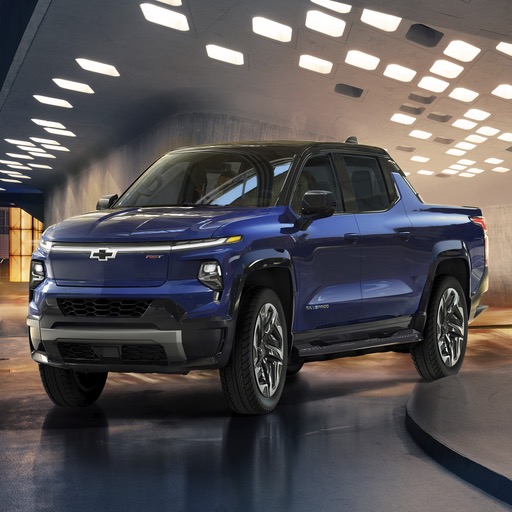} &
        \includegraphics[width=0.15\textwidth]{images/inputs/cars/red_vintage.jpg} &
        \includegraphics[width=0.15\textwidth]{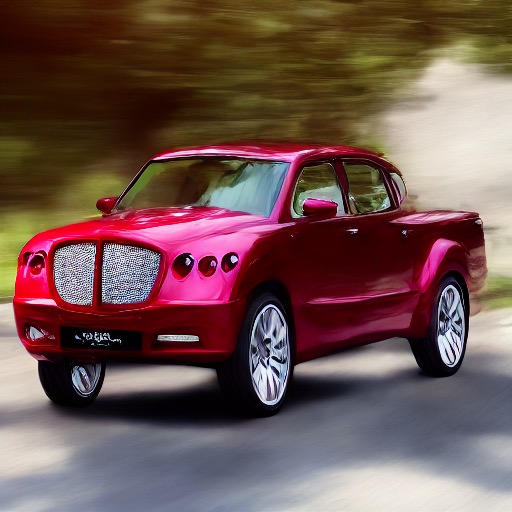} &

        \includegraphics[width=0.15\textwidth]{images/inputs/cars/red_vintage.jpg} &
        \includegraphics[width=0.15\textwidth]{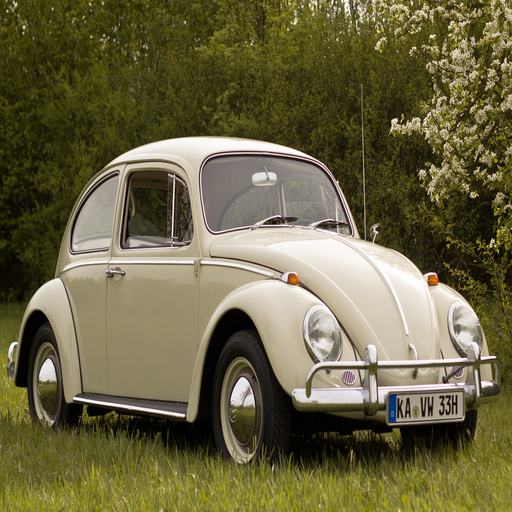} &
        \includegraphics[width=0.15\textwidth]{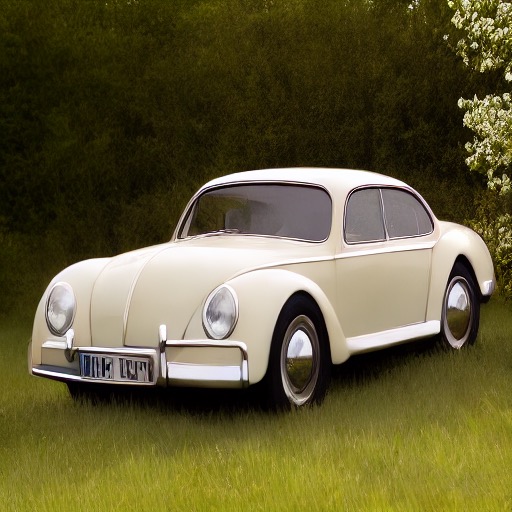} \\

        Structure & Appearance & Output & Structure & Appearance & Output \\ \\
        
        \includegraphics[width=0.15\textwidth]{images/inputs/animals/zebras.jpg} &
        \includegraphics[width=0.15\textwidth]{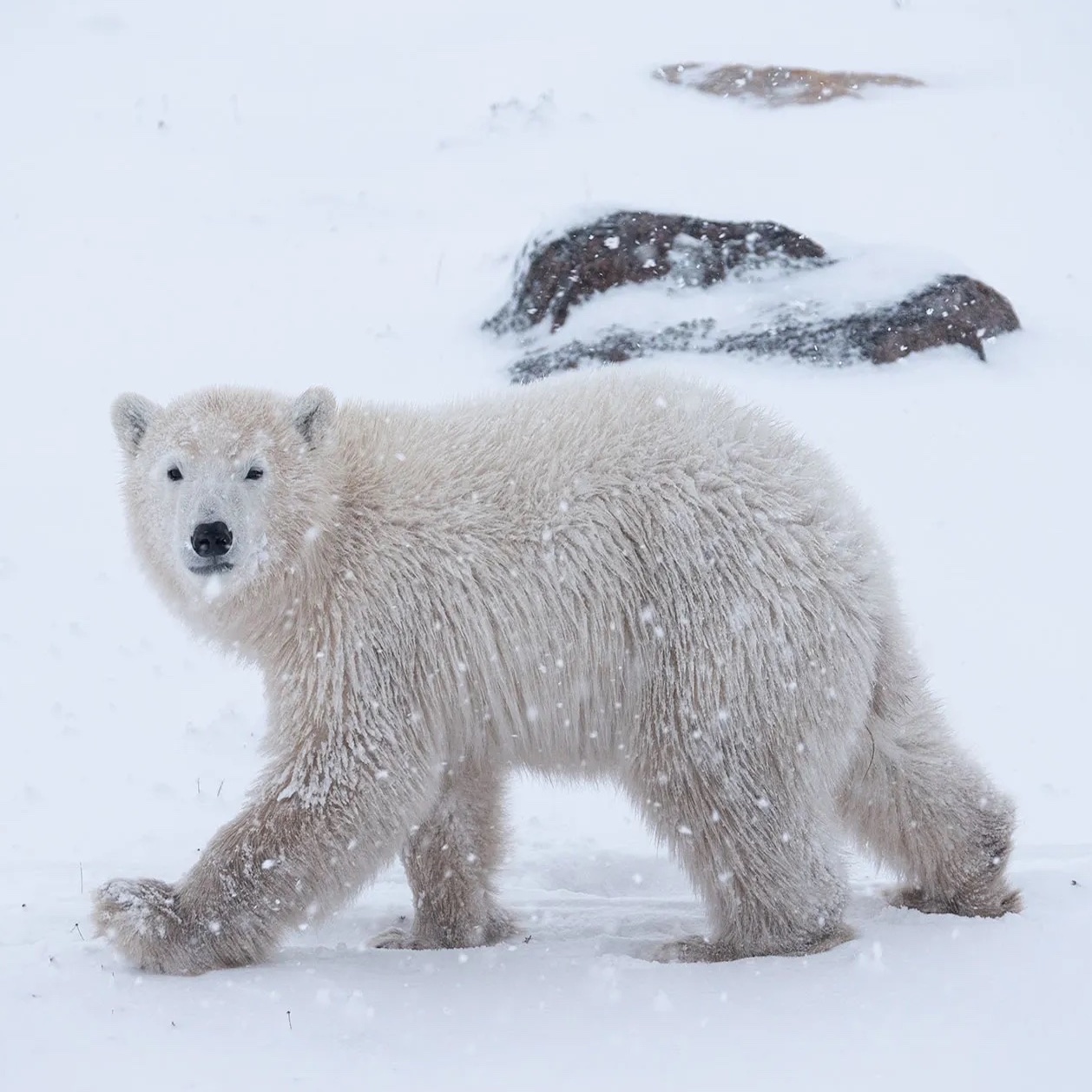} &
        \includegraphics[width=0.15\textwidth]{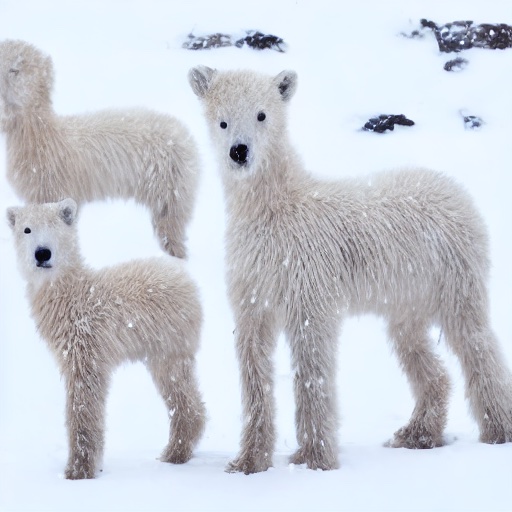} &

        \includegraphics[width=0.15\textwidth]{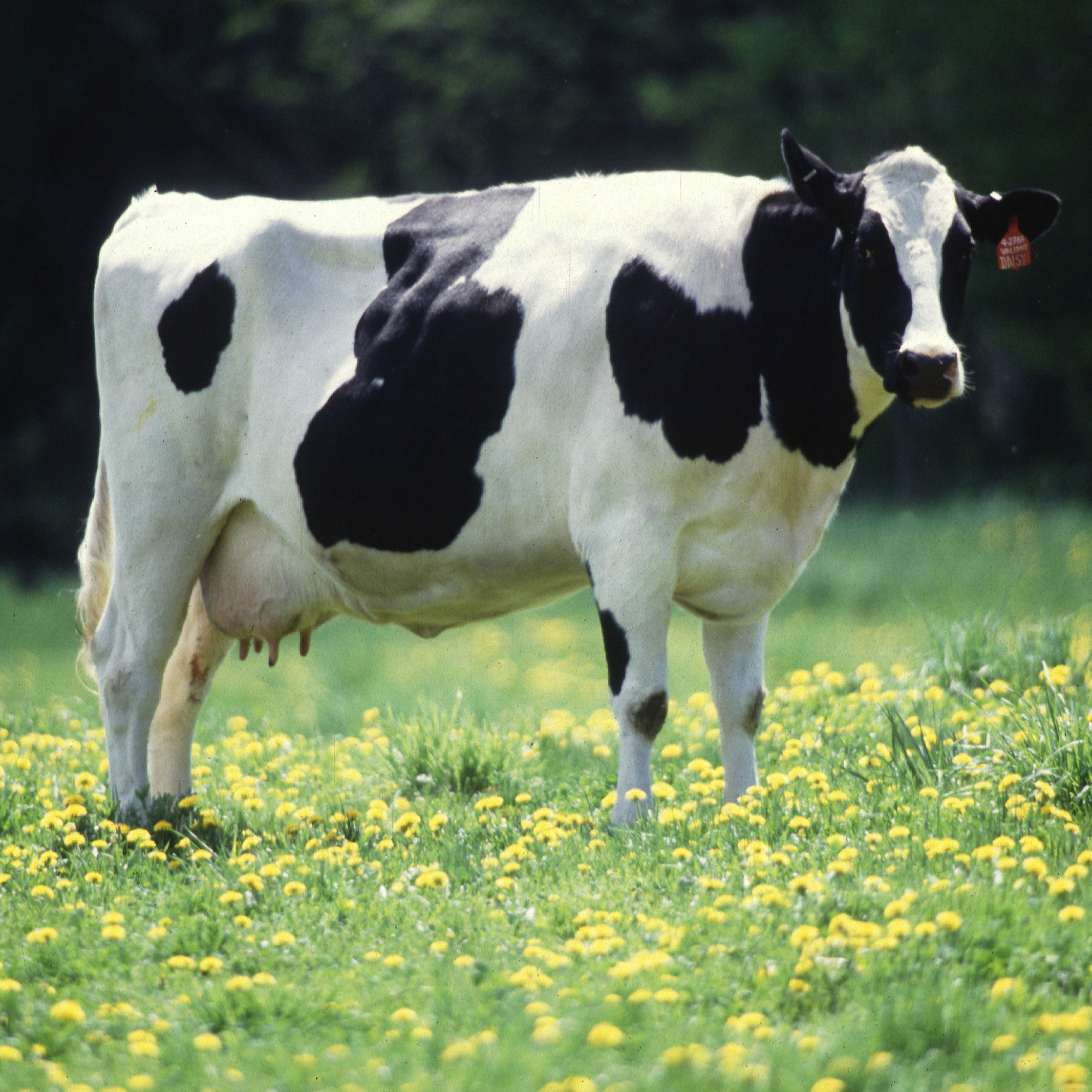} &
        \includegraphics[width=0.15\textwidth]{images/inputs/animals/zebras.jpg} &
        \includegraphics[width=0.15\textwidth]{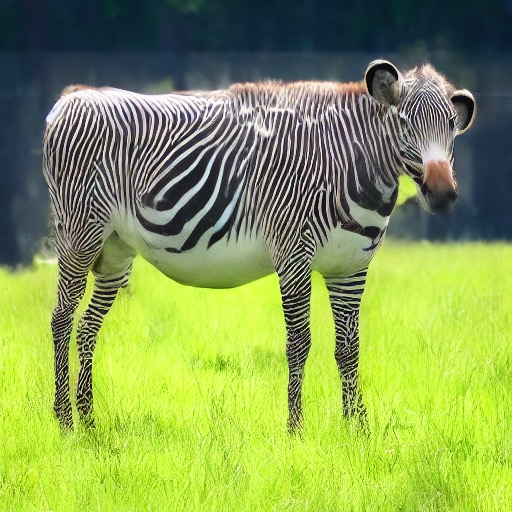} \\

        Structure & Appearance & Output & Structure & Appearance & Output \\ \\
        
        \includegraphics[width=0.15\textwidth]{images/inputs/animals/horse.jpg} &
        \includegraphics[width=0.15\textwidth]{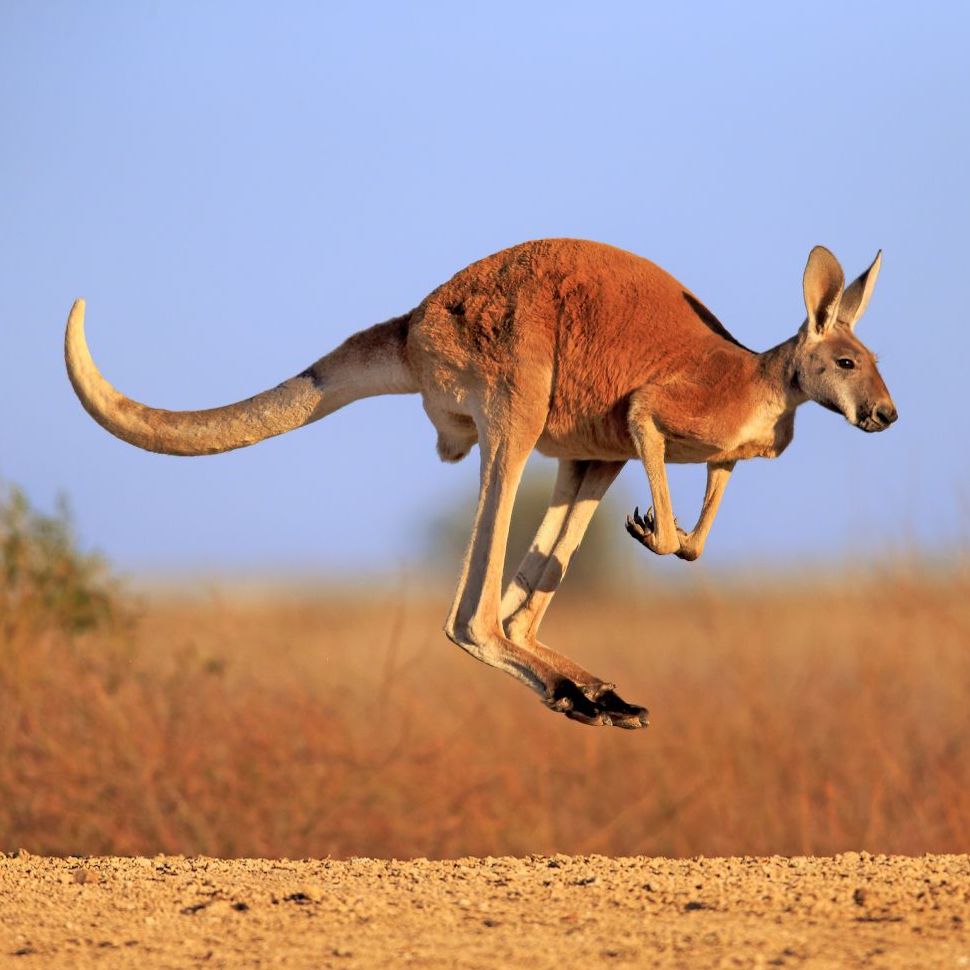} &
        \includegraphics[width=0.15\textwidth]{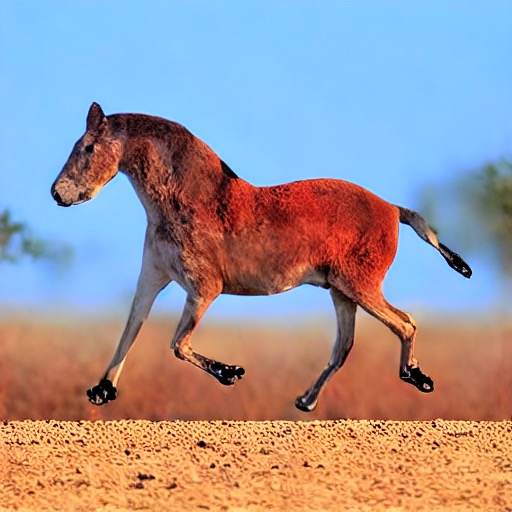} &

        \includegraphics[width=0.15\textwidth]{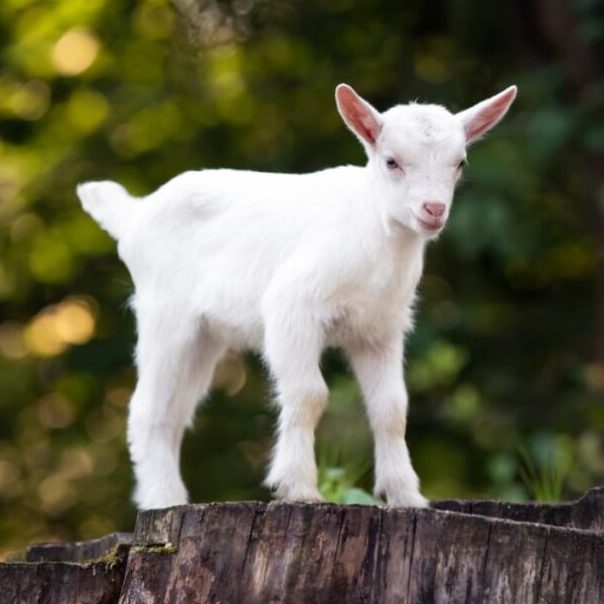} &
        \includegraphics[width=0.15\textwidth]{images/inputs/animals/tiger.jpg} &
        \includegraphics[width=0.15\textwidth]{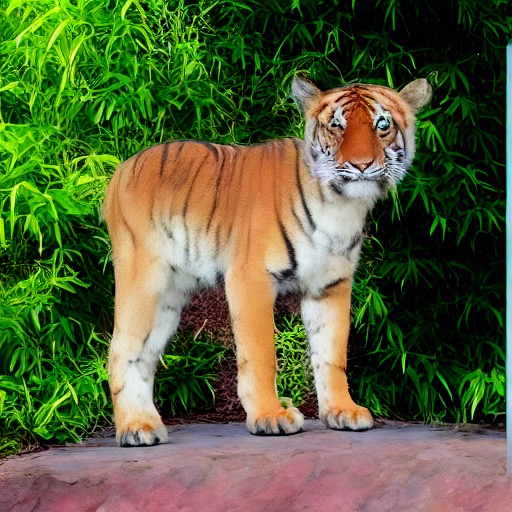} \\

        Structure & Appearance & Output & Structure & Appearance & Output \\ \\

        \includegraphics[width=0.15\textwidth]{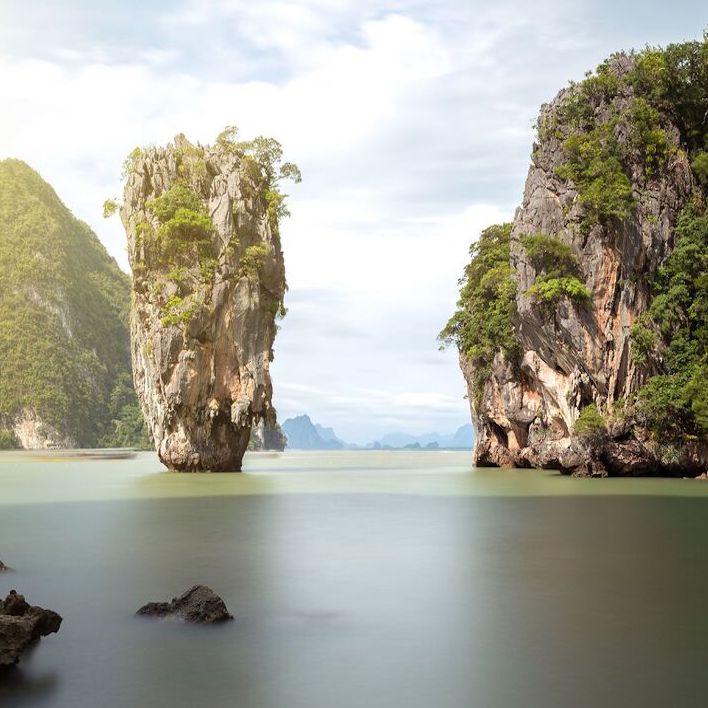} &
        \includegraphics[width=0.15\textwidth]{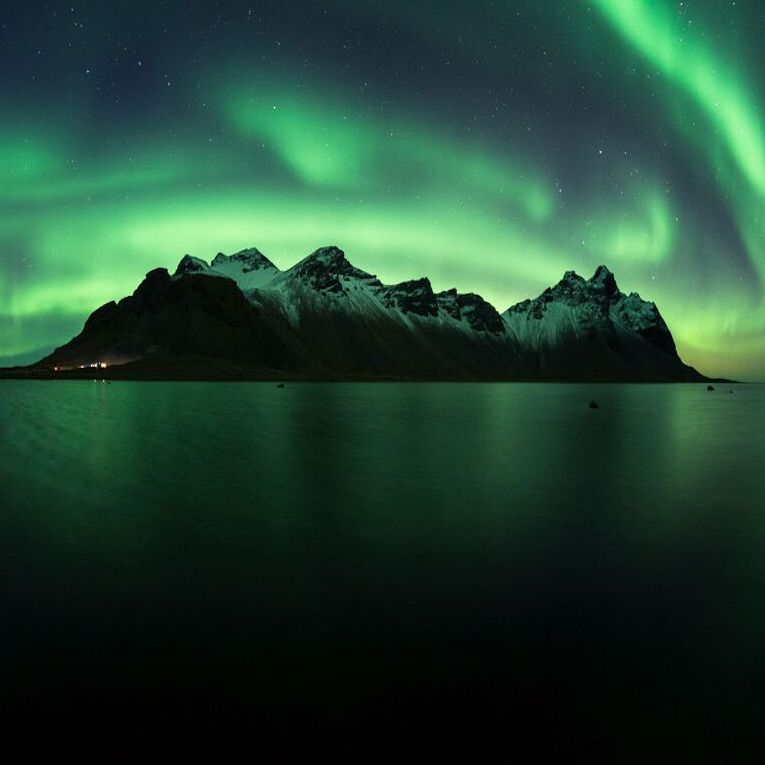} &
        \includegraphics[width=0.15\textwidth]{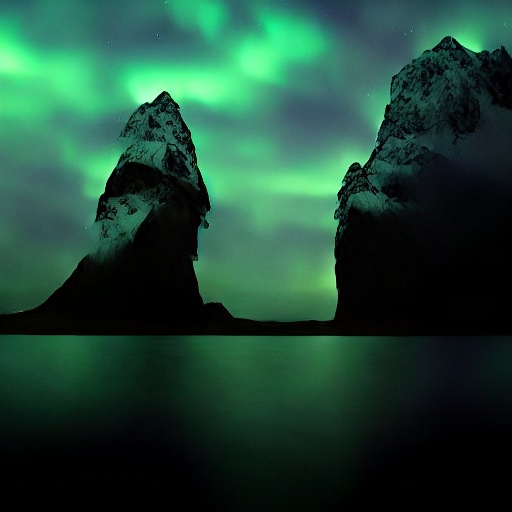} &

        \includegraphics[width=0.15\textwidth]{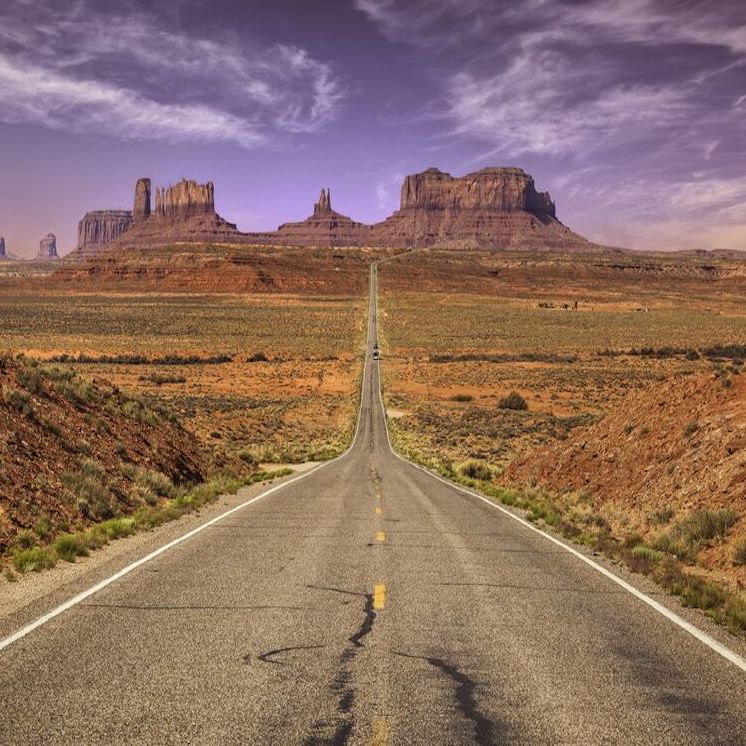} &
        \includegraphics[width=0.15\textwidth]{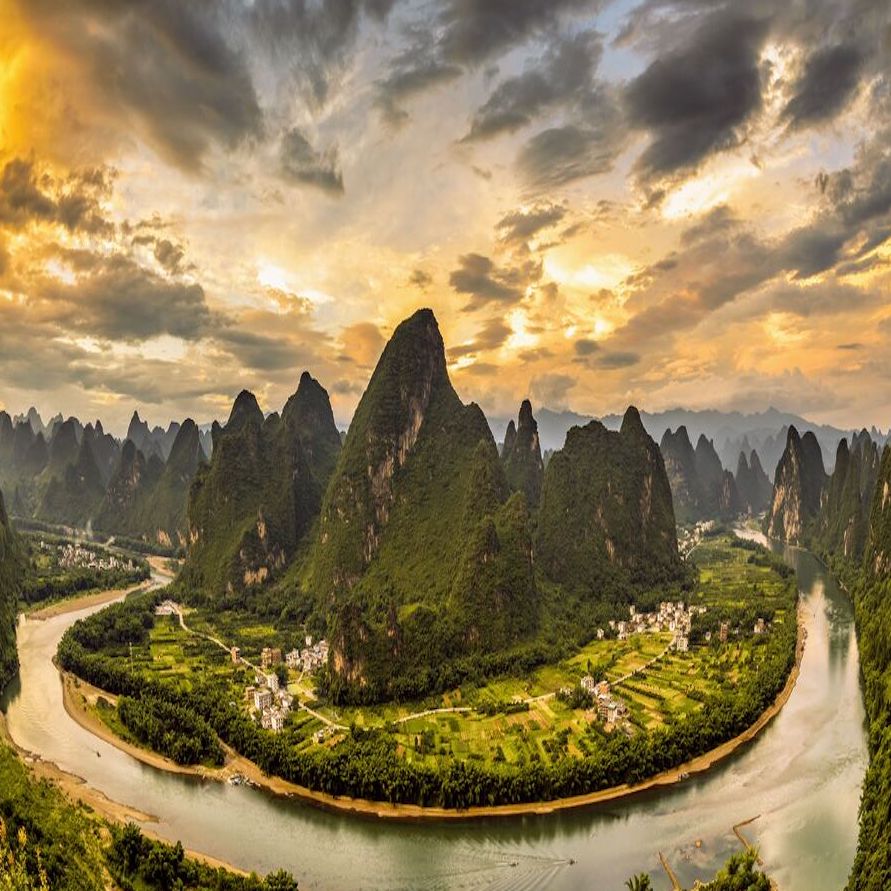} &
        \includegraphics[width=0.15\textwidth]{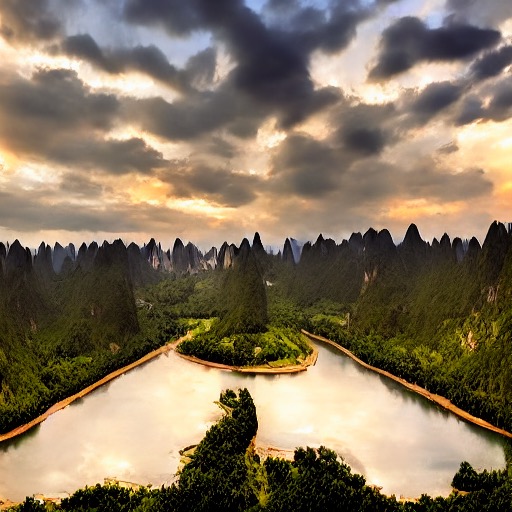} \\

        Structure & Appearance & Output & Structure & Appearance & Output \\ \\

    \end{tabular}
    }
    \vspace{-0.2cm}
    \caption{Additional appearance transfer results obtained by our method.}
    \label{fig:additional_results_pairs}
\end{figure*}
\begin{figure*}
    \centering
    \setlength{\tabcolsep}{0.5pt}
    \addtolength{\belowcaptionskip}{-10pt}
    {
    \begin{tabular}{c c c@{\hspace{0.4cm}} c c c@{\hspace{0.4cm}} c c c}

        \includegraphics[width=0.15\textwidth]{images/inputs/cake/square_yellow_cake.jpg} &
        \includegraphics[width=0.15\textwidth]{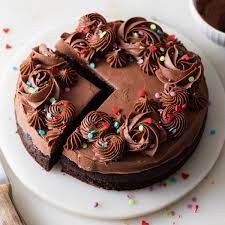} &
        \includegraphics[width=0.15\textwidth]{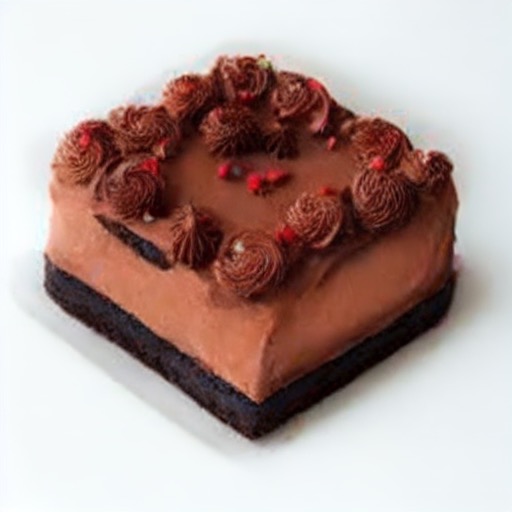} &

        \includegraphics[width=0.15\textwidth]{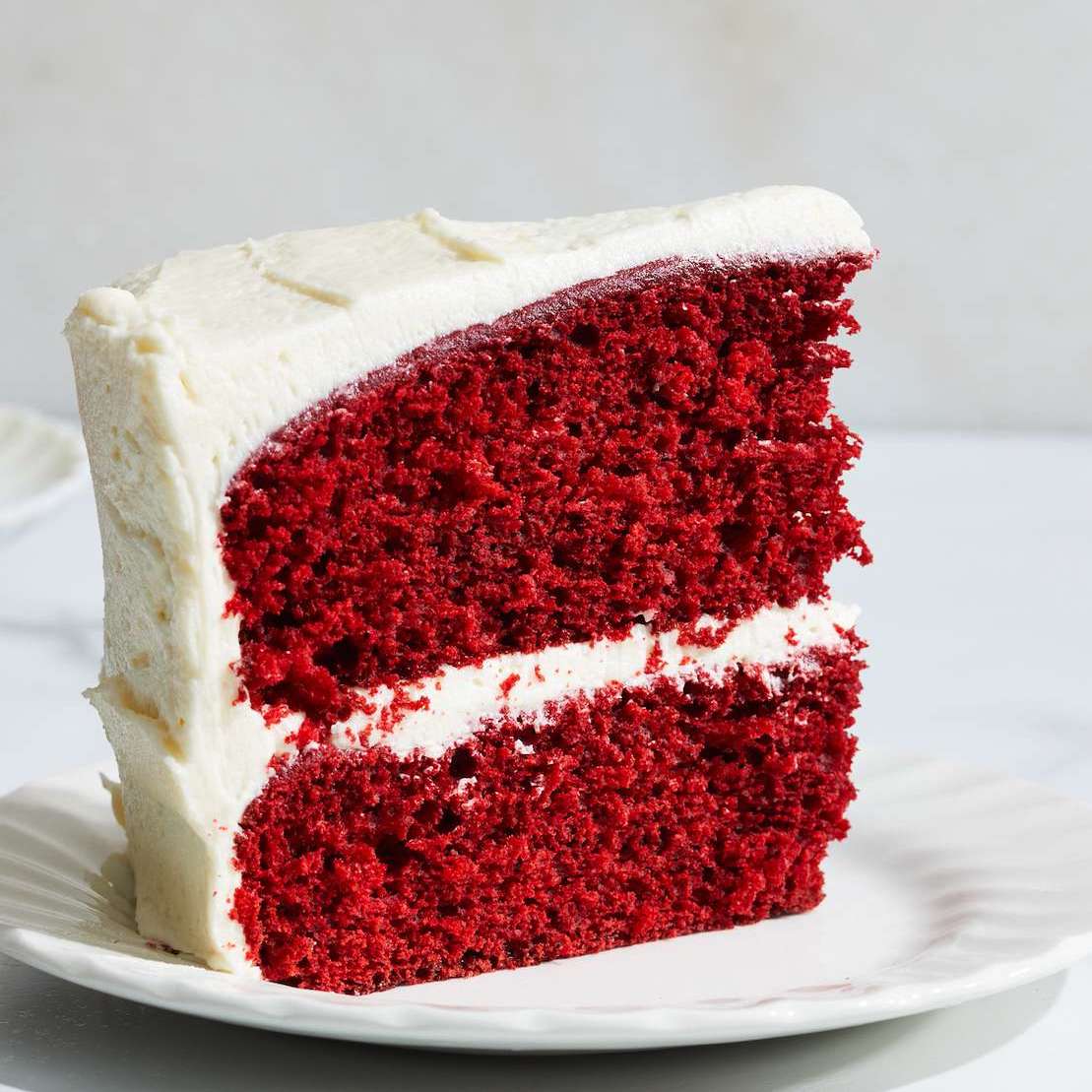} &
        \includegraphics[width=0.15\textwidth]{images/inputs/cake/vanilla_doughnut.jpg} &
        \includegraphics[width=0.15\textwidth]{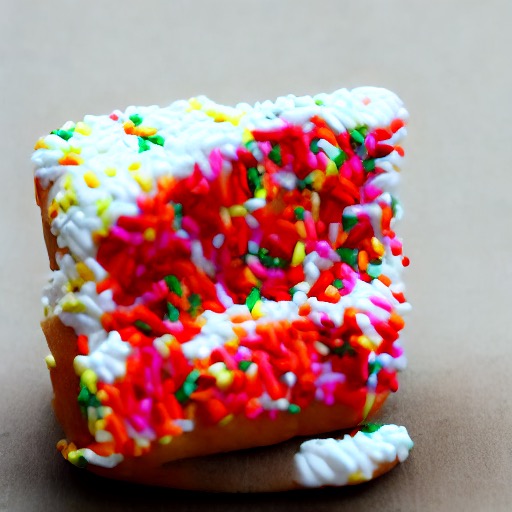} \\

        Structure & Appearance & Output & Structure & Appearance & Output \\ \\

        \includegraphics[width=0.15\textwidth]{images/inputs/cake/fruit.jpg} &
        \includegraphics[width=0.15\textwidth]{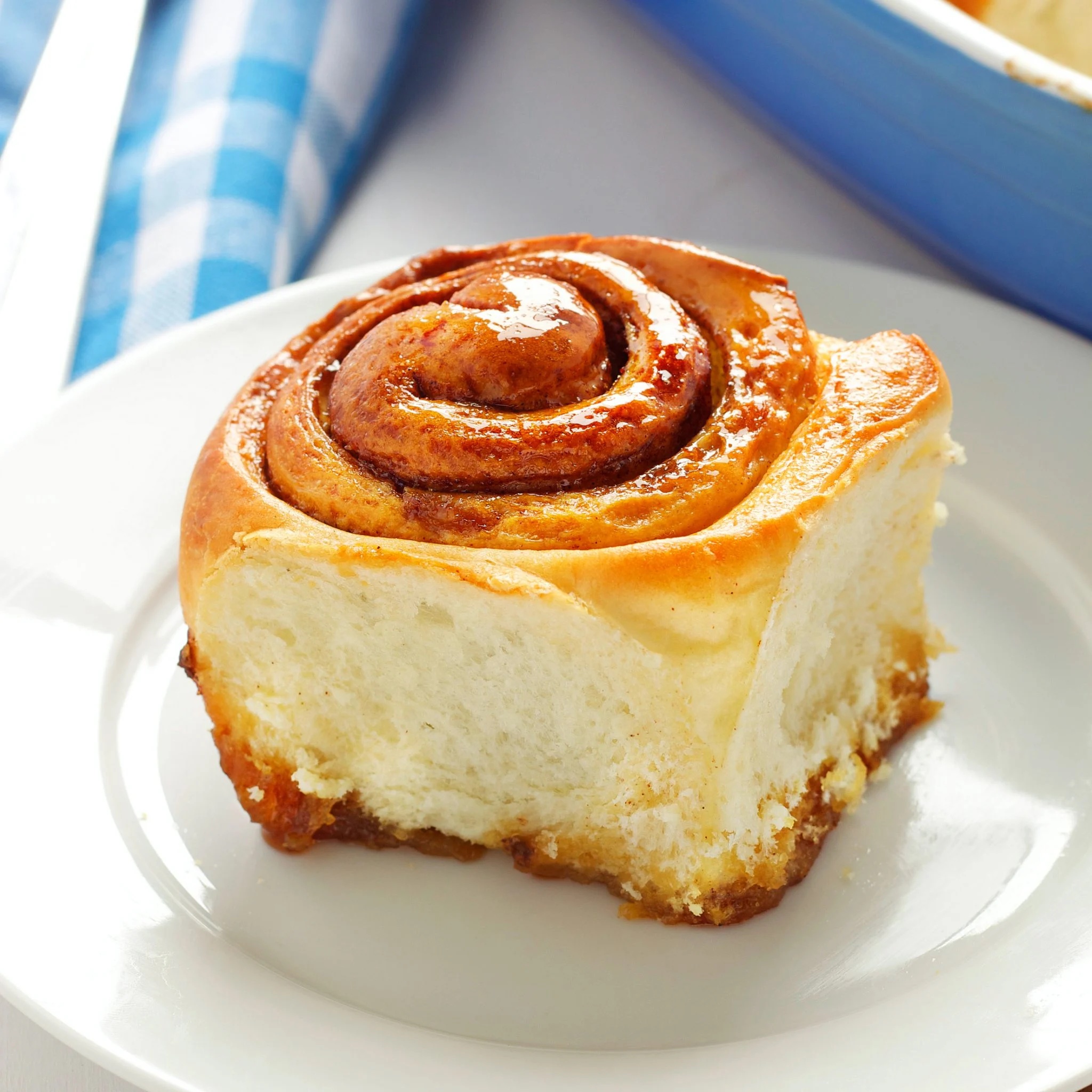} &
        \includegraphics[width=0.15\textwidth]{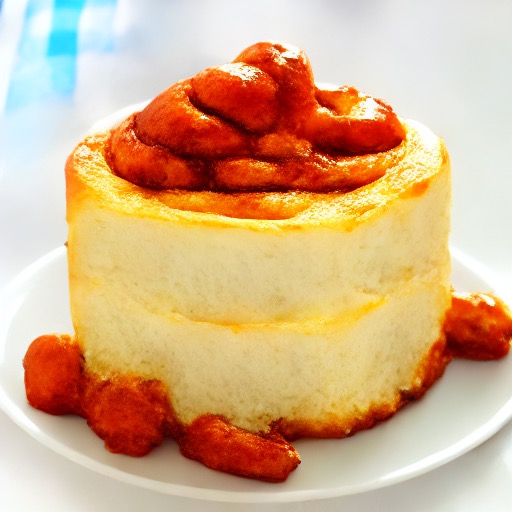} &

        \includegraphics[width=0.15\textwidth]{images/inputs/cake/fruit.jpg} &
        \includegraphics[width=0.15\textwidth]{images/inputs/cake/raspberry_cupcake.jpeg} &
        \includegraphics[width=0.15\textwidth]{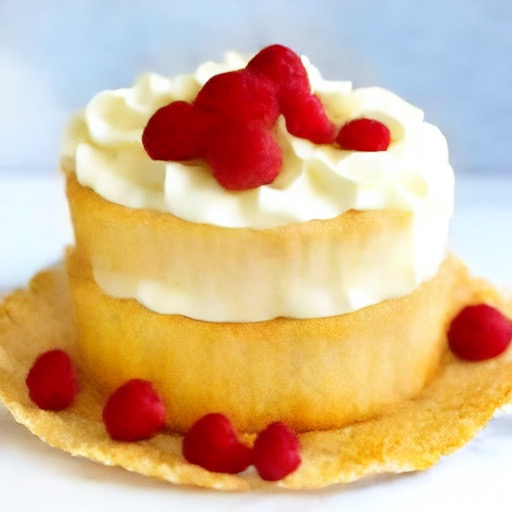} \\

        Structure & Appearance & Output & Structure & Appearance & Output \\ \\

        \includegraphics[width=0.15\textwidth]{images/inputs/birds/hummingbird.jpg} &
        \includegraphics[width=0.15\textwidth]{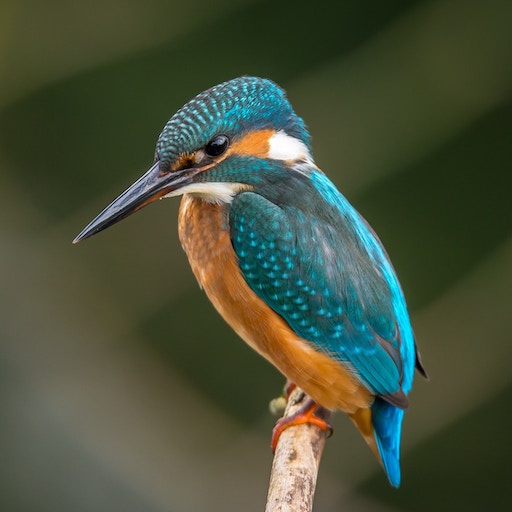} &
        \includegraphics[width=0.15\textwidth]{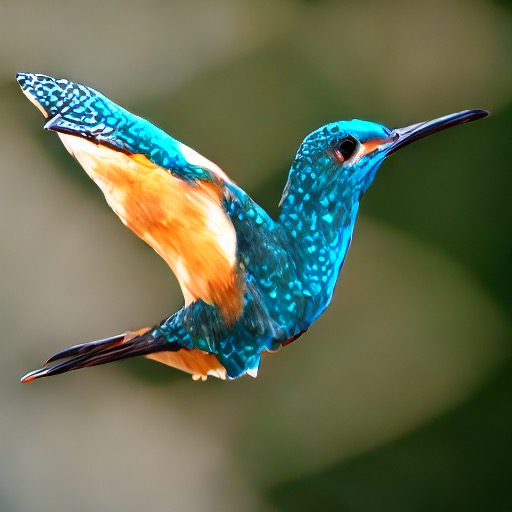} &

        \includegraphics[width=0.15\textwidth]{images/inputs/birds/flying_parrot.jpeg} &
        \includegraphics[width=0.15\textwidth]{images/inputs/birds/blue_jay.jpg} &
        \includegraphics[width=0.15\textwidth]{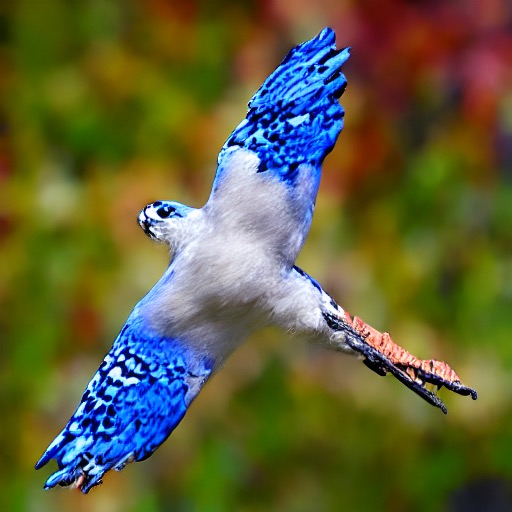} \\

        Structure & Appearance & Output & Structure & Appearance & Output \\ \\
        
        \includegraphics[width=0.15\textwidth]{images/inputs/house/house_1.jpg} &
        \includegraphics[width=0.15\textwidth]{images/inputs/house/wooden_house.jpg} &
        \includegraphics[width=0.15\textwidth]{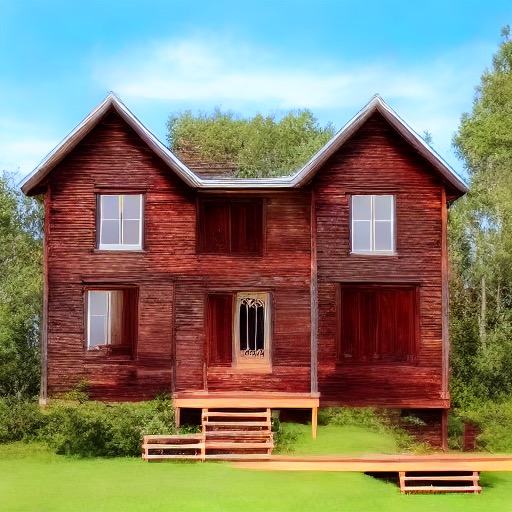} &

        \includegraphics[width=0.15\textwidth]{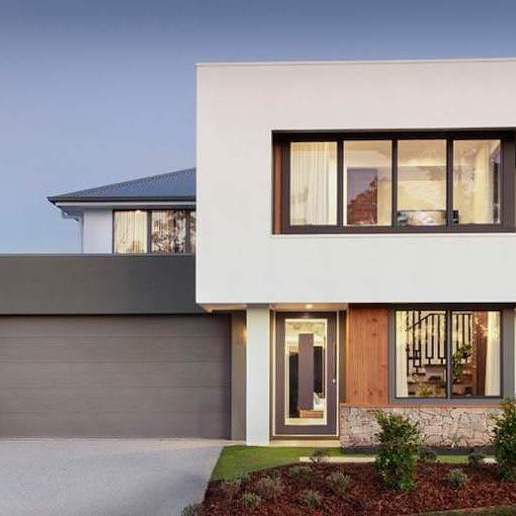} &
        \includegraphics[width=0.15\textwidth]{images/inputs/house/house_1.jpg} &
        \includegraphics[width=0.15\textwidth]{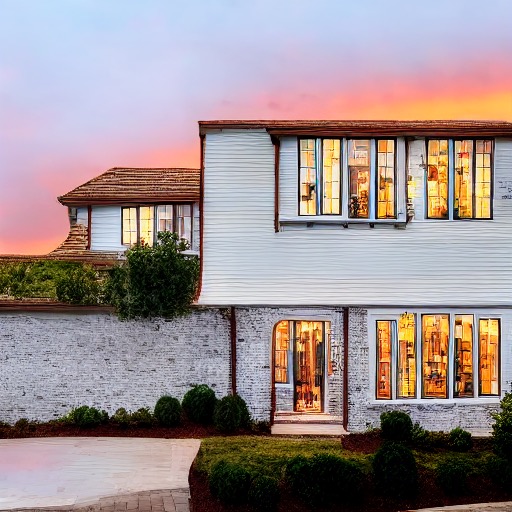} \\

        Structure & Appearance & Output & Structure & Appearance & Output \\ \\
        
        \includegraphics[width=0.15\textwidth]{images/inputs/fruits/oranges.jpg} &
        \includegraphics[width=0.15\textwidth]{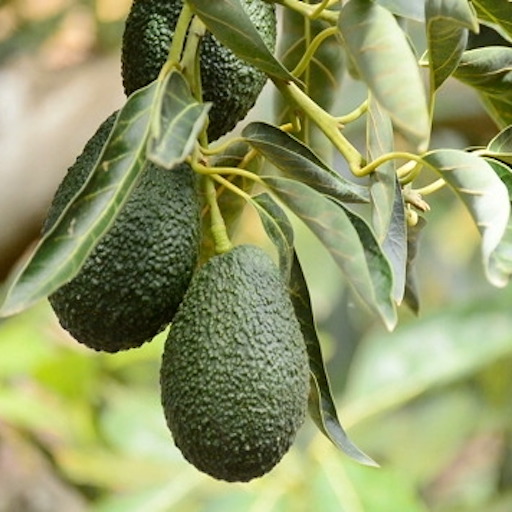} &
        \includegraphics[width=0.15\textwidth]{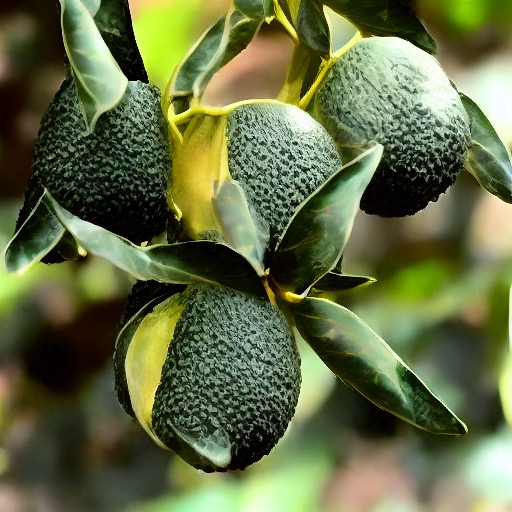} &

        \includegraphics[width=0.15\textwidth]{images/inputs/fruits/apples.jpg} &
        \includegraphics[width=0.15\textwidth]{images/inputs/fruits/oranges.jpg} &
        \includegraphics[width=0.15\textwidth]{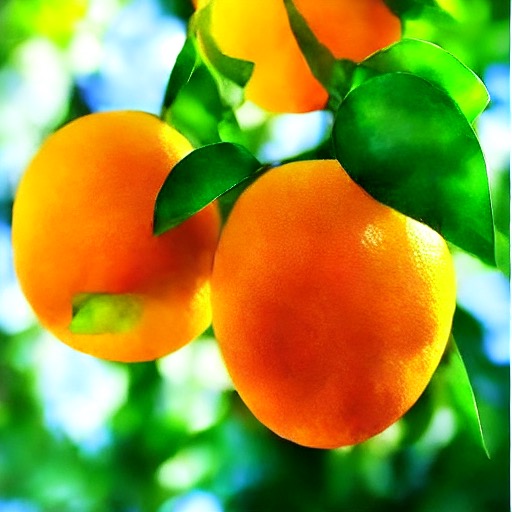} \\

        Structure & Appearance & Output & Structure & Appearance & Output \\ \\

        \includegraphics[width=0.15\textwidth]{images/inputs/fish/fish1.jpg} &
        \includegraphics[width=0.15\textwidth]{images/inputs/fish/fish6.jpg} &
        \includegraphics[width=0.15\textwidth]{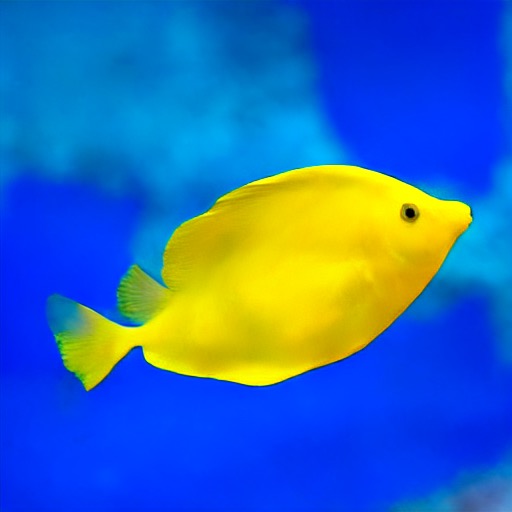} &

        \includegraphics[width=0.15\textwidth]{images/inputs/food/pizza.jpeg} &
        \includegraphics[width=0.15\textwidth]{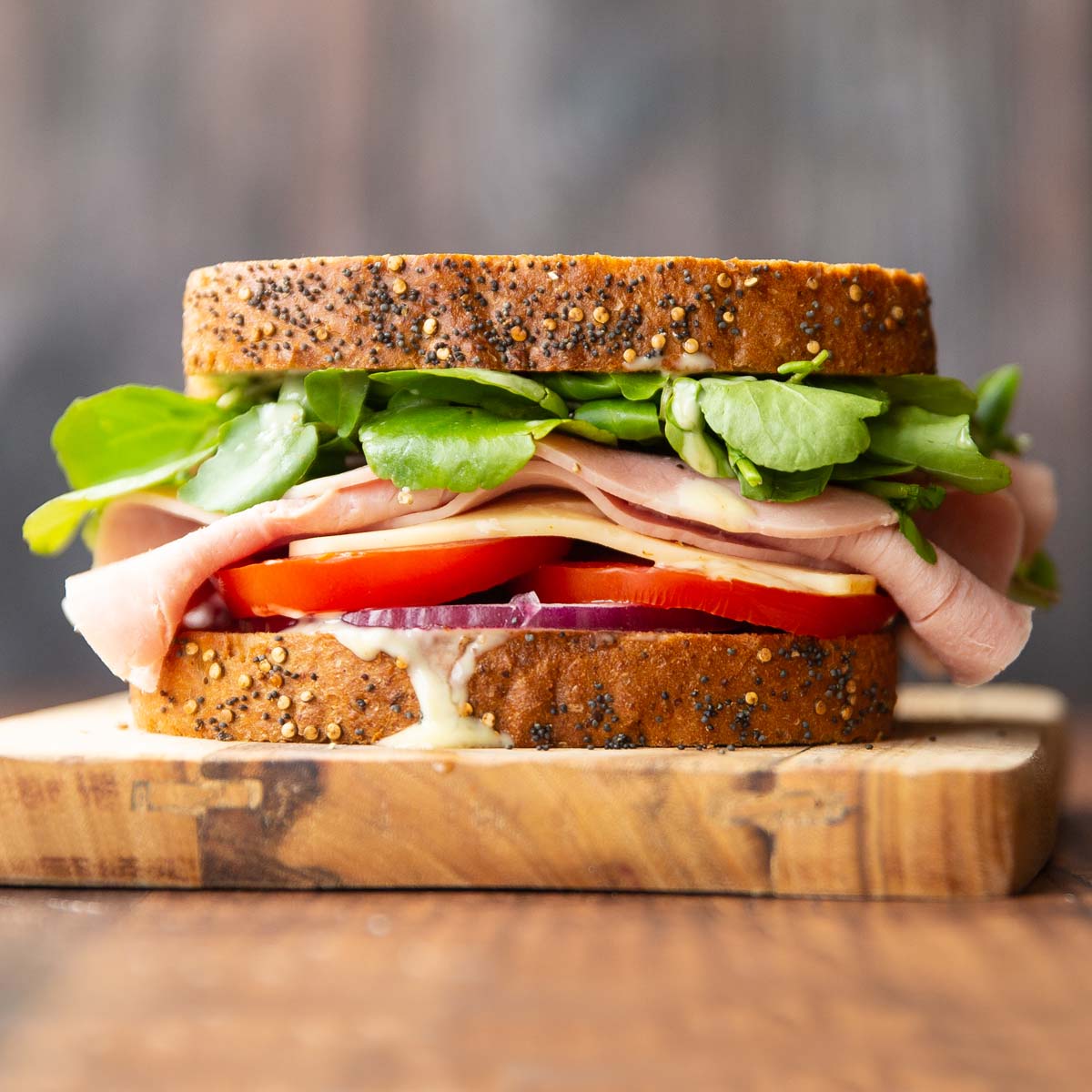} &
        \includegraphics[width=0.15\textwidth]{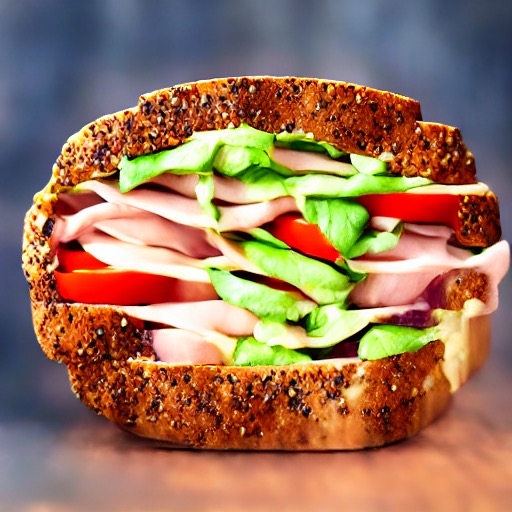} \\

        Structure & Appearance & Output & Structure & Appearance & Output \\ \\
        
    \end{tabular}
    }
    \vspace{-0.2cm}
    \caption{Additional appearance transfer results obtained by our method.}
    \label{fig:additional_results_pairs_2}
\end{figure*}
\begin{figure*}
    \centering
    \setlength{\tabcolsep}{0.5pt}
    \renewcommand{\arraystretch}{0.3}
    \addtolength{\belowcaptionskip}{-5pt}
    {

    \hspace{-0.2cm}
    \begin{minipage}{0.5\textwidth}
        \centering
        \begin{tabular}{c c c c}

        \\ \\ \\

        \includegraphics[width=0.24\textwidth]{images/struct_app.png} &
        \includegraphics[width=0.24\textwidth]{images/inputs/buildings/saint_basil.jpg} &
        \includegraphics[width=0.24\textwidth]{images/inputs/buildings/taj_mahal.jpg} &
        \includegraphics[width=0.24\textwidth]{images/inputs/buildings/Le_sacre_coeur.jpg} \\

        \includegraphics[width=0.24\textwidth]{images/inputs/buildings/saint_basil.jpg} &
        \includegraphics[width=0.24\textwidth]{images/our_results_grid/buildings/basil_to_basil.jpg} &
        \includegraphics[width=0.24\textwidth]{images/our_results_grid/buildings/basil_to_taj.jpg} &
        \includegraphics[width=0.24\textwidth]{images/our_results_grid/buildings/basil_to_sacre.jpg} \\
 
        \includegraphics[width=0.24\textwidth]{images/inputs/buildings/taj_mahal.jpg} &
        \includegraphics[width=0.24\textwidth]{images/our_results_grid/buildings/taj_to_basil.jpg} &
        \includegraphics[width=0.24\textwidth]{images/our_results_grid/buildings/taj_to_taj.jpg} &
        \includegraphics[width=0.24\textwidth]{images/our_results_grid/buildings/taj_to_sacre.jpg} \\
        
        \includegraphics[width=0.24\textwidth]{images/inputs/buildings/Le_sacre_coeur.jpg} &
        \includegraphics[width=0.24\textwidth]{images/our_results_grid/buildings/sacre_to_basil.jpg} &
        \includegraphics[width=0.24\textwidth]{images/our_results_grid/buildings/sacre_to_taj.jpg} &
        \includegraphics[width=0.24\textwidth]{images/our_results_grid/buildings/sacre_to_sacre.jpg} \\
        
        \end{tabular}
        
    \end{minipage}%
    \begin{minipage}{0.5\textwidth}
        \centering
        \begin{tabular}{c c c c}

        \\ \\ \\

        \includegraphics[width=0.24\textwidth]{images/struct_app.png} &
        \includegraphics[width=0.24\textwidth]{images/inputs/cars/vintage.jpg} &
        \includegraphics[width=0.24\textwidth]{images/inputs/cars/red_vintage.jpg} &
        \includegraphics[width=0.24\textwidth]{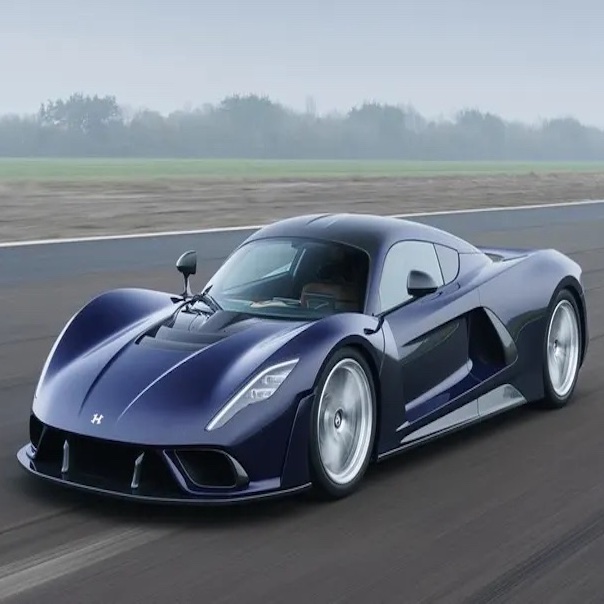} \\

        \includegraphics[width=0.24\textwidth]{images/inputs/cars/vintage.jpg} &
        \includegraphics[width=0.24\textwidth]{images/our_results_grid/car/blue_vintage_to_blue_vintage.jpg} &
        \includegraphics[width=0.24\textwidth]{images/our_results_grid/car/blue_vintage_to_red_vintage.jpg} &
        \includegraphics[width=0.24\textwidth]{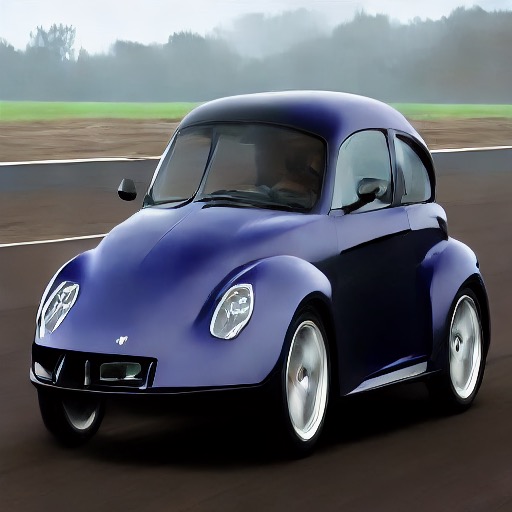} \\

        \includegraphics[width=0.24\textwidth]{images/inputs/cars/red_vintage.jpg} &
        \includegraphics[width=0.24\textwidth]{images/our_results_grid/car/red_vintage_to_blue_vintage.jpg} &
        \includegraphics[width=0.24\textwidth]{images/our_results_grid/car/red_vintage_to_red_vintage.jpg} &
        \includegraphics[width=0.24\textwidth]{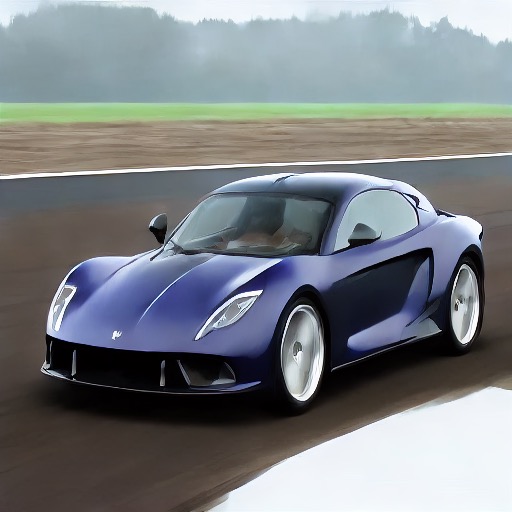} \\
        
        \includegraphics[width=0.24\textwidth]{images/inputs/cars/black_sports.jpeg} &
        \includegraphics[width=0.24\textwidth]{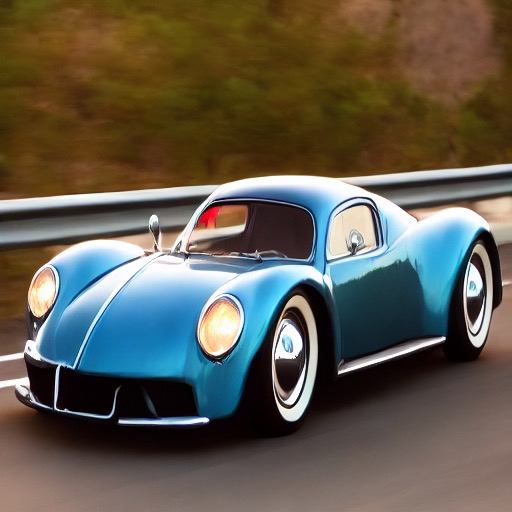} &
        \includegraphics[width=0.24\textwidth]{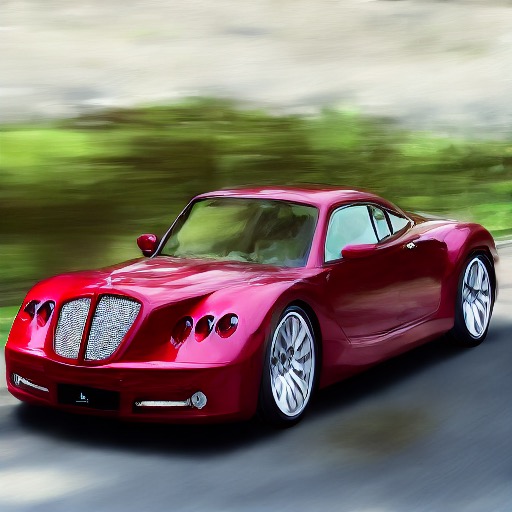} &
        \includegraphics[width=0.24\textwidth]{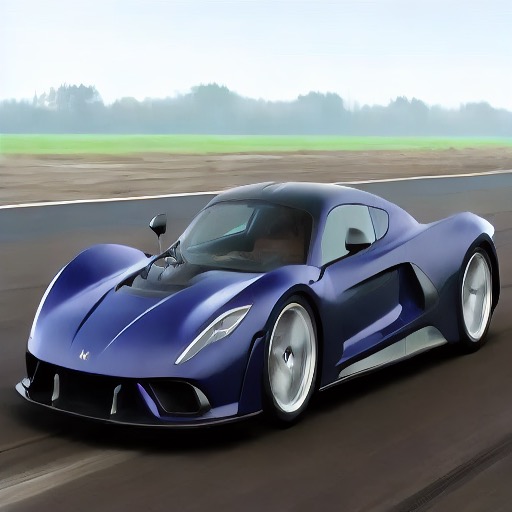} \\
        
        \end{tabular}
        
    \end{minipage}%
    
    \begin{minipage}{0.5\textwidth}
        \centering
        \begin{tabular}{c c c c}

        \\ \\ \\ \\

        \includegraphics[width=0.24\textwidth]{images/struct_app.png} &
        \includegraphics[width=0.24\textwidth]{images/inputs/birds/hummingbird.jpg} &
        \includegraphics[width=0.24\textwidth]{images/inputs/birds/lilac_roller.jpg} &
        \includegraphics[width=0.24\textwidth]{images/inputs/birds/small_orange.jpg} \\

        \includegraphics[width=0.24\textwidth]{images/inputs/birds/hummingbird.jpg} &
        \includegraphics[width=0.24\textwidth]{images/our_results_grid/birds/hummingbird_to_hummingbird.jpg} &
        \includegraphics[width=0.24\textwidth]{images/our_results_grid/birds/hummingbird_to_lilac.jpg} &
        \includegraphics[width=0.24\textwidth]{images/our_results_grid/birds/hummingbird_to_orange_bird.jpg} \\

        \includegraphics[width=0.24\textwidth]{images/inputs/birds/lilac_roller.jpg} &
        \includegraphics[width=0.24\textwidth]{images/our_results_grid/birds/lilac_to_hummingbird.jpg} &
        \includegraphics[width=0.24\textwidth]{images/our_results_grid/birds/lilac_to_lilac.jpg} &
        \includegraphics[width=0.24\textwidth]{images/our_results_grid/birds/lilac_to_orange_bird.jpg} \\
        
        \includegraphics[width=0.24\textwidth]{images/inputs/birds/small_orange.jpg} &
        \includegraphics[width=0.24\textwidth]{images/our_results_grid/birds/orange_bird_to_hummingbird.jpg} &
        \includegraphics[width=0.24\textwidth]{images/our_results_grid/birds/orange_bird_to_lilac.jpg} &
        \includegraphics[width=0.24\textwidth]{images/our_results_grid/birds/small_orange_to_small_orange.jpg} \\
        \end{tabular}
        
    \end{minipage}%

    }

    \caption{
    Enlarged versions of our appearance transfer results from~\Cref{fig:our_results_grid}.
    }
    \label{fig:our_results_grid_big}
\end{figure*}

\end{document}